\documentclass[acmtog]{acmart}

\acmSubmissionID{1687}

\makeatletter
\DeclareRobustCommand\onedot{\futurelet\@let@token\@onedot}
\def\@onedot{\ifx\@let@token.\else.\null\fi\xspace}

\makeatother

\newcommand{\ignore}[1]{}   

\usepackage{enumitem}
\usepackage{array}   
\usepackage{makecell} 
\usepackage{subcaption}

\AtBeginDocument{%
  \providecommand\BibTeX{{%
    \normalfont B\kern-0.5em{\scshape i\kern-0.25em b}\kern-0.8em\TeX}}}

\copyrightyear{2025}
\acmYear{2025}
\setcopyright{cc}
\setcctype{by}
\acmConference[SA Conference Papers '25]{SIGGRAPH Asia 2025 Conference Papers}{December 15--18, 2025}{Hong Kong, Hong Kong}
\acmBooktitle{SIGGRAPH Asia 2025 Conference Papers (SA Conference Papers '25), December 15--18, 2025, Hong Kong, Hong Kong}\acmDOI{10.1145/3757377.3763910}
\acmISBN{979-8-4007-2137-3/2025/12}

\usepackage{float}
\usepackage{placeins}
\usepackage{dblfloatfix}
\usepackage{multirow}
\begin{document}
\title{Prior-Enhanced Gaussian Splatting for Dynamic Scene Reconstruction from Casual Video}

\newcommand{\emptyline}{\relax} 
\author{Meng-Li Shih}
\affiliation{%
  \institution{University of Washington}
  \city{Seattle}
  \state{Washington}
  \country{\emptyline}
}
\email{mlshih@cs.washington.edu}

\author{Ying-Huan Chen}
\affiliation{%
  \institution{National Yang Ming Chiao Tung University}
  \city{Hsinchu City}
  \country{\emptyline}
}
\email{yinghuan0419@gmail.com}

\author{Yu-Lun Liu}
\affiliation{%
  \institution{National Yang Ming Chiao Tung University}
  \city{Hsinchu City}
  \country{\emptyline}
}
\email{yulunliu@cs.nycu.edu.tw}

\author{Brian Curless}
\affiliation{%
  \institution{University of Washington}
  \city{Seattle}
  \state{Washington}
  \country{\emptyline}
}
\email{curless@cs.washington.edu}

\newcommand{\yulunliu}[1]{{\textcolor{red}{[yulunliu: #1]}}}

\newcommand{\yhchen}[1]{{\textcolor{orange}{[yhchen: #1]}}}

\begin{abstract}
We introduce a fully automatic pipeline for dynamic scene reconstruction from casually captured monocular RGB videos.
Rather than designing a new scene representation, we enhance the \emph{priors} that drive Dynamic Gaussian Splatting.
Video segmentation combined with epipolar-error maps yields object-level masks that closely follow thin structures; these masks (i) guide an object-depth loss that sharpens the consistent video depth, and (ii) support skeleton-based sampling plus mask-guided re-identification to produce reliable, comprehensive 2-D tracks.
Two additional objectives embed the refined priors in the reconstruction stage: a virtual-view depth loss removes floaters, and a scaffold-projection loss ties motion nodes to the tracks, preserving fine geometry and coherent motion.
The resulting system surpasses previous monocular dynamic scene reconstruction methods and delivers visibly superior renderings. Project page: \href{https://priorenhancedgaussian.github.io/}{\textcolor{magenta}{https://priorenhancedgaussian.github.io/}}
\end{abstract}


\begin{teaserfigure}
  \centering
  \includegraphics[width=1.0\textwidth]{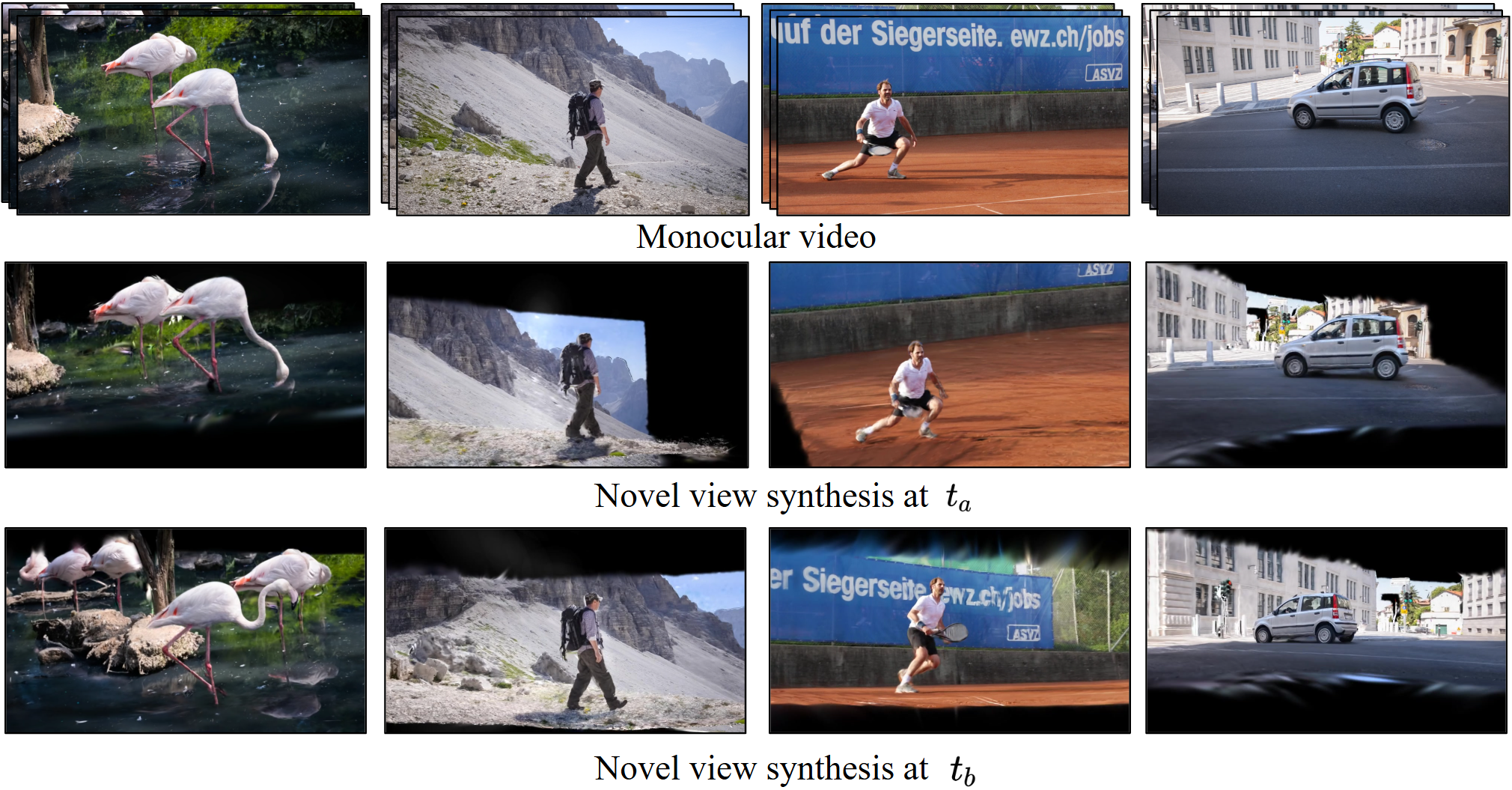}
  \caption{\textbf{Dynamic free-view synthesis of reconstruction from monocular video.}}
  \Description{}
  \label{fig:teaser}
\end{teaserfigure}


\maketitle
\section{Introduction}
\label{sec:intro}

Reconstructing a faithful, time‐varying 3-D scene from a casual hand-held
monocular RGB video is a long-standing goal in computer graphics and
vision.  
With the advent of \emph{3-D Gaussian Splatting} (3D-GS)~\cite{kerbl20233d}, a static scene
can be distilled into tens of thousands of anisotropic Gaussians and
rasterized in real time.  
Extending this idea to moving footage leads to
\emph{dynamic Gaussian Splatting} (DGS): Gaussians translate and rotate
through time, enabling free-view playback and immersive AR/VR
experiences (see Sec.~\ref{sec:related} for detail). 

Solving this ill-posed problem requires both a carefully designed representation for motion and geometry, and a diverse set of 2-D priors from foundation models such as consistent depth, dynamic masks, and 2-D point trajectories to initialize and supervise the reconstruction.
Recent systems such as MoSca \cite{lei2024mosca} and Shape-of-Motion \cite{wang2024shape} combine these aspects and achieve striking realism.

\begin{figure}[h]
\centering

\begin{subfigure}[t]{0.495\linewidth}\centering
  \includegraphics[width=\linewidth]{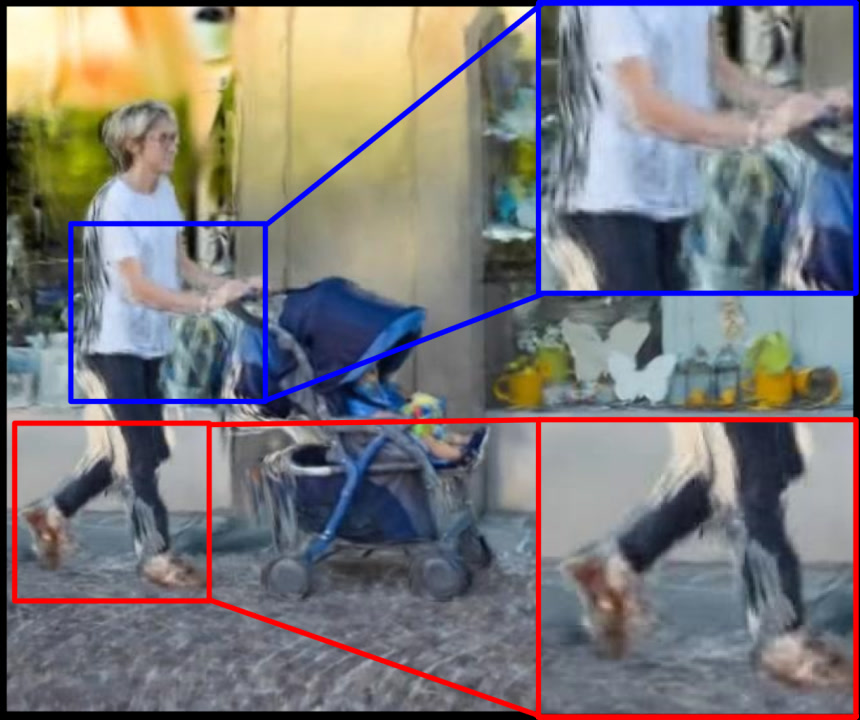}
  \caption{Shape of Motion}
\end{subfigure}\hfill
\begin{subfigure}[t]{0.495\linewidth}\centering
  \includegraphics[width=\linewidth]{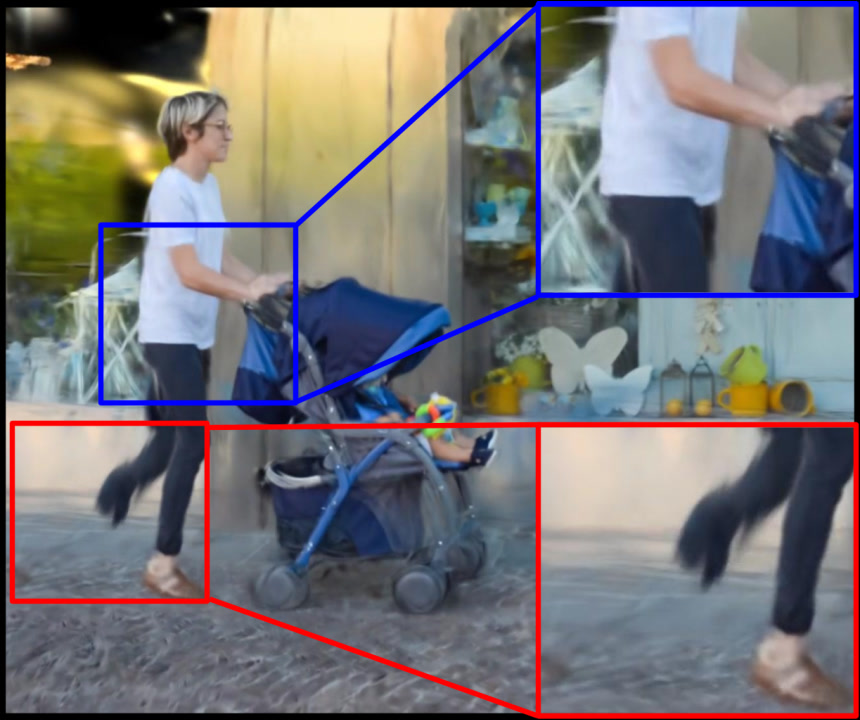}
  \caption{MoSca}
\end{subfigure}

\caption{\textbf{Limitations of existing methods.}
Shape of Motion~\cite{wang2024shape} shows incoherent motion near
boundaries and dis-occlusions of the object (red and blue boxes in (a)),
indicating insufficient regularization. MoSca~\cite{lei2024mosca} improves
coherence by constraining motion with a set of scaffolds, yet it still
struggles to represent thin objects with complex motion, as highlighted by
the red box in (b).}
\label{fig:motivation}
\end{figure}

Yet dynamic objects still reveal the limits of current pipelines: thin structures blur or vanish, and complex motions lose coherence (Fig.~\ref{fig:motivation}). 
Most prior work tackles these artifacts by introducing richer scene
representations or elaborate optimization schedules, while paying little attention to the details of 2-D priors from foundation models. 
We find that the quality of these priors has become one of the major bottlenecks. Upgrading the depth, masks, and tracks fed into a DGS system markedly improves reconstruction quality.
We therefore focus on \textbf{\emph{enhancing prior quality}}. Our pipeline \textbf{(1)} extracts \textbf{salient dynamic-object masks} that tightly follow thin parts such as limbs, \textbf{(2)} \textbf{refines the consistent video depth}, recovering detailed dynamic structures while preserving global geometry, and \textbf{(3)} builds \textbf{robust, comprehensive 2-D trajectories} that survive occlusions and provide thorough coverage of moving surfaces
These higher-fidelity priors supply far stronger supervision.  To let
the model exploit them fully, we further introduce two additional loss terms, \textbf{scaffold-projection loss} and \textbf{virtual-view depth loss}), that
propagate the improved priors directly into the Gaussian cloud and its
underlying motion model. 

We integrate these components and develop a fully automatic pipeline for dynamic-scene reconstruction from monocular RGB video. 
Experiments on the DyCheck dataset show quantitative gains, while qualitative comparisons on DAVIS reveal significant improvements in visual quality over previous monocular DGS systems. 

\section{Related Work}
\label{sec:related}
\begin{figure*}[t]    
\centering
\includegraphics[width=\linewidth]{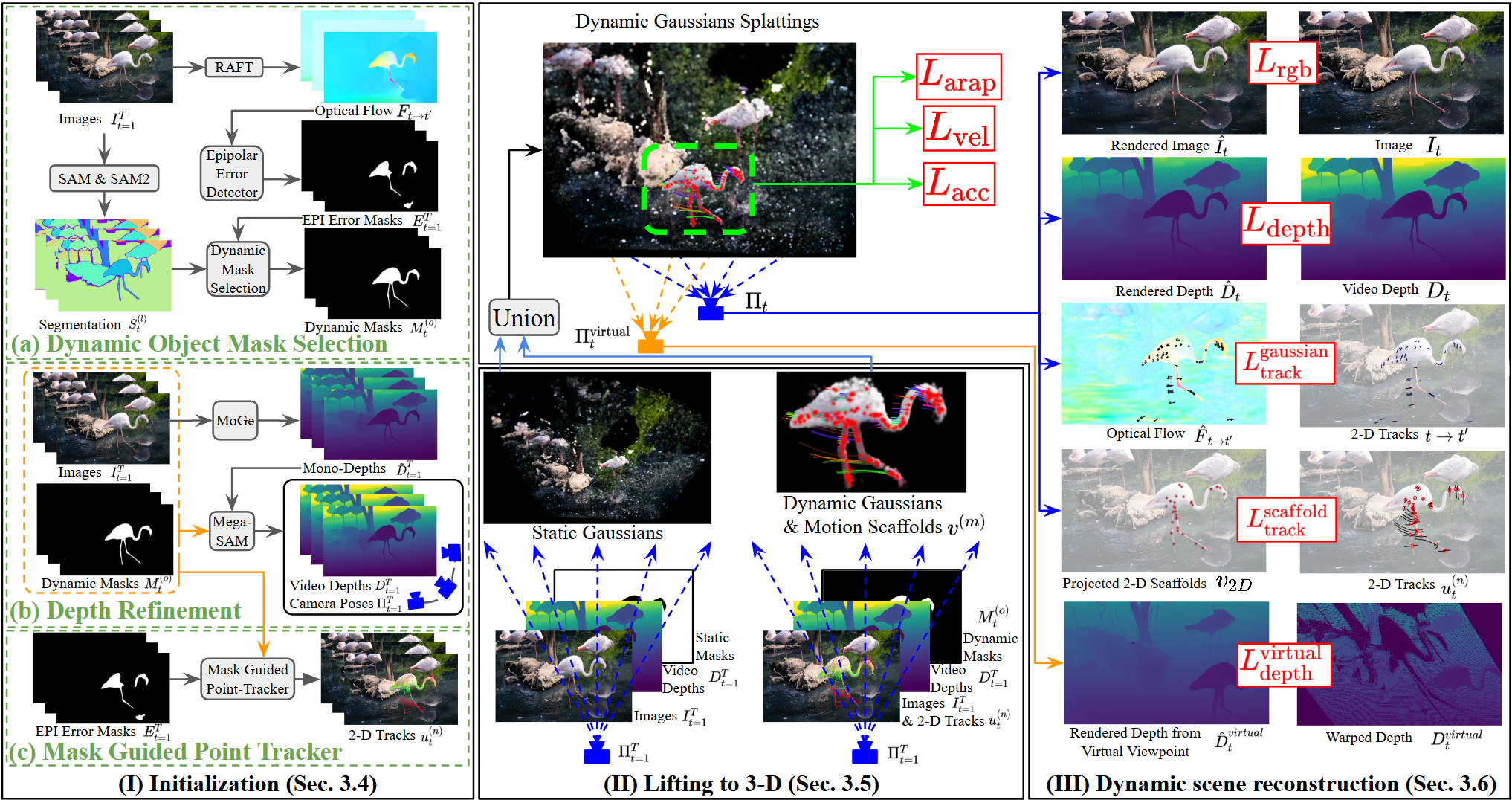}
\caption{
\textbf{Method Overview.}
Our three-stage approach is: (I) Initialization, (II) Lifting to 3-D, and (III) Dynamic scene reconstruction.
\textbf{(I) Initialization(Sec.~\ref{sec:init}).} (a) Video segmentation $S_t^{(l)}$~\cite{kirillov2023segment,ravi2024sam,AutoSeg_SAM2} is combined with epipolar-error masks $E_{t=1}^T$ derived from optical flow $F_{t\rightarrow t'}$~\cite{teed2020raft}, yielding dynamic-object masks $M_t^{(o)}$ that cover each dynamic object in its entirety. (b) Using mask $M_t^{(o)}$ and mono-depths $\tilde D_{t=1}^T$~\cite{wang2024moge}, we apply an object-depth loss that sharpens Mega-SAM’s consistent depths $D_{t=1}^T$~\cite{li2024megasam}. (c) From images $I_{t=1}^T$ and masks $M_t^{(o)}$ we identify and sample tracks $u_t^{(o)}$ along thin structures; combined with uniformly sampled tracks from $E_{t=1}^T$, they fully covering each moving object. Mask-guided re-identification, using the dynamic masks $M_t^{(o)}$, restores tracks lost to occlusion and further improves reconstruction quality.
(II) \textbf{Lifting to 3-D(Sec.~\ref{sec:lift}).} Following \cite{lei2024mosca}, 2-D pixels and tracks are promoted to 3-D Gaussians and motion-scaffold nodes $v^{(m)}$.  A space–time regularizer then produces a motion-coherent 3-D initialization. (III) \textbf{Dynamic scene reconstruction(Sec.~\ref{sec:dyn_recon}).} 
For a training view $\Pi_t$ we render $\hat I_t$, $\hat D_t$, and $\hat F_{t\rightarrow t_1}$ and supervise them with $L_\mathrm{rgb}$, $L_\mathrm{depth}$, and $L_\mathrm{track}^\mathrm{gaussian}$. Geometry regularizer $L_\mathrm{arap}$, $L_\mathrm{vel}$, and $L_\mathrm{acc}$ act on the scaffold nodes $v^{(m)}$. Two extra terms, $L_\mathrm{track}^\mathrm{scaffold}$ and $L_\mathrm{depth}^\mathrm{virtual}$, tether $v^{(m)}$ to accurately preserve object structure in motion and remove floaters, respectively. The final reconstruction result renders photorealistic views at any time and viewpoint.
}
\label{fig:method_overview}
\end{figure*}

\paragraph{Dynamic Novel View Synthesis.}
Recent years have seen rapid progress in dynamic scene reconstruction for novel-view synthesis. Many methods assume multi-view inputs with known calibration~\cite{kratimenos2023dynmf, wu2023four, attal2023hyperreel, bansal2020four, cao2023hexplane, fridovich2023kplanes, li2021neural, lin2023im4d, luiten2023dynamic, wang2022fourier, li2020neural, lombardi2019neural,
xu2024representing,
yan20244d,
lee2024fully,
zhu2024motiongs,
bae2024per,
sun20243dgstream,
duan20244d,
li2024spacetime,
shaw2023swings,
huang2024sc,
liang2025gaufre,
kratimenos2024dynmf,
wu20244d,
lin2024gaussian}, 
while others focus on the more practical but challenging monocular setting~\cite{lei2024mosca, wang2024shape, park2024splinegs, kwak2025modec, liu2024modgs, stearns2024dynamic, zhao2024dynomo, li2023fast, bui2023dyblurf, li2020neural, liang2023gaufre, zhao2023pseudo, zhang2023dynpoint, miao2024ctnerf, li2022layeredgs, wang2022fourier, you2023decoupling,
zhang2025motion,
das2023neural,
yoon2020novel, jeong2024rodygs}. 
Monocular reconstruction is particularly challenging due to limited parallax, occlusion, and motion blur, often leading to unstable or incomplete geometry.

Alongside this divide in input assumptions, approaches also differ in their scene representations. NeRF-based models~\cite{mildenhall2021nerf, li2020neural, shih2024modeling, li2023dynibar, gao2021dynamic, fang2022fast, jiang2022neuman, athar2022rignerf, du2021neural,
xian2021space, song2023nerfplayer} rely on implicit volumetric fields, but are slow to train and difficult to edit. Recent dynamic extensions of 3D Gaussian Splatting~\cite{kerbl20233d, lei2024mosca, wang2024shape, park2024splinegs, liu2024modgs, stearns2024dynamic, yang2024deformable, yang2023real} offer real-time rendering and more explicit control via point-based structures, making them especially suitable for dynamic content. Some recent methods also adopt a feedforward strategy, directly predicting novel views from monocular input without explicit 3D reconstruction~\cite{liang2024feed, wu2024cat4d}.

Similar to MoSca~\cite{lei2024mosca}, SplineGS~\cite{park2024splinegs}, and Shape of Motion~\cite{wang2024shape}, our method adopts an explicit deformation representation to model dynamic scenes. In contrast to these approaches, we further enhance the quality of priors obtained from foundation models and boost the quality of reconstruction. 
\paragraph{Camera Pose Estimation from Monocular Video.}
Most prior dynamic reconstruction pipelines rely on COLMAP for camera pose estimation. While effective in static environments, COLMAP often fails in dynamic or narrow-baseline videos due to foreground motion and small baseline. Recent methods such as Robust CVD~\cite{kopf2020robust}, CasualSAM~\cite{10.1007/978-3-031-19827-4_2}, and MegaSAM~\cite{li2024megasam} address this limitation by jointly optimizing depth and pose from monocular videos with the prior from mono-depth. Other approaches adopt feedforward networks that directly regress poses from video frames~\cite{zhang2024monst3r, wang2025cut3r, feng2025st4rtrack}, enabling faster inference but often requiring large-scale training data and suffering in low-texture regions.

For dynamic novel view synthesis, approaches like RoDynRF~\cite{liu2023robust}, SplineGS~\cite{park2024splinegs}, and MoSca~\cite{lei2024mosca} propose COLMAP-free pipelines by using motion masks to isolate static background regions, enabling geometry learning and camera estimation without external SfM tools. 

\paragraph{Geometry and motion prior.}
Recent advances in vision foundation models have led to high-quality predictions across a variety of image-based tasks. This progress has made them increasingly popular in dynamic reconstruction pipelines, where monocular input suffers from ambiguity in geometry and motion. Depth models~\cite{yang2024depth, wang2024moge, hu2024depthcrafter, piccinelli2024unidepth} provide geometric cues. MoGe~\cite{wang2024moge} predicts affine-invariant 3D point maps, and MegaSAM~\cite{li2024megasam} produces consistent video depth and pose from DepthAnything ~\cite{yang2024depth}; Point tracking methods ~\cite{harley2022particle, doersch2023tapir, karaev2023cotracker, xiao2024spatialtracker, doersch2022tap, doersch2024bootstap, tapip3d}
 estimate dense point trajectories across time. ~\cite{xiao2024spatialtracker, doersch2024bootstap, karaev2023cotracker} can track long-range pixels' trajectory; and segmentation models~\cite{kirillov2023segment, ravi2024sam2, yang2023track, huang2025segment}
 produce masks that help isolate foreground motion and preserve fine structural detail. 
 Recent work train learnable modules to extract dynamic objects~\cite{huang2025segment, goli2024romo, karazija2024learning}. In contrast, we directly combine video segmentation with geometric heuristics for a simpler yet effective solution.
\section{Method}
\label{sec:method}
We first review Motion Scaffold (MoSca)~\cite{lei2024mosca} in Sec.~\ref{sec:preliminaries}. Sec.~\ref{sec:overview} gives a high-level outline of our pipeline, and Secs.~\ref{sec:init}–\ref{sec:dyn_recon} describe the three stages in detail.
\subsection{Preliminary: Motion Scaffold}
\label{sec:preliminaries}


Motion Scaffold (MoSca)~\cite{lei2024mosca} represents a dynamic scene with (i) a collection of dynamic 3-D Gaussians that translate and rotate over time to represent geometry and appearance, and (ii) a sparse set of 3-D \emph{motion scaffold nodes} $v^{(m)} \in \mathcal V$ whose time-varying rigid transforms $Q^{(m)}_t=[\mathbf R^{(m)}_t,\mathbf t^{(m)}_t]$ encode the motion. These nodes $v^{(m)}$ are connected as a graph $\mathcal G=(\mathcal V,\mathcal E)$ by $K$-nearest-neighbor links; dual-quaternion blending over $\mathcal G$ yields a smooth deformation field that drives every dynamic Gaussian.

\subsection{Problem Statement}
We tackle \textbf{dynamic view synthesis} from a single handheld RGB video of everyday scenes.  
While prior work concentrates on expressive scene representations, we show that \emph{higher-quality priors} obviously boosts final reconstruction quality.


\subsection{Method Overview}
\label{sec:overview}
Figure~\ref{fig:method_overview} sketches our three-stage pipeline.  Given $T$ input frames $I_{t=1}^{T}$, we reconstruct a scene of dynamic Gaussians whose rotation and translation are governed by a motion-scaffold graph $\mathcal G=(\mathcal V,\mathcal E)$ in three stages, (1) \textbf{Initialization} (Sec.~\ref{sec:init}), (2) \textbf{Lifting to 3-D} (Sec.~\ref{sec:lift}), and (3) \textbf{Dynamic-scene reconstruction} (Sec.~\ref{sec:dyn_recon}).


\subsection{Initialization}
\label{sec:init}

In this stage (see Fig.~\ref{fig:method_overview}(a)), we prepare the following ingredients that enables the subsequent 3-D/4-D reconstruction: \textbf{(1)} video segmentation $\{S_t^{(l)}\}_{t=1,l=1}^{T,L}$~\cite{chan2022efficient,ravi2024sam2,AutoSeg_SAM2}, where $L$ is number of segments, \textbf{(2)}  optical flow $F_{t\rightarrow t'}$~\cite{teed2020raft}, \textbf{(3)}  epipolar-error masks (EPI error masks) $E_{t=1}^T$ ~\cite{liu2023robust}, \textbf{(4)}  dynamic-object masks $\{M_t^{(o)}\}_{t=1,o=1}^{T,O}$, where $O$ is number of dynamic objects, \textbf{(5)} single-image (mono) depth maps $\tilde{D}_{t=1}^T$~\cite{wang2024moge}, \textbf{(6)} camera poses $\Pi_{t=1}^T$, 
\textbf{(7)} consistent video depth maps $D_{t=1}^T$~\cite{li2024megasam}, and \textbf{(8)} 2-D point tracks $\{u_t^{(n)}\}_{t=1,n=1}^{T,N}$, where $N$ is number of tracks.

EPI error masks $E_{t=1}^T$ reveal dynamic surfaces but cover only the moving parts. We enlarge these regions with video-segmentation cues $S_t^{(l)}$ to obtain robust object-level masks $M_t^{(o)}$ that cover each dynamic object in its entirety. The following subsections detail how these masks are constructed and how they improve both the video depths $D_{t=1}^{T}$ and the 2-D point tracks $u_t^{(n)}$ on the dynamic objects.



\begin{figure}[t]  
\centering

\begin{subfigure}[t]{0.33\linewidth}
  \includegraphics[width=\linewidth]{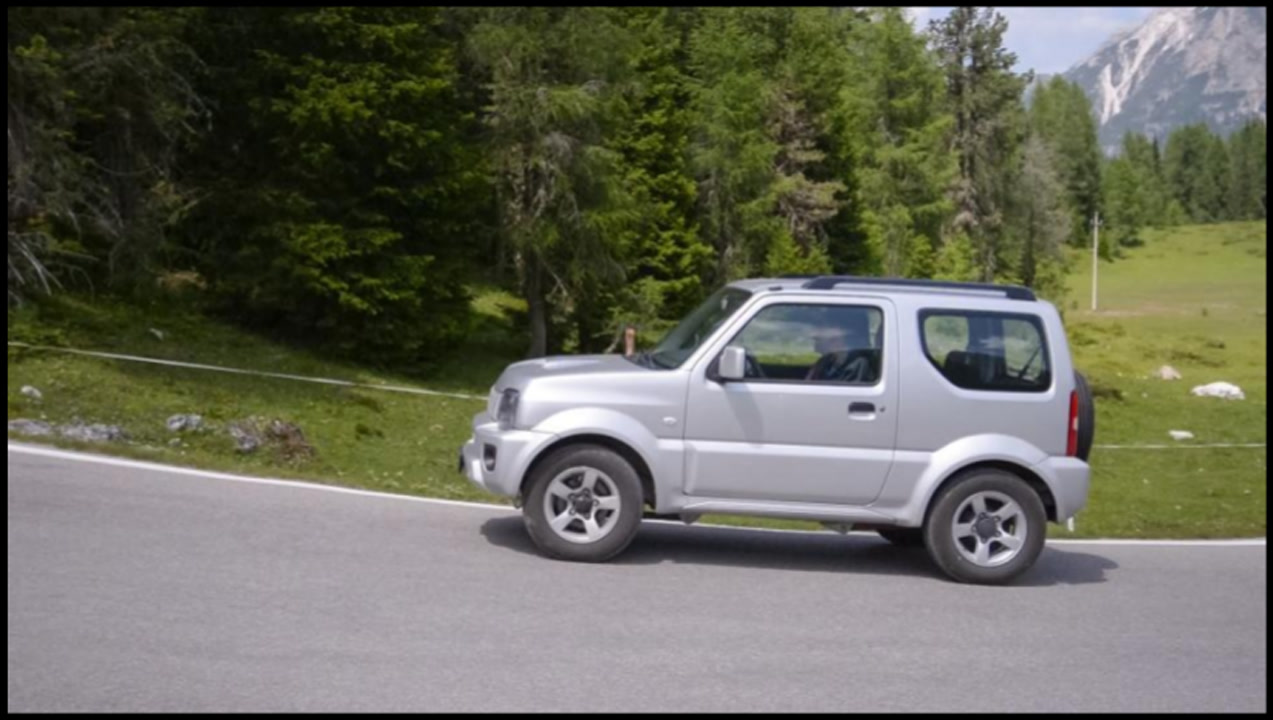}
  \caption{Image}
\end{subfigure}\hfill
\begin{subfigure}[t]{0.33\linewidth}
  \includegraphics[width=\linewidth]{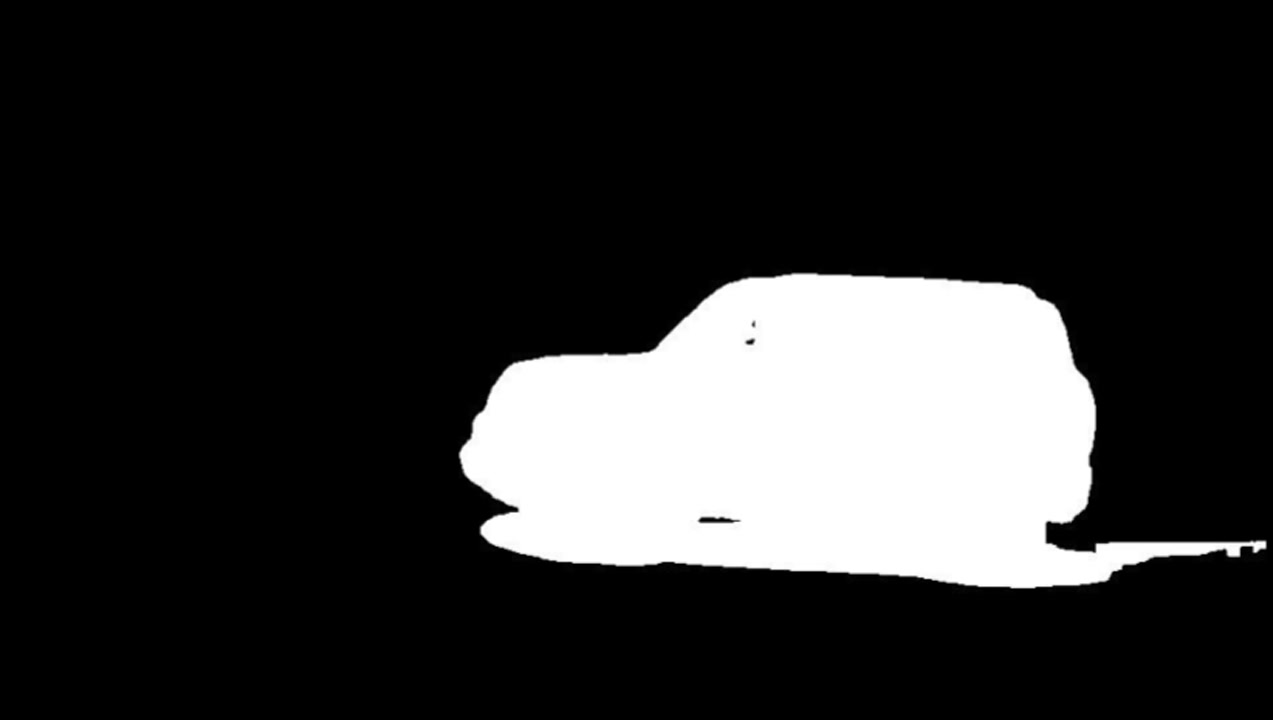}
  \caption{EPI error mask}
\end{subfigure}\hfill
\begin{subfigure}[t]{0.33\linewidth}
  \includegraphics[width=\linewidth]{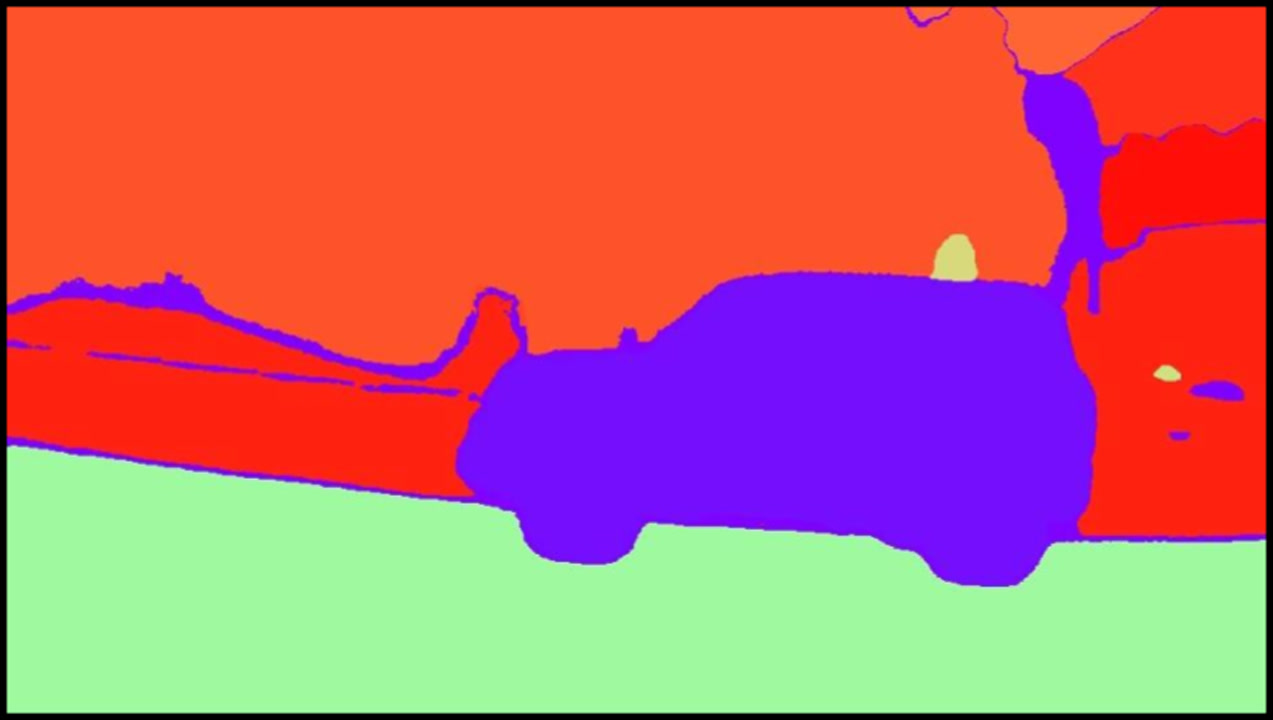}
  \caption{Video Segmentation}
\end{subfigure}


\begin{subfigure}[t]{0.33\linewidth}
  \includegraphics[width=\linewidth]{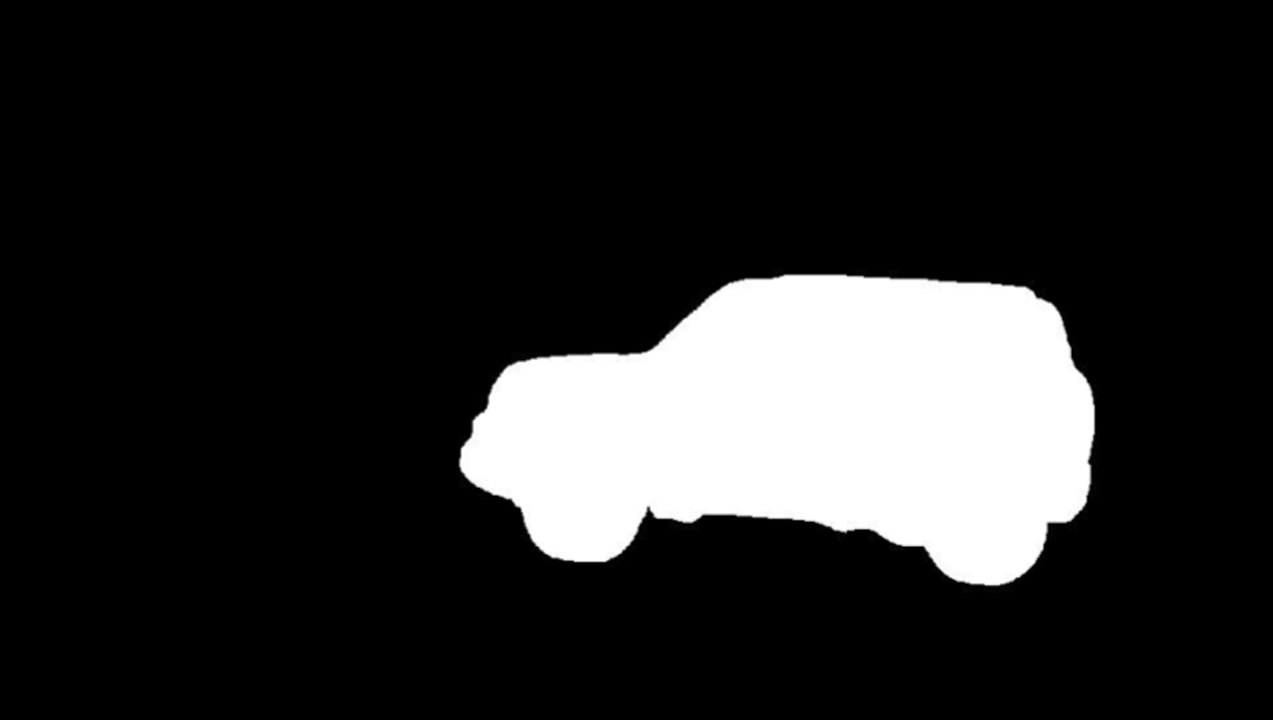}
  \caption{Car Segment}
\end{subfigure}\hfill
\begin{subfigure}[t]{0.33\linewidth}
  \includegraphics[width=\linewidth]{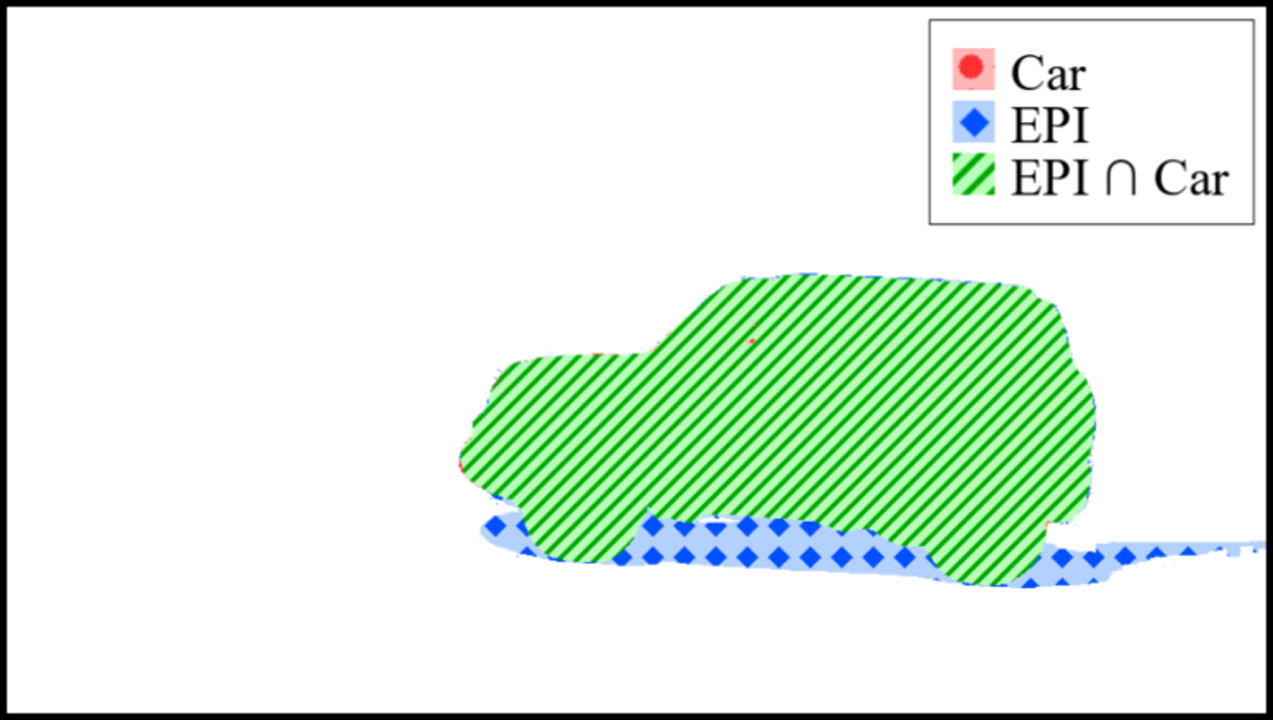}
  \caption{Car Segment \& EPI Error Mask}
\end{subfigure}\hfill
\begin{subfigure}[t]{0.33\linewidth}
  \includegraphics[width=\linewidth]{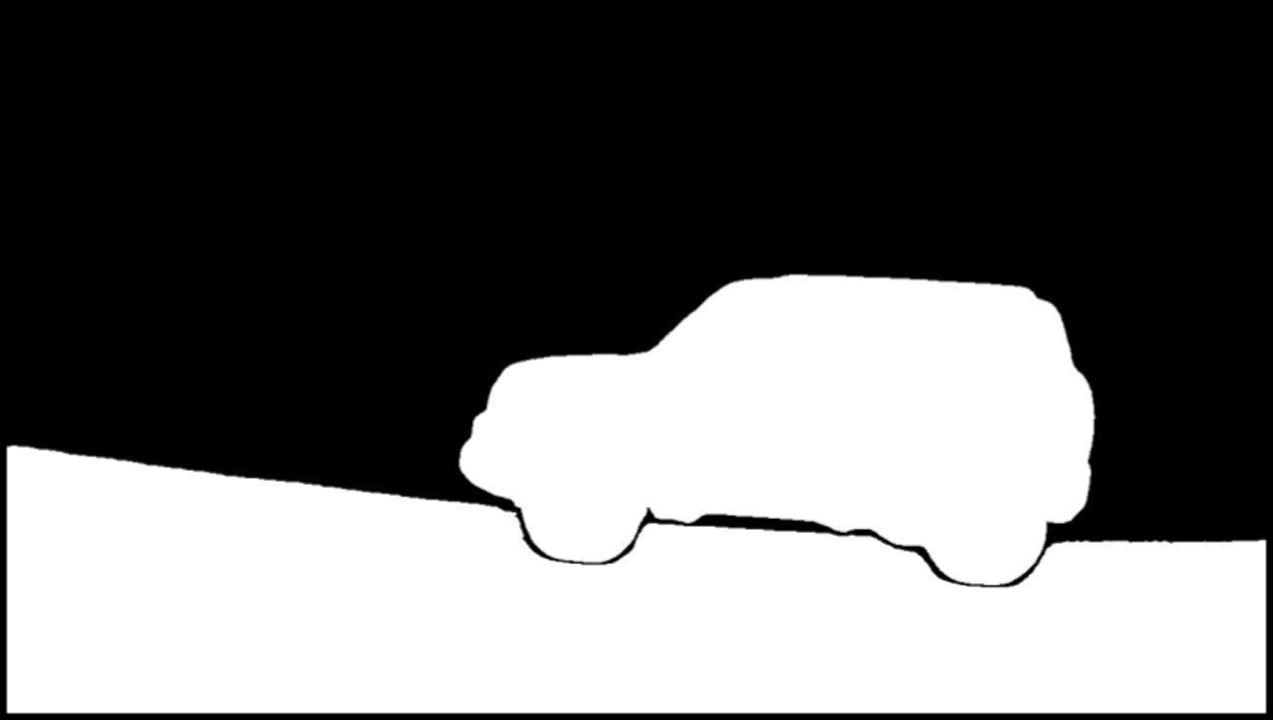}
  \caption{Dynamic mask pass 1$^{\text{st}}$ test}
\end{subfigure}


\begin{subfigure}[t]{0.33\linewidth}
  \includegraphics[width=\linewidth]{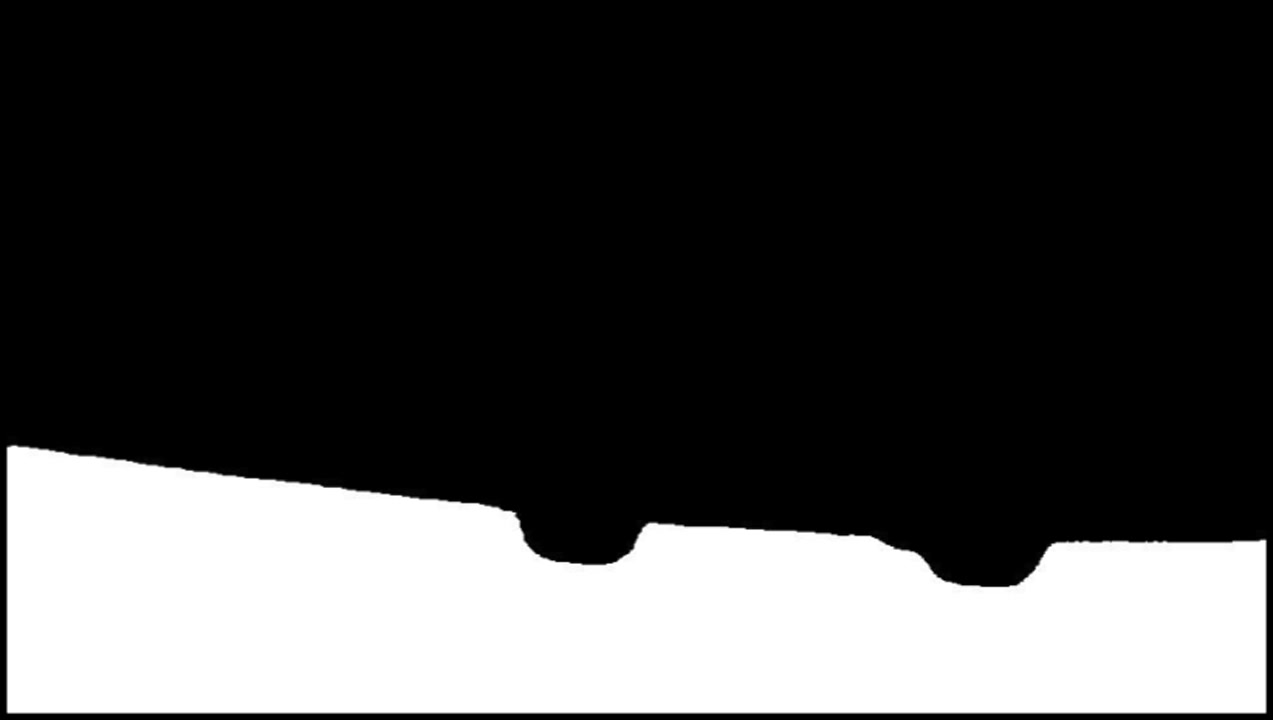}
  \caption{Road Segment}
\end{subfigure}\hfill
\begin{subfigure}[t]{0.33\linewidth}
  \includegraphics[width=\linewidth]{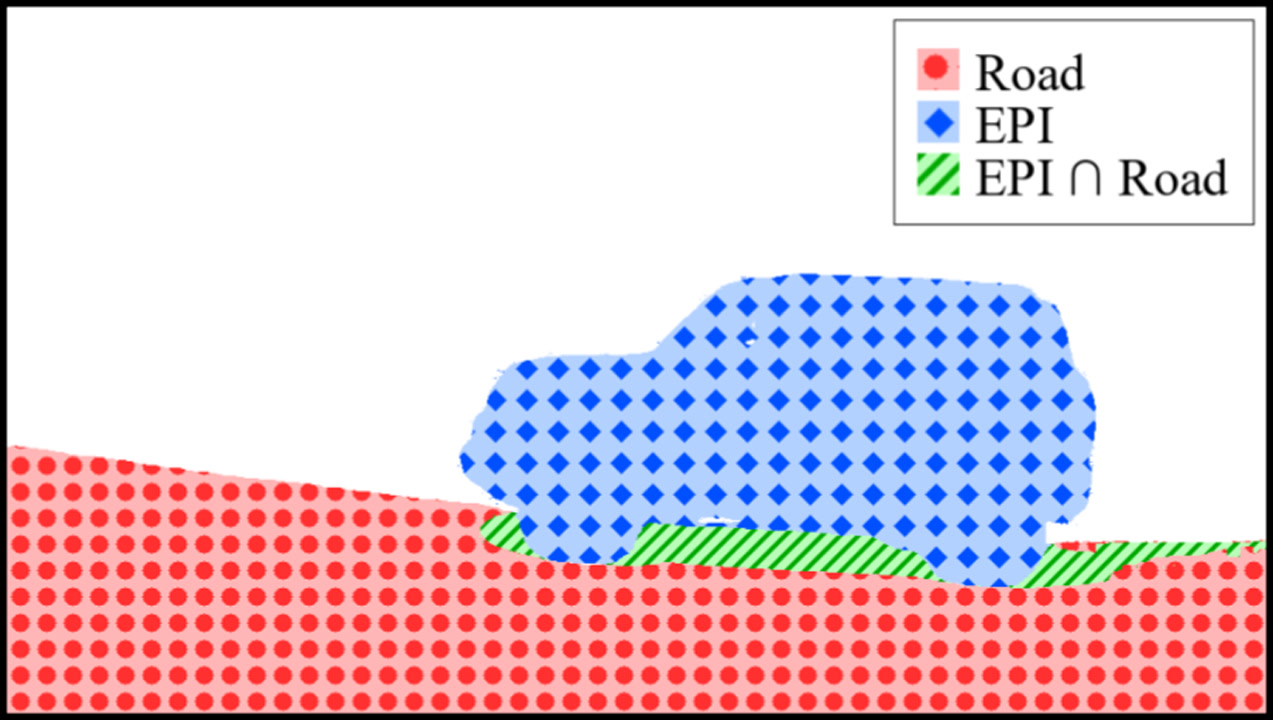}
  \caption{Road Segment \& EPI Error Mask}
\end{subfigure}\hfill
\begin{subfigure}[t]{0.33\linewidth}
  \includegraphics[width=\linewidth]{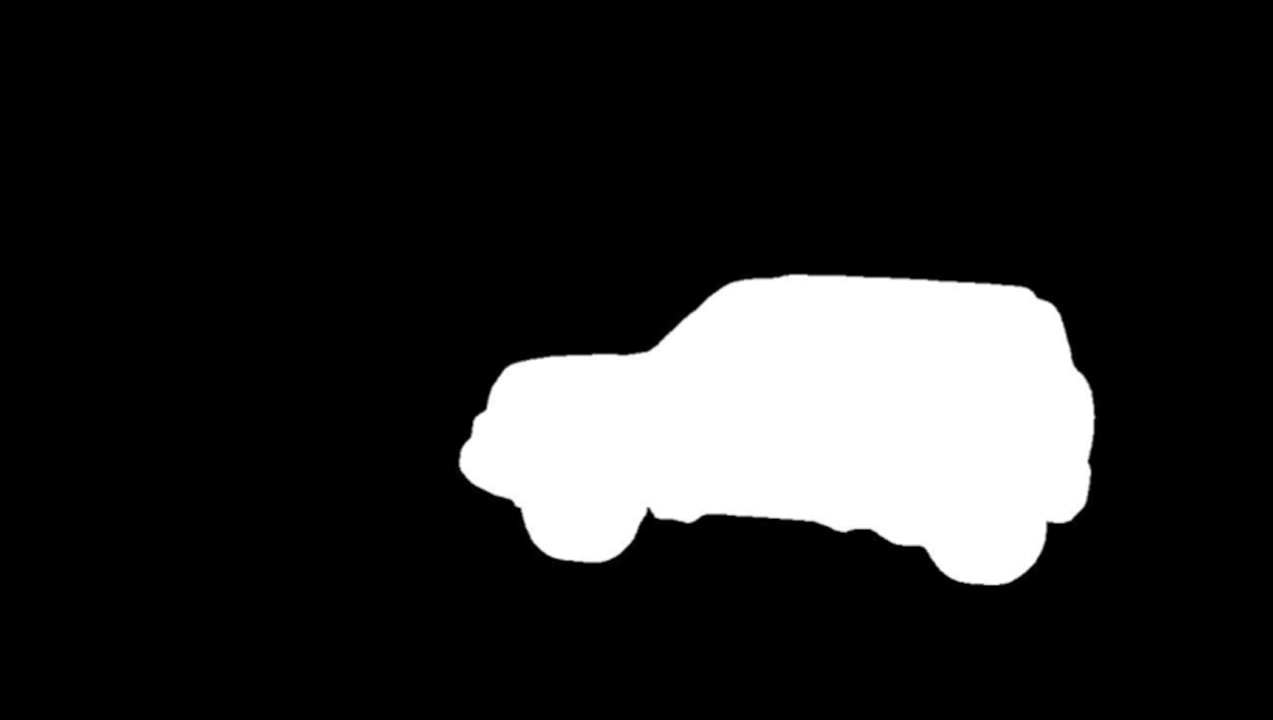}
  \caption{Dynamic mask pass 1$^{\text{st}}$ \& 2$^{\text{nd}}$ test}
\end{subfigure}

\caption{
\textbf{Dynamic Object Mask Selection.} We intersect the EPI error mask $E_{t=1}^T$ (b) with each video segment $S^{(l)}$ (c).
The intersections for the car and road segments appear in (e) \& (h).
Applying the two-pass test from Sec.~\ref{sec:init} removes road shadows as outliers, leaving only the car as the dynamic-object mask $M_t^{(o)}$ (i).
} 
\label{fig:dynamic_mask}
\end{figure}

\paragraph{Dynamic Object Mask Selection.}
To identify the \emph{salient} dynamic objects we compute the intersection between $S_t^{(l)}$ and $E_{t=1}^T$ (Fig.~\ref{fig:dynamic_mask}).
A segment is kept if it covers at least
$\tau_\mathrm{salient}=0.05$ of the total moving surface:  
\[
\frac{|\,S^{(l)}\cap E_{t=1}^{T}|}{|E_{t=1}^{T}|}\;\ge\;\tau_\mathrm{salient}.
\]
Static segments occasionally pass this test because of appearance change (e.g. moving shadows);
we discard those whose \emph{own} motion area is small
($\tau_\mathrm{appearance}=0.2$):
\[
\frac{|\,S^{(l)}\cap E_{t=1}^{T}|}{|S^{(l)}|}
\;\ge\;\tau_\mathrm{appearance}.
\]
The filtered results become the per-frame dynamic object masks
$\{M_t^{(k)}\}$.


\captionsetup[figure]{name=Fig.,font=small}
\captionsetup[subfigure]{font=footnotesize,justification=centering,skip=1pt}
\captionsetup[figure]{aboveskip=2pt,belowskip=0pt}

\begin{figure}[h]  
\centering

\begin{subfigure}[t]{0.33\linewidth}\centering
  \includegraphics[width=\linewidth]{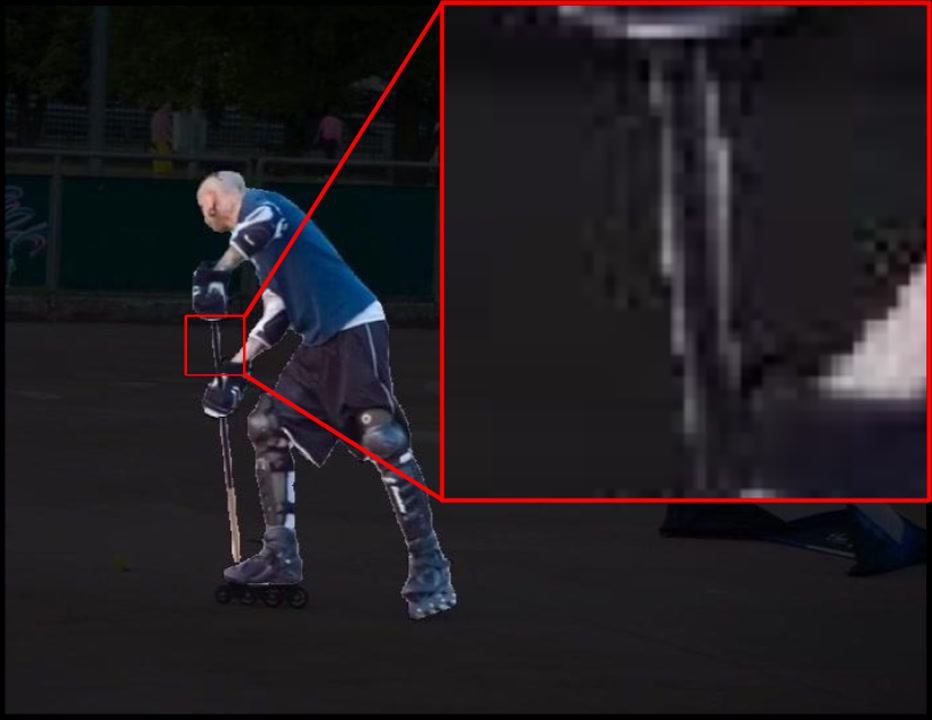}
  \caption{Image \& Dynamic mask}
\end{subfigure}\hfill
\begin{subfigure}[t]{0.33\linewidth}\centering
  \includegraphics[width=\linewidth]{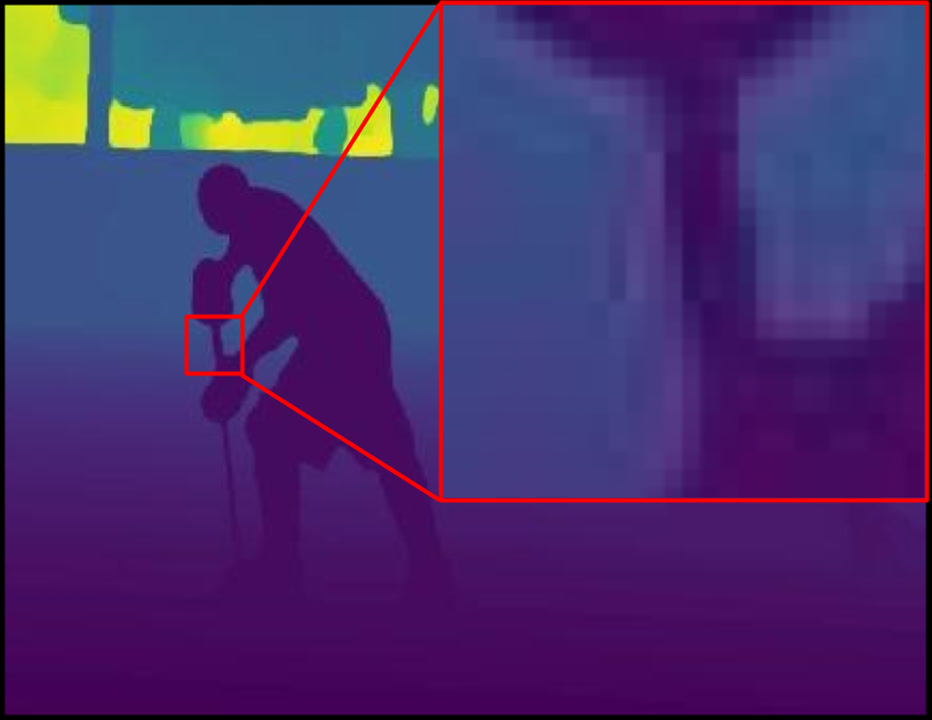}
  \caption{MoGe Mono-Depth $\tilde{D}$}
\end{subfigure}\hfill
\begin{subfigure}[t]{0.33\linewidth}\centering
  \includegraphics[width=\linewidth]{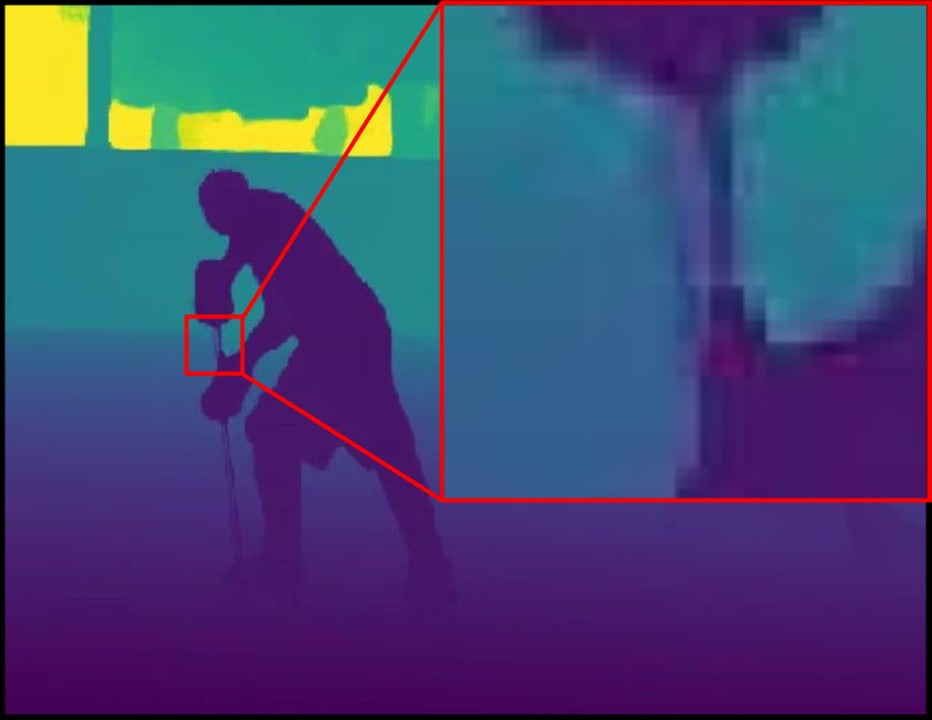}
  \caption{Consistent Video Depth $D$ w/o $L^{\text{object}}_{\text{depth}}$}
\end{subfigure}


\begin{subfigure}[t]{0.33\linewidth}\centering
  \includegraphics[width=\linewidth]{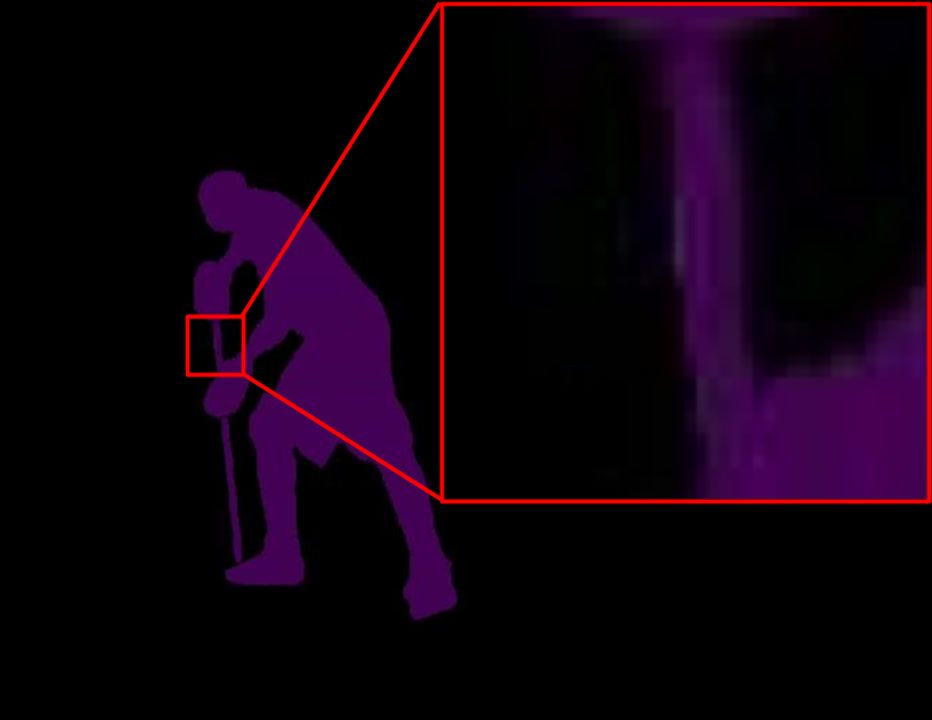}
  \caption{Aligned object depth}
\end{subfigure}\hfill
\begin{subfigure}[t]{0.33\linewidth}\centering
  \includegraphics[width=\linewidth]{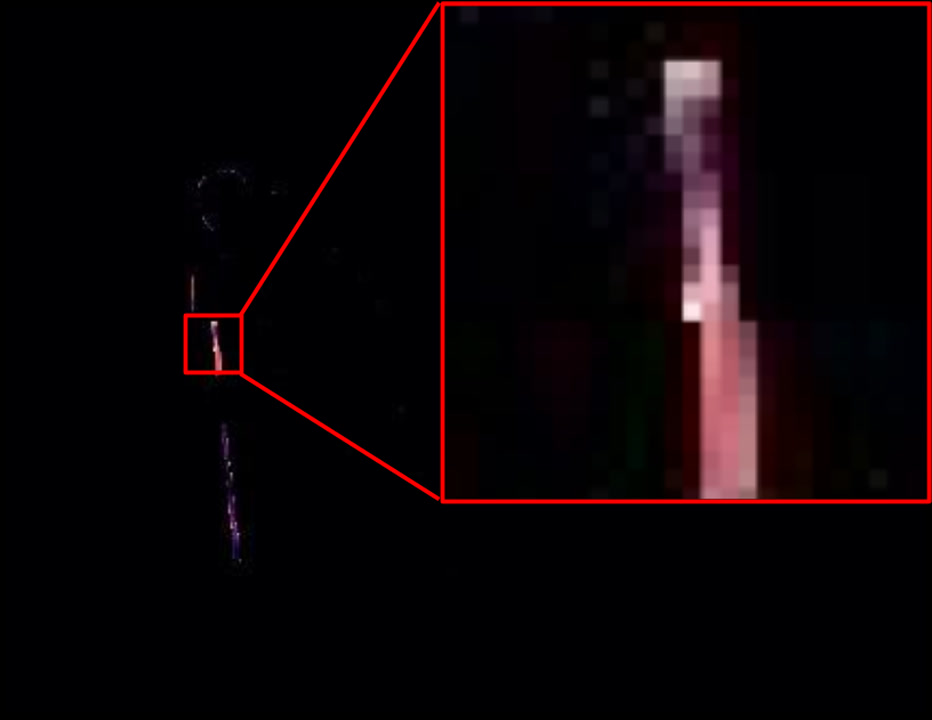}
  \caption{$L^{\text{object}}_{\text{depth}}$}
\end{subfigure}\hfill
\begin{subfigure}[t]{0.33\linewidth}\centering
  \includegraphics[width=\linewidth]{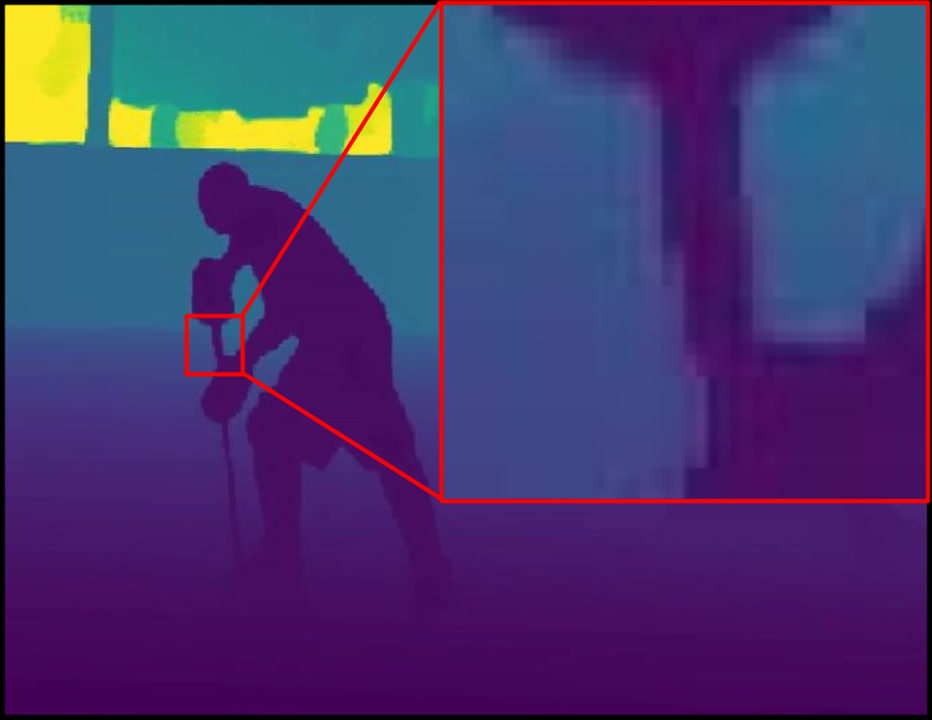}
  \caption{Consistent Video Depth $D$ w/ $L^{\text{object}}_{\text{depth}}$}
\end{subfigure}

\caption{
\textbf{Depth Refinement.} Mega-SAM depths lack fine detail: in (c) the hockey stick is swallowed by the background. We restore this detail using the high-quality MoGe depth in (b). We extract the object-only depth $\tilde D_t^{(o)}$ for dynamic mask from (b) and align it to the consistent depth $D_t$, producing (d). Then, the object-depth loss $L_\mathrm{depth}^\mathrm{object}$ between (c) and (d) (visualized in (e)) is added it to Mega-SAM’s optimization. The refinement produces depth maps that cleanly preserve the hockey stick, leading to higher-quality reconstruction (see (f) and Fig.~~\ref{fig:qual_depth_refine_recon}; details in Sec.~\ref{sec:init}).
}
\label{fig:refine_depth}
\end{figure}

\paragraph{Depth Refinement.} Although Mega-SAM ~\cite{li2024megasam} provides temporally consistent video depths $D_{t=1}^{T}$, its estimates are oversmoothed on thin, fast-moving parts of dynamic objects (Fig.~\ref{fig:refine_depth}).
To restore the fine detail, we (1) replace Mega-SAM's initial mono-depths $\tilde{D}_{t=1}^{T}$ ~\cite{yang2024depth} with ~\cite{wang2024moge} and (2) add an \emph{object-depth loss} $L_\mathrm{depth}^\mathrm{object}$ to its consistency optimization.

Specifically, for each dynamic-object mask $M_t^{(o)}$ we
(i) crop the mono-depth to obtain an object-only map
$\tilde D_t^{(o)} = M_t^{(o)}\!\odot\!\tilde D_t$;
(ii) align this map to the current video depth $D_t$ via a
mask-restricted scale–shift fit~\cite{wang2024moge} $\alpha_t^{(o)}\tilde D_t^{(o)}+\beta_t^{(o)}$; and
(iii) add $L_\mathrm{depth}^\mathrm{object}$ to Mega-SAM's consistency optimization stage 
\[
L_\mathrm{depth}^\mathrm{object}
=
\frac{1}{T\,|\Omega|}
\sum_{o=1}^{O}
\sum_{t=0}^{T}
\sum_{p\in\Omega}
      M_t^{(o)}(p)\,
      \bigl|
        D_t(p)\;-\;
        \bigl(\alpha_t^{(o)}\,\tilde D_t(p)+\beta_t^{(o)}\bigr)
      \bigr|,
\]
where $\Omega$ denotes domain of image $I$.

Because the loss is confined to object pixels, it sharpens thin structures while leaving the global scale and
overall temporal consistency of $D_{t=1}^{T}$ intact. As a result, fine details are enriched and thin, moving objects are reconstructed more faithfully (see Fig.\ref{fig:refine_depth} and Fig.\ref{fig:qual_depth_refine_recon}).

\begin{figure}[t]
\centering

\begin{subfigure}[t]{0.33\linewidth}\centering
  \includegraphics[width=\linewidth]{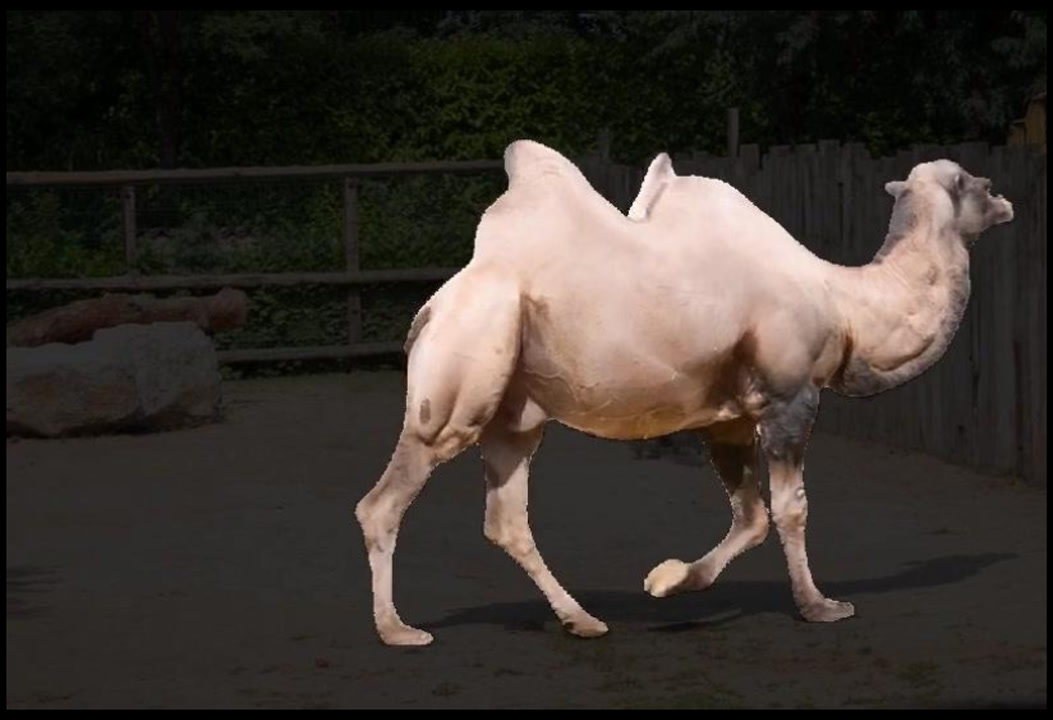}
  \caption{Image \& Dynamic Mask}
\end{subfigure}\hfill
\begin{subfigure}[t]{0.33\linewidth}\centering
  \includegraphics[width=\linewidth]{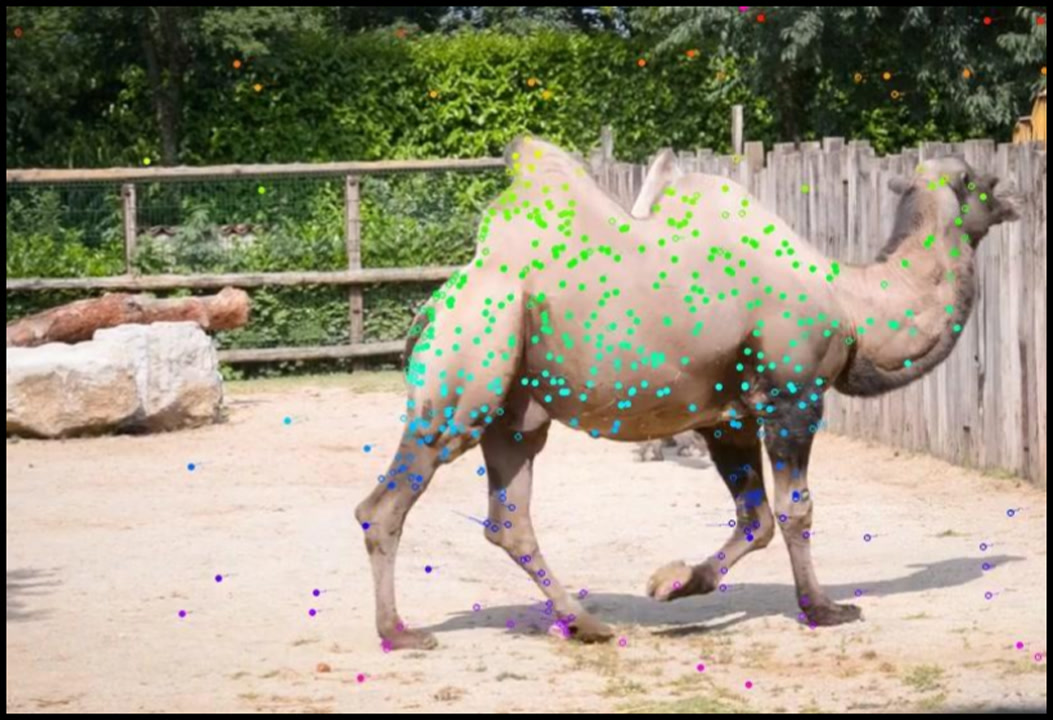}
  \caption{Uniform Sampling from EPI Error Mask}
\end{subfigure}\hfill
\begin{subfigure}[t]{0.33\linewidth}\centering
  \includegraphics[width=\linewidth]{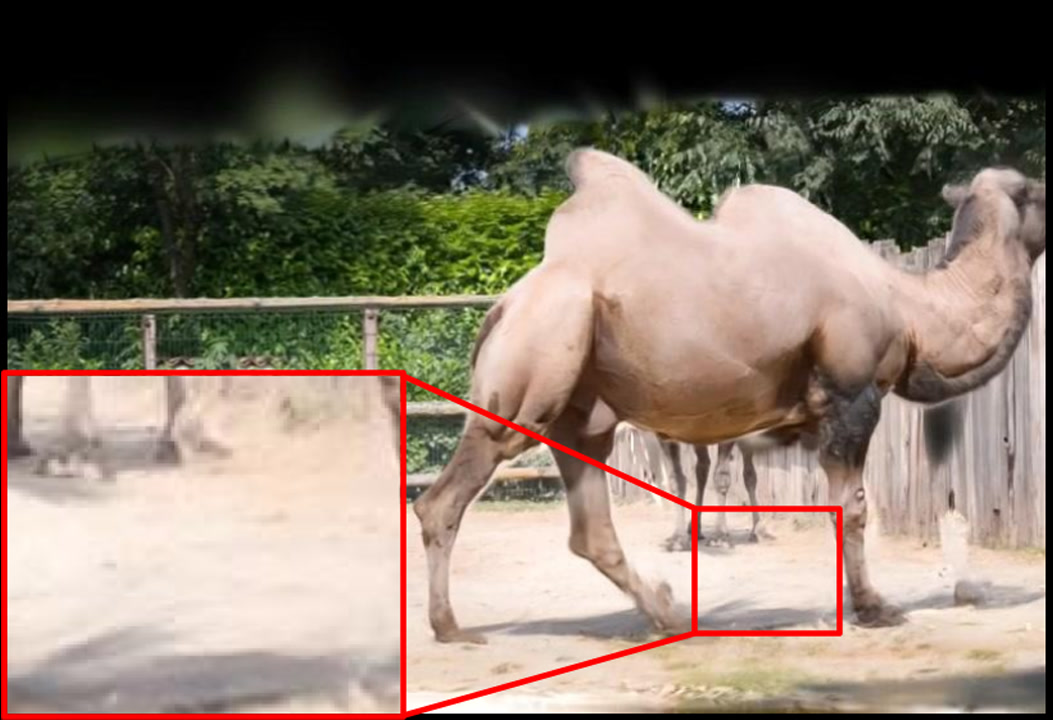}
  \caption{Novel View Synthesis w/o Skeleton Sampling}
\end{subfigure}


\begin{subfigure}[t]{0.33\linewidth}\centering
  \includegraphics[width=\linewidth]{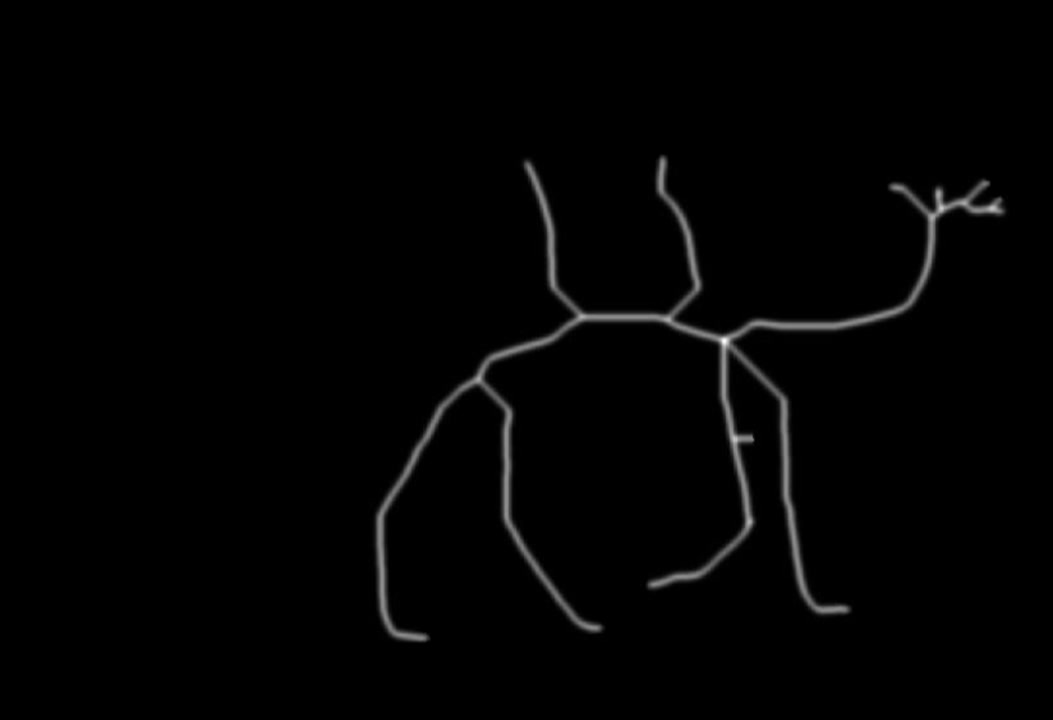}
  \caption{Skeleton of Dynamic Mask}
\end{subfigure}\hfill
\begin{subfigure}[t]{0.33\linewidth}\centering
  \includegraphics[width=\linewidth]{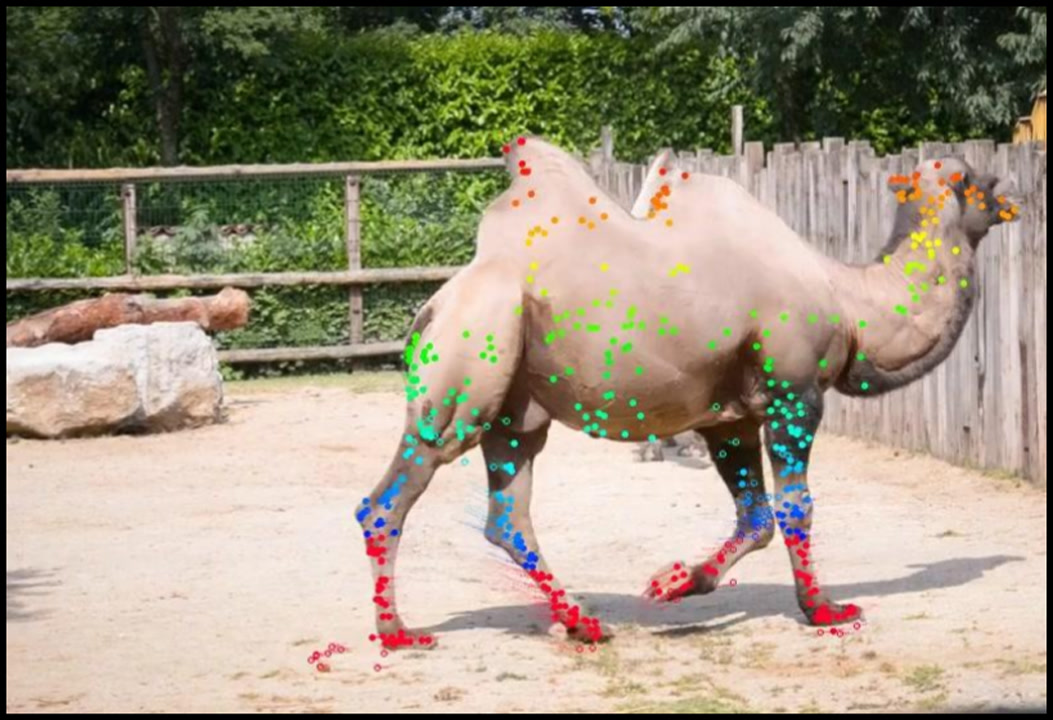}
  \caption{Skeleton Sampling from Dynamic Mask}
\end{subfigure}\hfill
\begin{subfigure}[t]{0.33\linewidth}\centering
  \includegraphics[width=\linewidth]{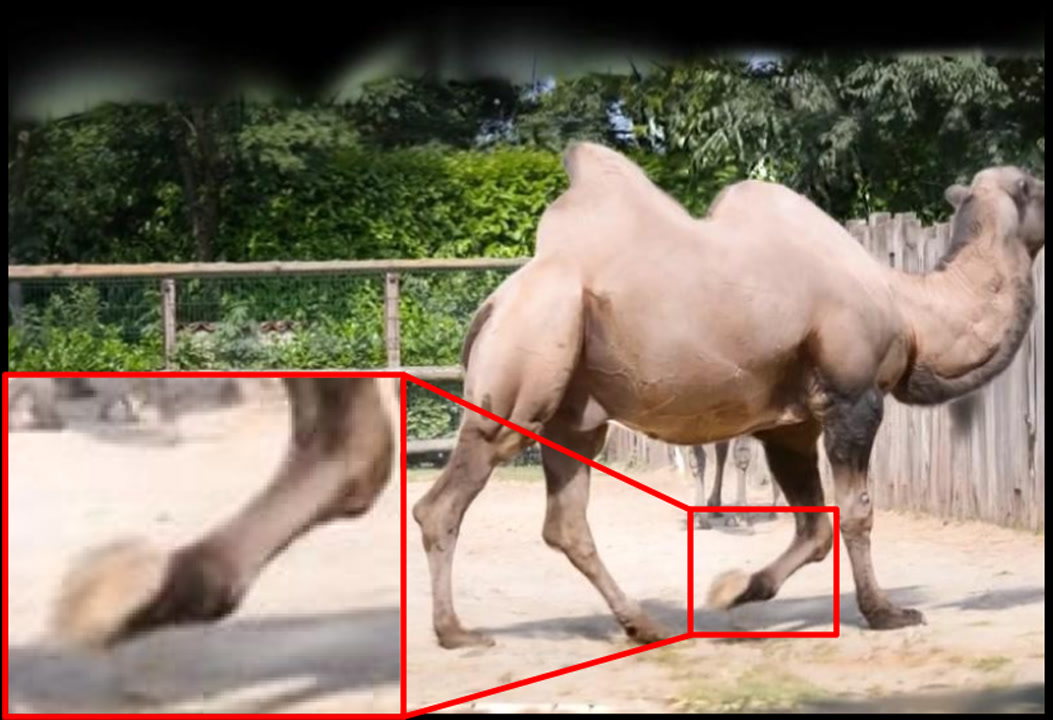}
  \caption{Novel View Synthesis w/ Skeleton Sampling}
\end{subfigure}

\caption{\textbf{Skeleton Sampling.} 2-D tracks seed object motion.
MoSca draws them uniformly inside EPI error masks (b), so thin parts are
undersampled and disappear in the result (c). We instead extract a skeleton
from each dynamic mask $M_t^{(o)}$ (d) and add extra samples in a narrow band
around it (e). These skeleton points, only $\tfrac{1}{6}$ of the uniform set,
provide dense coverage of limbs, sharply resolving them in the reconstruction
(f). Detail is in Sec.~\ref{sec:init}.}
\label{fig:skeleton_sampling} 
\end{figure}

\paragraph{Mask-guided point tracker.}
Uniformly sampling tracks from $E_{t=1}^{T}$~\cite{lei2024mosca} biases toward large, rigid parts (e.g.\ torsos) and under-samples thin, high-motion regions (e.g.\ limbs). We therefore adopt \emph{skeleton-sample}: for each object, we extract its 2-D medial-axis skeleton from the object mask $M_t^{(o)}$, dilate it by 5 pixels, weight pixels by inverse distance to the mask boundary, and draw an additional $\tfrac{1}{6}$ of tracks from this distribution. During trajectory estimation, frames are resized so that the longer side is 512 pixels, ensuring all trajectory-related heuristics operate at a consistent resolution. Compared with uniform sampling, this strategy yields noticeably better quality in thin regions (see Fig.\ref{fig:skeleton_sampling} and Fig.~\ref{fig:qual_skeleton}).

\begin{figure}[h]
\centering

\begin{subfigure}[t]{0.33\linewidth}\centering
  \includegraphics[width=\linewidth]{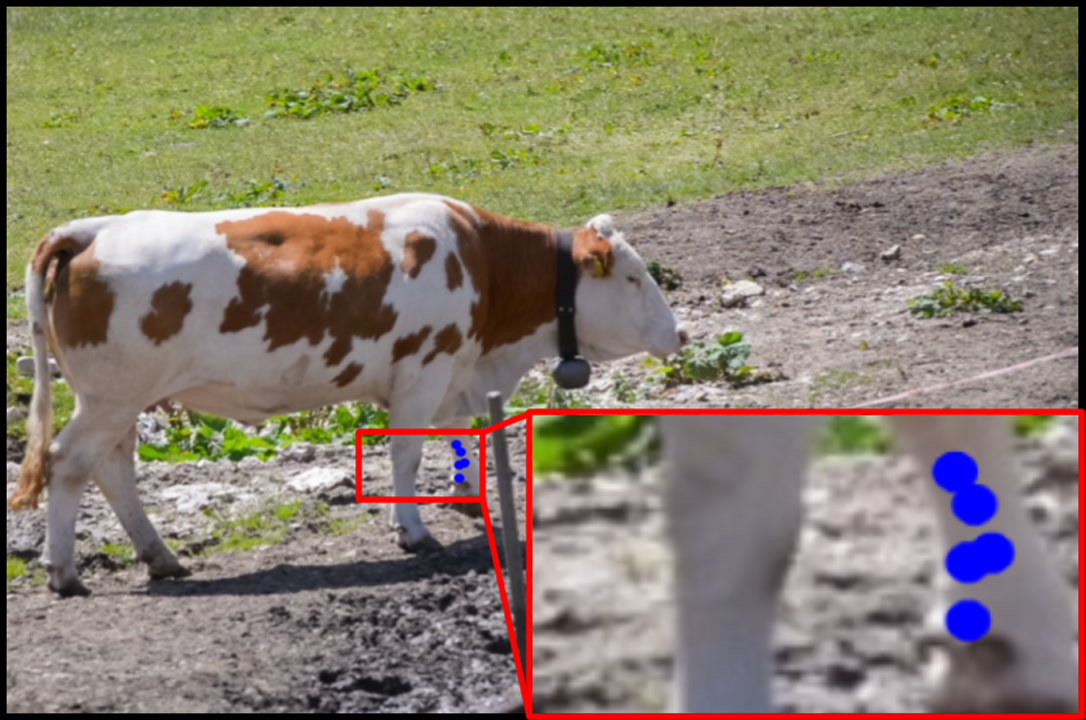}
  \caption{Sample Point at $t_a$}
\end{subfigure}\hfill
\begin{subfigure}[t]{0.33\linewidth}\centering
  \includegraphics[width=\linewidth]{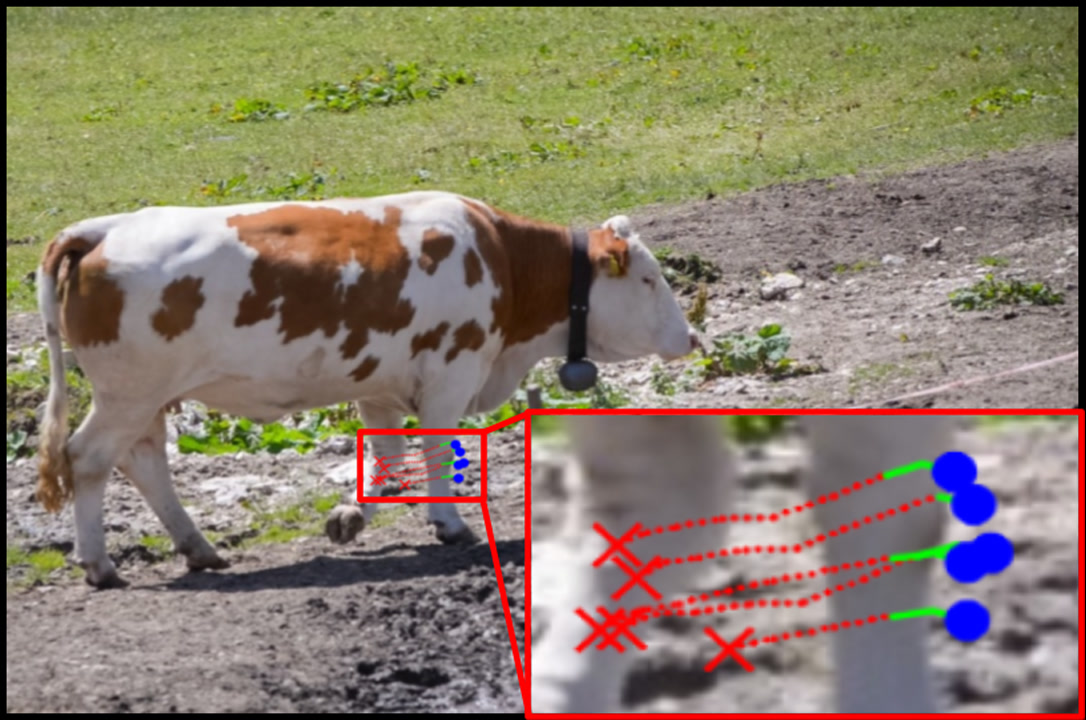}
  \caption{Track $t_a \rightarrow t_b$}
\end{subfigure}\hfill
\begin{subfigure}[t]{0.33\linewidth}\centering
  \includegraphics[width=\linewidth]{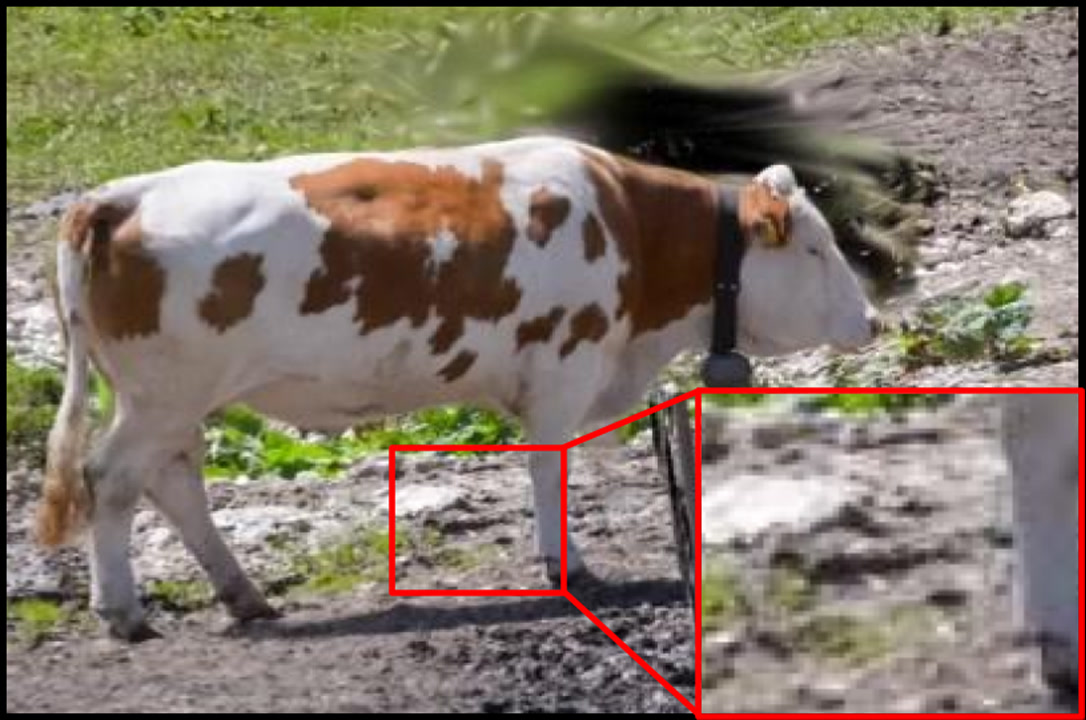}
  \caption{Novel View Synthesis at $t_b$ w/o Re-identification}
\end{subfigure}


\begin{subfigure}[t]{0.33\linewidth}\centering
  \includegraphics[width=\linewidth]{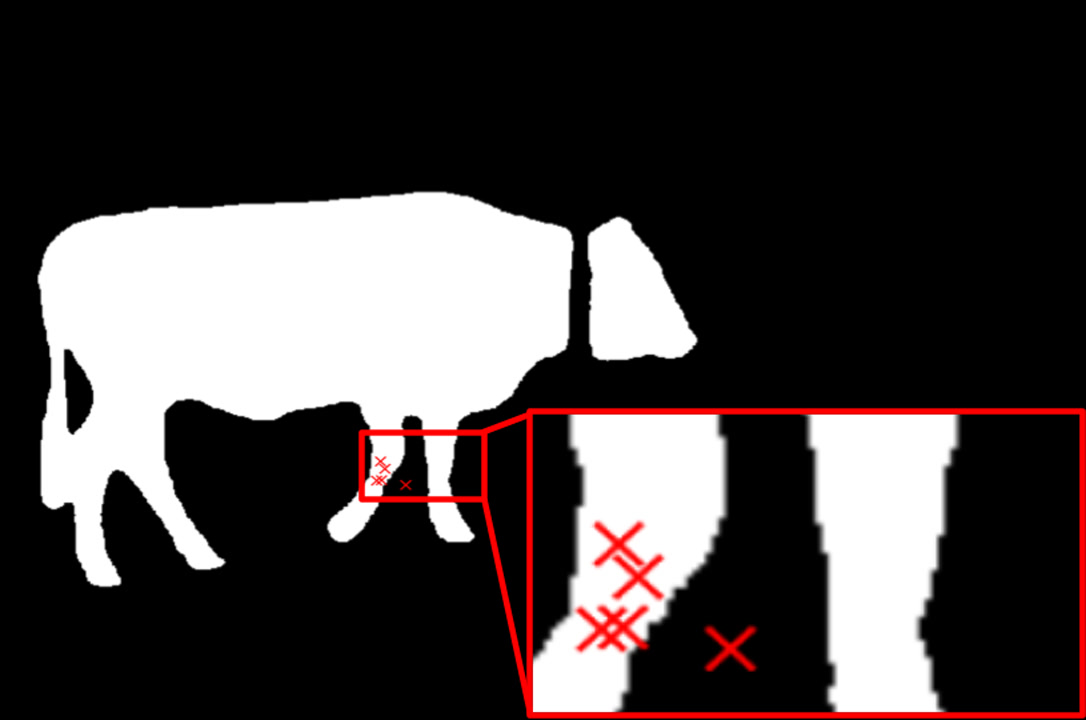}
  \caption{Find Tracks in Dynamic Mask at $t_b$}
\end{subfigure}\hfill
\begin{subfigure}[t]{0.33\linewidth}\centering
  \includegraphics[width=\linewidth]{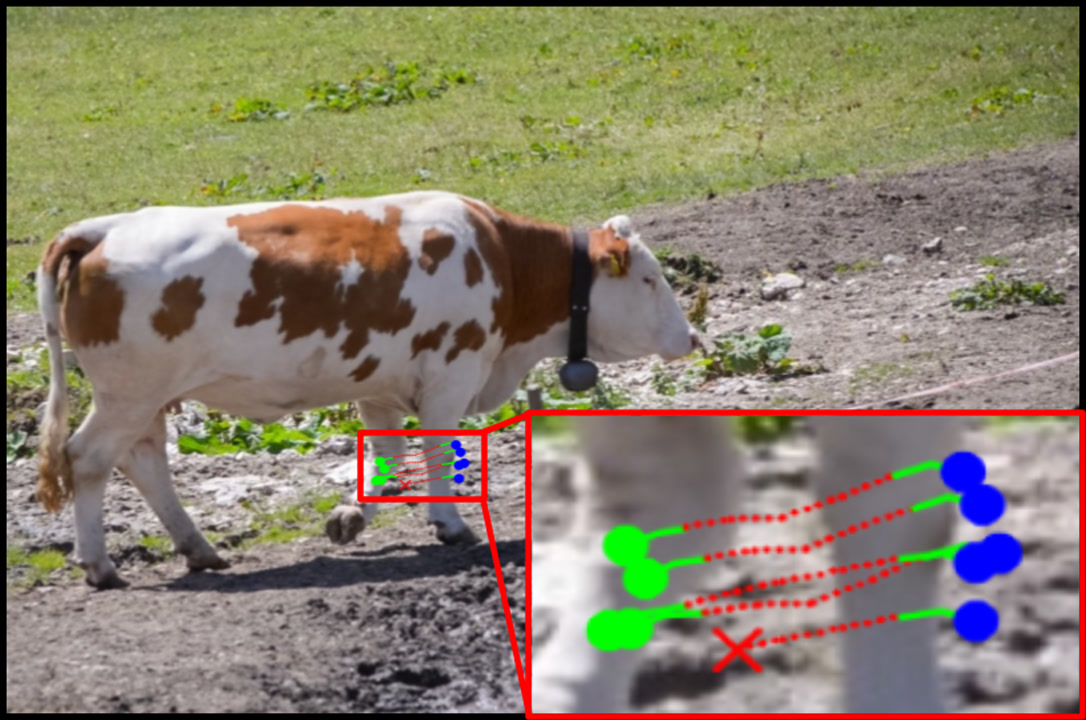}
  \caption{Re-identify Tracks at $t_b$}
\end{subfigure}\hfill
\begin{subfigure}[t]{0.33\linewidth}\centering
  \includegraphics[width=\linewidth]{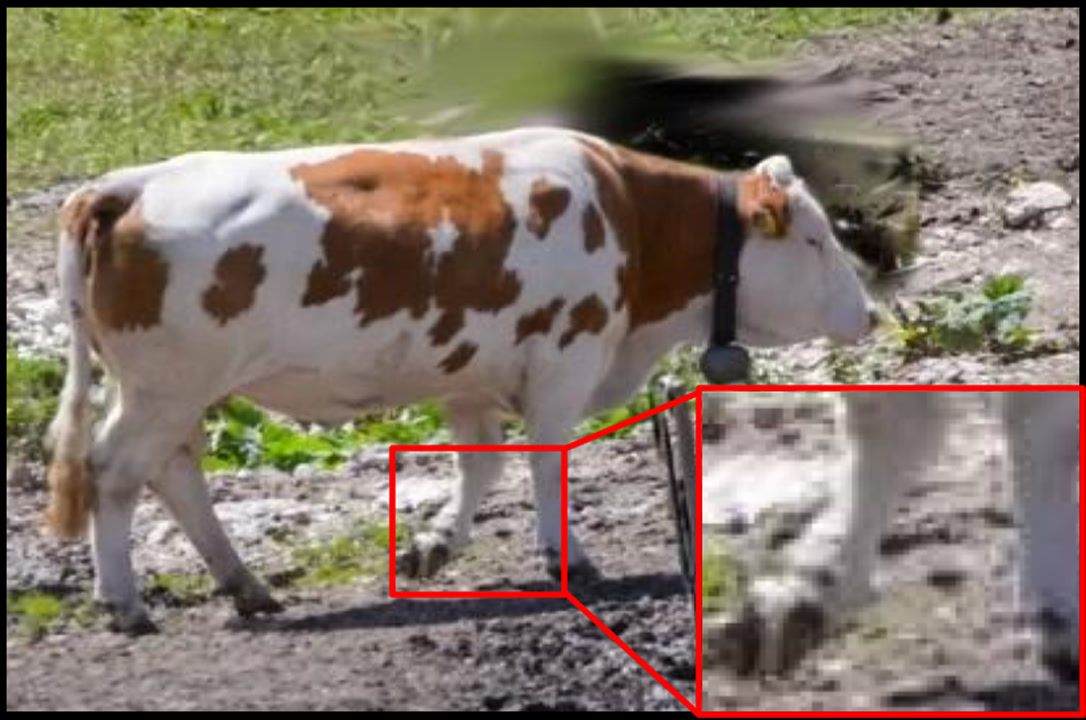}
  \caption{Novel View Synthesis at $t_b$ w/ Re-identification}
\end{subfigure}

\caption{\textbf{Track Re-identification.}
Points sampled at $t_a$ are tracked toward $t_b$ (blue dots, (a)).
After an occlusion, a re-emerged point is not re-identified (red dashed
path and cross, (b)), leaving a gap in the reconstruction (c).
We mark a track as \emph{visible} whenever its position lies inside the
same dynamic mask $M_t^{(o)}$ where it originated.
This restores the missing trajectory (e) and completes the
reconstruction (f). Detail is in Sec.~\ref{sec:init}.}
\label{fig:track_reidentification}
\end{figure}

2-D trackers often lose a point when they re-emerge from occlusion,
leaving the track labelled \emph{occluded} and creating holes in the
reconstruction (Fig.~\ref{fig:track_reidentification}).  
Points that remain on the same object usually move coherently with its surface,
so we treat the object masks $M_t^{(o)}$ as a re-identification
oracle: if a track originates in $M^{(o)}$ and its current position is
still inside that mask at frame $t$, we relabel it \emph{visible}.  
The recovered tracks $u_t^{(n)}$ re-enter optimization, supply the
missing supervision, and restore the surrounding geometry
(Fig.~\ref{fig:track_reidentification}).

\begin{figure}[h]
\centering

\begin{subfigure}[t]{0.495\linewidth}\centering
  \includegraphics[width=\linewidth]{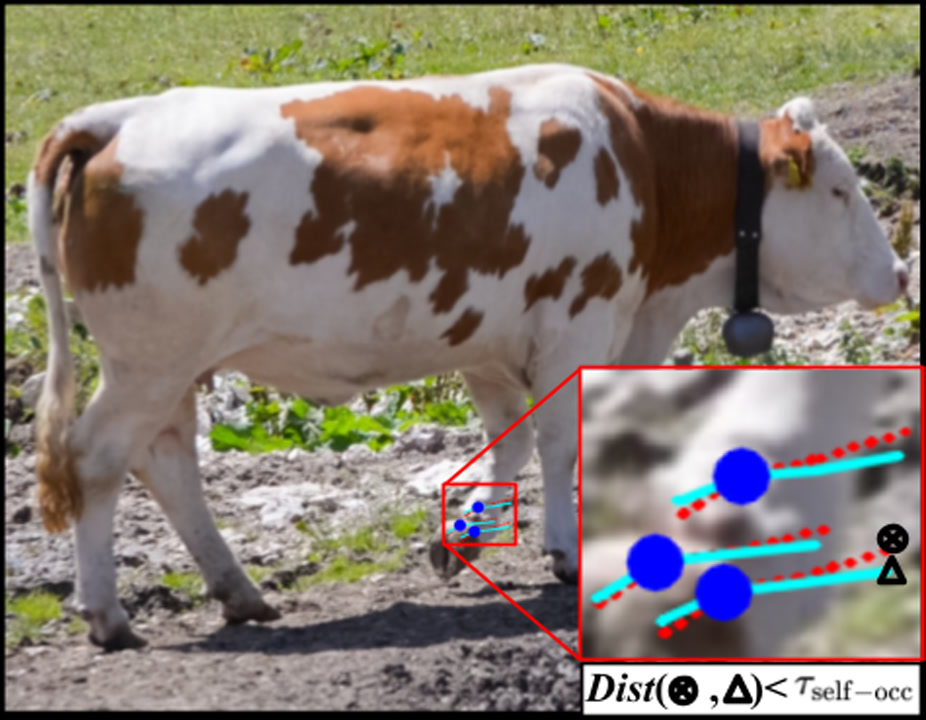}
  \caption{Non self-occlusion}
\end{subfigure}\hfill
\begin{subfigure}[t]{0.495\linewidth}\centering
  \includegraphics[width=\linewidth]{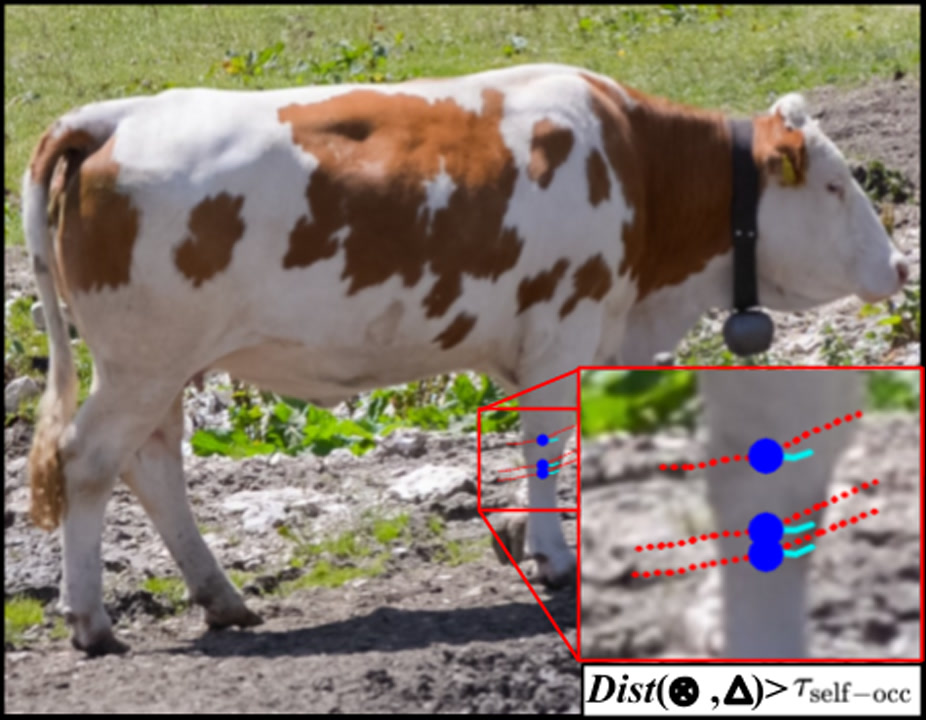}
  \caption{Self-occlusion}
\end{subfigure}

\caption{\textbf{Self-occlusion filtering.}
Self-occlusion occurs when one part of an object hides another.
During re-identification (Sec.~\ref{sec:init}) such points can be
mistakenly marked \emph{visible}. We resample each suspect point and
retrack it over a short window (cyan curve). If the point is truly
self-occluded, the new track adheres to the occluding surface and
diverges from the original trajectory (red dashed curve). We measure
the 2-D distance $Dist(\cdot,\cdot)$ between the two paths and relabel
the track \emph{occluded} whenever this distance exceeds
$\tau_{\text{self-occ}}$.}
\label{fig:self_occlusion}
\end{figure}

To suppress self-occlusion cases where a track is hidden by another part of the same object, yet still falls inside the same mask $M_t^{(o)}$ (Fig.~\ref{fig:self_occlusion}), we leverage motion cues. For each recovered track, we resample its position over a short window $[t\!-\!2,\,t\!+\!2]$. If the resampled trajectory $\{\hat{u}_{t'}\}$ ever diverges from the original $\{u_{t'}\}$ by more than $\tau_{\text{self-occ}}=10$ pixels:
\[
\max_{t' \in [t-2,t+2]} \text{Dist}(\hat{u}_{t'},u_{t'}) > \tau_{\text{self-occ}},
\]
we keep the track marked as \emph{invisible}.


Leveraging the masks $M_t^{(o)}$, we (i) generate high-quality, temporally consistent depth maps $D_{t=1}^T$ and (ii) obtain numerous robust tracks $u_t^{(n)}$, especially along thin, complexly moving structures, thereby enhancing the quality of dynamic-scene reconstruction.


\subsection{Lifting to 3-D}
\label{sec:lift}
We keep the lifting stage identical to the procedure in MoSca~\cite{lei2024mosca}. In brief, dynamic pixels are back-projected to initialize 3-D Gaussians, and long-term 2-D tracks are promoted to motion-scaffold
nodes \(v^{(m)}\). The ensuing joint space–time optimization enforces coherent motion and
fills in trajectories through occlusion. (See.Fig.~\ref{fig:method_overview} (II)) 


\subsection{Dynamic Scene Reconstruction}
\label{sec:dyn_recon}

We render RGB $\hat{I}_t$, depth $\hat{D}_t$, and optical flow $\hat{F}_{t\rightarrow t'}$ with a Gaussian-splatting renderer~\cite{ye2025gsplat} and supervise them with the 2-D priors from Sec.~\ref{sec:init}:

\[
\begin{aligned}
L_\mathrm{rgb} &=
  \lVert \hat{I}_t - I_t \rVert_1
  + 0.1\cdot SSIM(\hat{I}_t,I_t),
\\
L_\mathrm{depth} &=
  \lVert \hat{D}_t - D_t \rVert_1,
\\
L_\mathrm{track}^\mathrm{gaussian} &=
  \lVert
    u_t^{(n)}
    + \hat{F}_{t\rightarrow t'}[u_t^{(n)}]
    - u_{t'}^{(n)}
  \rVert_2 .
\end{aligned}
\]

Appearance $L_\mathrm{rgb}$ and geometry $L_\mathrm{depth}$ losses supervise Gaussians; track loss propagates motion from Gaussians to the scaffold nodes, which are regularized with as-rigid-as-possible (i.e. ARAP) ($L_\mathrm{arap}$), velocity ($L_\mathrm{vel}$), and acceleration ($L_\mathrm{acc}$) terms to ensure smooth, coherent motion.
Although these losses follow \cite{lei2024mosca}, we introduce two additional terms that better constrain structure of scene and dynamic-object.

\begin{figure}[h]
\centering

\mbox{%
  \makebox[0.33\linewidth][c]{\footnotesize\bfseries }%
  \makebox[0.33\linewidth][c]{\footnotesize\bfseries Before Regularization}%
  \makebox[0.33\linewidth][c]{\footnotesize\bfseries After Regularization}%
}
\par\smallskip

\begin{subfigure}[t]{0.33\linewidth}\centering
  \includegraphics[width=\linewidth]{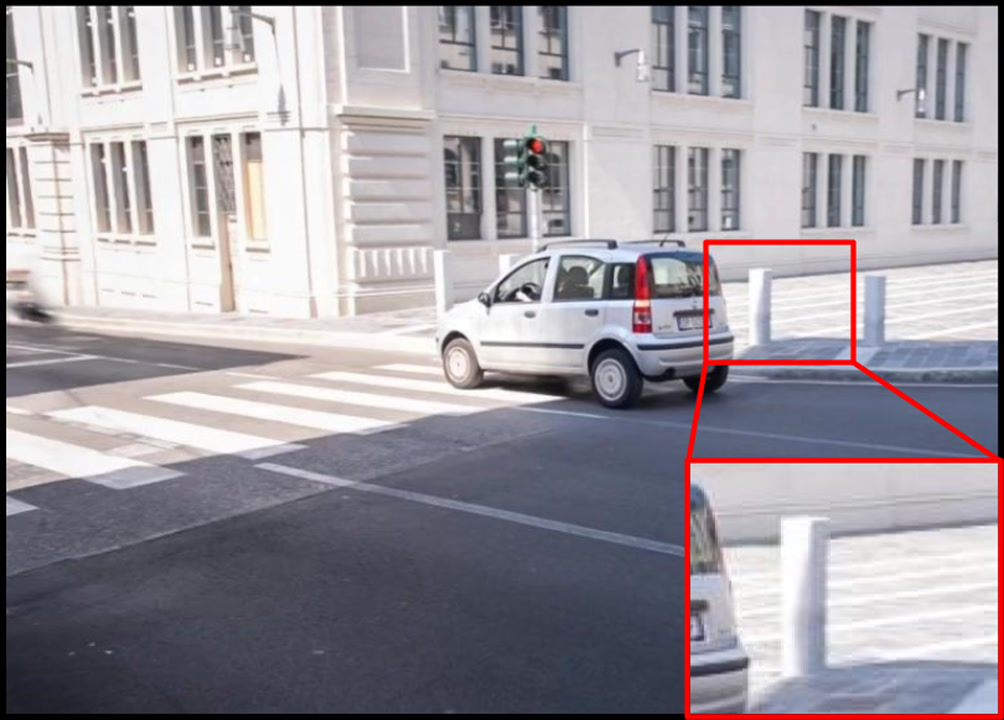}
  \caption{Training View Synthesis}
\end{subfigure}\hfill
\begin{subfigure}[t]{0.33\linewidth}\centering
  \includegraphics[width=\linewidth]{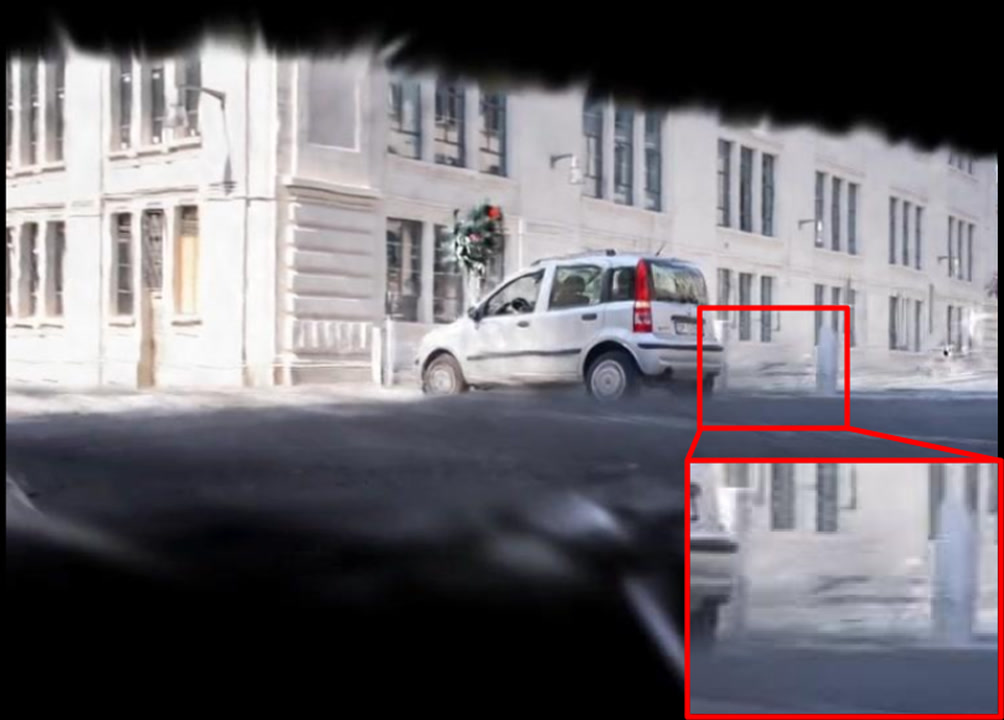}
  \caption{Virtual View Synthesis}
\end{subfigure}\hfill
\begin{subfigure}[t]{0.33\linewidth}\centering
  \includegraphics[width=\linewidth]{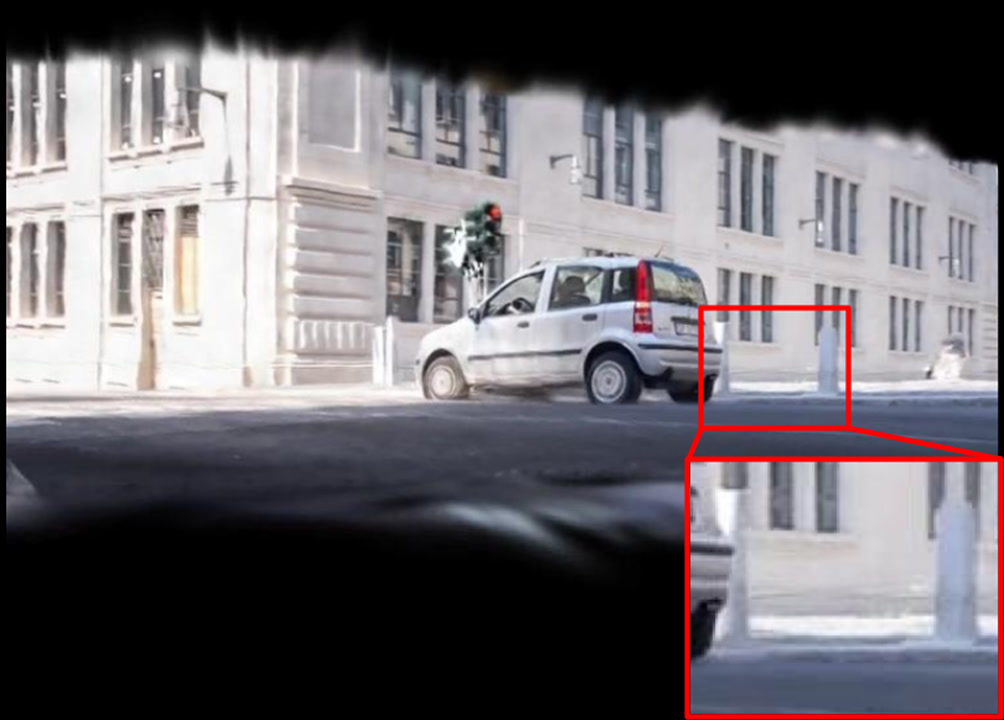}
  \caption{Virtual View Synthesis}
\end{subfigure}

\par\smallskip 

\begin{subfigure}[t]{0.33\linewidth}\centering
  \includegraphics[width=\linewidth]{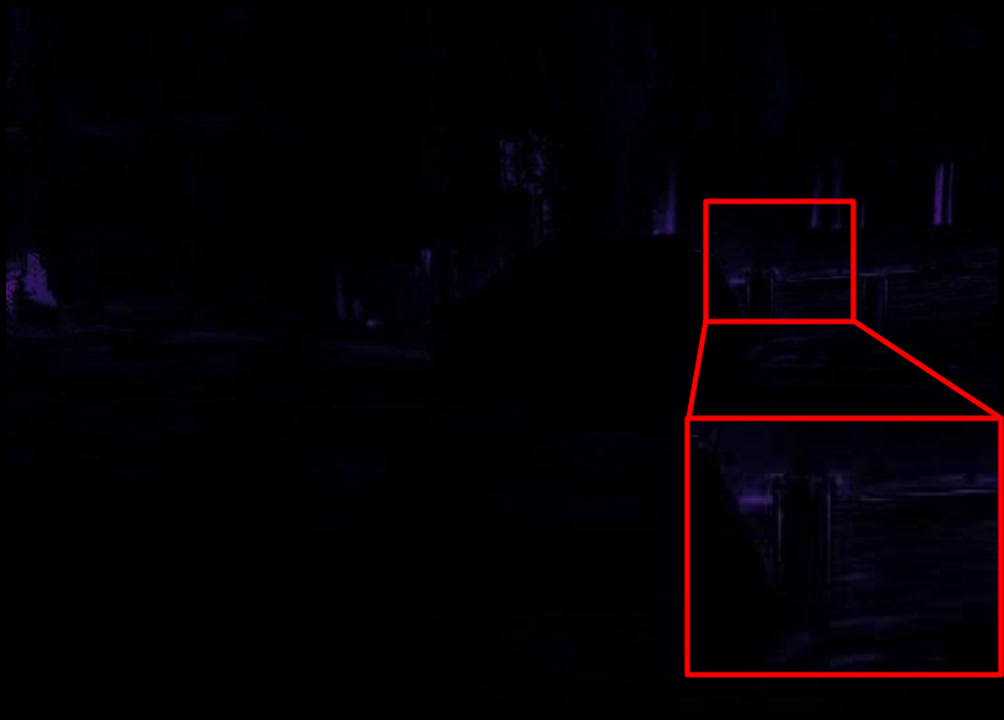}
  \caption{Training View Depth Loss}
\end{subfigure}\hfill
\begin{subfigure}[t]{0.33\linewidth}\centering
  \includegraphics[width=\linewidth]{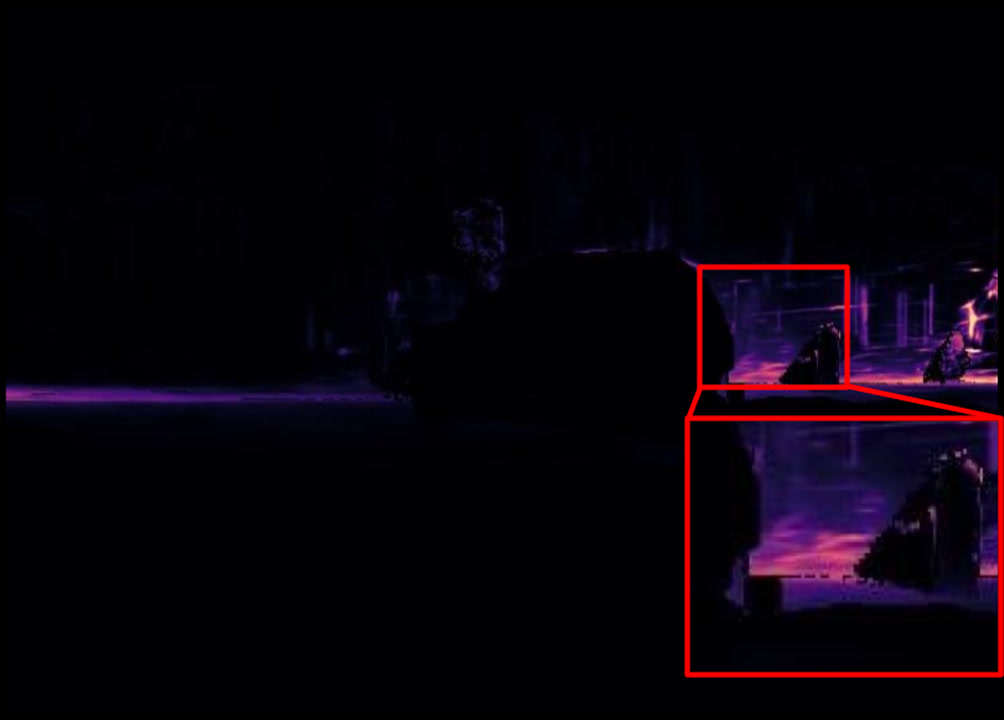}
  \caption{Virtual View Depth Loss}
\end{subfigure}\hfill
\begin{subfigure}[t]{0.33\linewidth}\centering
  \includegraphics[width=\linewidth]{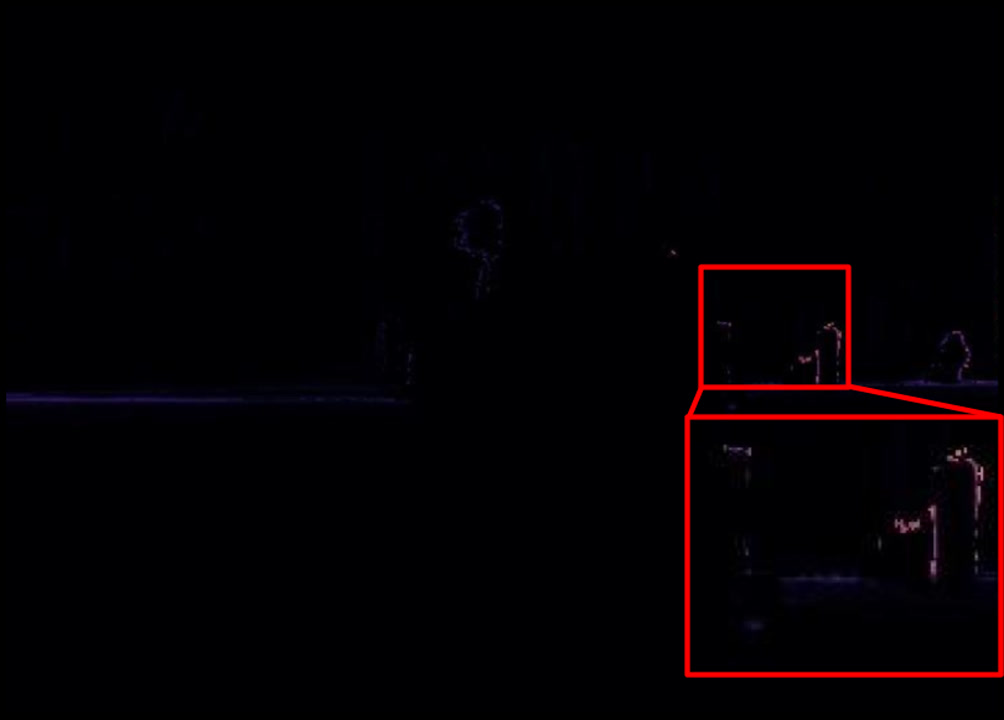}
  \caption{Virtual View Depth Loss}
\end{subfigure}

\caption{\textbf{Virtual-view depth loss.}
Gaussian splatting fits the training views well (a, d) but shows
floaters from virtual viewpoints (b).
A depth loss computed in these views, $L_\mathrm{depth}^\mathrm{virtual}$,
localizes the errors of floaters (e); adding it as a regularizer removes
the floaters and produces cleaner renderings (c, f), enabling broader
novel-view synthesis.}
\label{fig:virtual_depth_loss}
\end{figure}

\paragraph{Virtual-view depth loss.}
Small camera baseline videos let Gaussians overfit training views, producing floaters visible only from novel viewpoints (Fig.~\ref{fig:virtual_depth_loss}). Inspired by few-shot reconstruction regularizers \cite{chen2022geoaug,niemeyer2022regnerf,jain2021putting,truong2023sparf,yin2024fewviewgs}, we generate a virtual camera $\Pi_t^\mathrm{virtual}$ by randomly translating $\Pi_t$ in the image plane, warp $D_t$ to that view to obtain $D_t^\mathrm{virtual}$, and compute the following regularization for reconstruction.
\[
L_\mathrm{depth}^\mathrm{virtual}
=
\lVert \hat{D}_t^\mathrm{virtual} - D_t^\mathrm{virtual} \rVert_1 .
\]



\begin{figure}[h]
\centering
\begin{subfigure}[t]{0.495\linewidth}\centering
  \includegraphics[width=\linewidth]{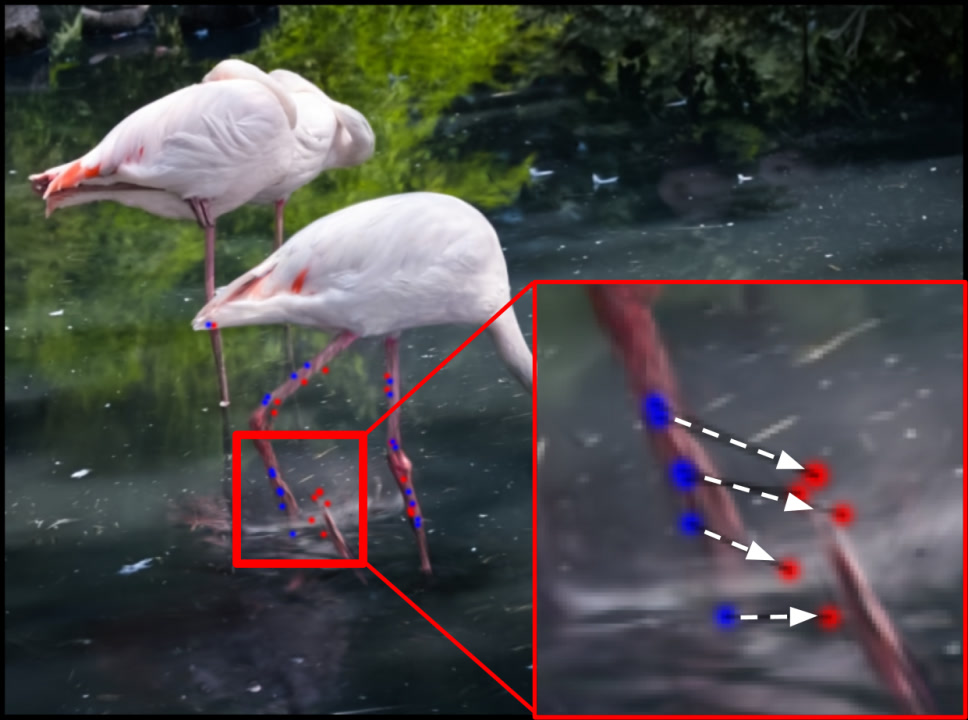}
  \caption{w/o $L^{\text{scaffold}}_{\text{track}}$}
\end{subfigure}\hfill
\begin{subfigure}[t]{0.495\linewidth}\centering
  \includegraphics[width=\linewidth]{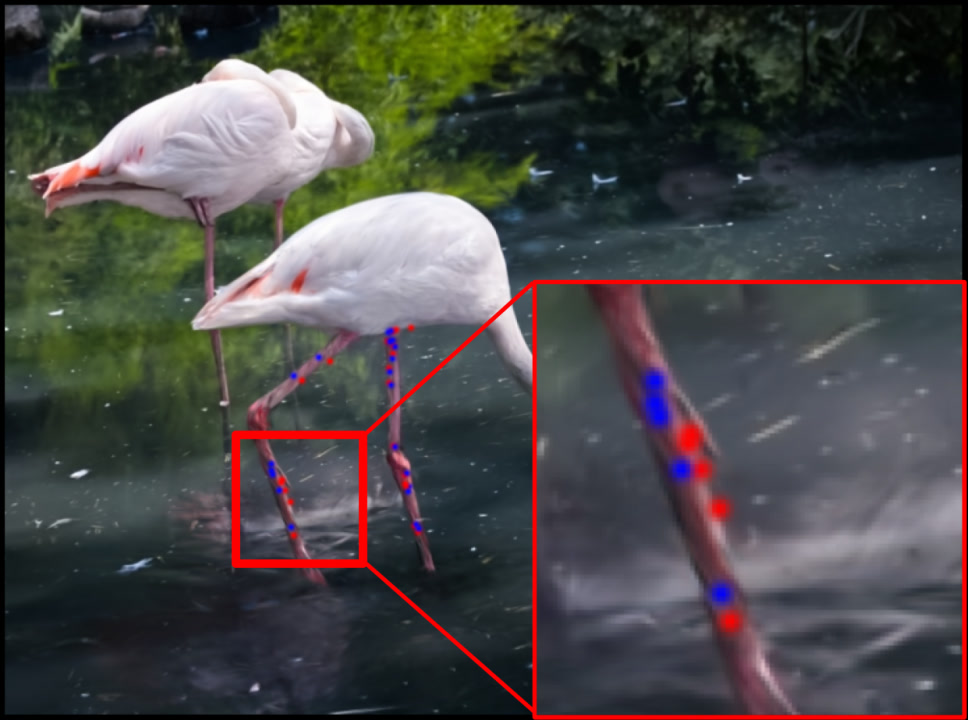}
  \caption{w/ $L^{\text{scaffold}}_{\text{track}}$}
\end{subfigure}

\caption{\textbf{Scaffold tracking regularization.}
ARAP regularization can let scaffold nodes drift off the object, especially on
thin or fast-moving parts. The initial scaffolds $v^{(m)}$ (blue dots) are
pushed away from the dynamic object during training (red dots). We add the
loss $L^{\text{scaffold}}_{\text{track}}$ to regularize the scaffold’s projected
2-D position toward the tracked points $u_t^{(n)}$ (see (b)).}
\label{fig:loss_scaffold}
\end{figure}

\paragraph{Scaffold-projection loss.} 
ARAP regularization ($L_\mathrm{arap}$) occasionally lets scaffold nodes drift off the object, especially on thin, fast-moving parts (Fig.~\ref{fig:loss_scaffold}). Because each node $v^{(m)}$ originates from a track $u_t^{(n)}$, we project the node into the camera, get $v_{2D}$ and penalize its 2-D distance to the corresponding
track:
\[
L_\mathrm{track}^\mathrm{scaffold}
=
\lVert v_{2D} - u_t^{(n)} \rVert_2 .
\]


\section{Experimental Results}

\begin{table}[t]
\centering
\caption{\textbf{Quantitative comparison on iPhone dataset~\cite{gao2022monocular}.}}
\vspace{-3mm}
\label{tab:quantitative_iphone}
\resizebox{\columnwidth}{!}{%
\begin{tabular}{lccc ccc}
\toprule
\multirow{2}{*}{Method} & \multicolumn{3}{c}{(a) \emph{Pose-free RGB}} & \multicolumn{3}{c}{(b) \emph{Depth \& pose}} \\ \cmidrule(lr){2-4} \cmidrule(lr){5-7}
 & \textbf{mPSNR} & \textbf{mSSIM} & \textbf{mLPIPS} & \textbf{mPSNR} & \textbf{mSSIM} & \textbf{mLPIPS} \\
 \midrule
S.o.M. & 17.15 & 0.625 & 0.278 & 17.32     & 0.598   & 0.296 \\
MoSca  & 17.24 & 0.603 & 0.304 & 19.32     & 0.706   & 0.264 \\
DpDy   & --    & --    & --    & --        & 0.559   & 0.516 \\
Cat4D  & --    & --    & --    & 18.24     & 0.666   & \textbf{0.227} \\
Ours & \textbf{17.63} & \textbf{0.648} & \textbf{0.268} & \textbf{19.43}     & \textbf{0.711}   & 0.260 \\
\bottomrule
\end{tabular}
}
\vspace{-1mm}
\end{table}
\begin{table}[t]
\centering
\caption{\textbf{Quantitative comparison on NVIDIA dataset without ground-truth camera poses.}}
\label{tab:quantitative_nvidia}
\vspace{-5mm}
\setlength{\tabcolsep}{6pt}      
\renewcommand{\arraystretch}{1.05}

\begin{tabular}{
  >{\footnotesize}l|   
  >{\footnotesize}c|   
  >{\footnotesize}c|   
  >{\footnotesize}c    
}
        & RoDynRF & MoSca & \textbf{Ours} \\ \hline
\textbf{PSNR/LPIPS} & 25.38/0.079 & 26.54/0.073 & \textbf{26.58/0.067} \\
\end{tabular}
\end{table}

\begin{table}[t]
\centering
\caption{\textbf{Ablation study.}}
\label{tab:ablation_study}
\vspace{-3mm}
\setlength{\tabcolsep}{2pt}      
\renewcommand{\arraystretch}{0.8}

\begin{tabular}{                       
  l|
  >{\footnotesize}c|
  >{\footnotesize}c|
  >{\footnotesize}c|
  >{\footnotesize}c
}
        & {\footnotesize\textbf{Ours}}
        & {\footnotesize w/o $\mathcal{L}_{\text{track}}^{\text{scaffold}}$}
        & {\footnotesize w/o Mask-Guided Track}
        & {\footnotesize w/o $\mathcal{L}_{\text{depth}}^{\text{virtual}}$} \\ \hline
{\footnotesize\textbf{PSNR}}  & 17.63 & 17.61 & 17.55 & \textbf{17.64} \\
{\footnotesize\textbf{SSIM}}  & \textbf{0.648} & 0.637 & 0.632 & 0.630 \\
{\footnotesize\textbf{LPIPS}} & \textbf{0.268} & 0.270 & 0.274 & 0.277 \\
\end{tabular}
\end{table}

\subsection{Implementation details.}
We set $\tau_\mathrm{salient}=0.05$ and $\tau_\mathrm{appearance}=0.2$ to pick dynamic masks.
Skeletons (OpenCV) guide sampling: 3000 of 19384 tracks are drawn from the skeleton, the remaining 16,384 are uniformly sampled as in \cite{lei2024mosca}.
Tracks are propagated with SpatialTracker~\cite{xiao2024spatialtracker}, and self-occluded ones are pruned using $\tau_\mathrm{self-occ}=10$ px.
Virtual views are generated by translating $\Pi_t$ up to $0.18 \mathrm{median}(D_t)$ in the image plane.
All other settings follow \cite{lei2024mosca}. The pipeline proceeds through (a) dynamic mask selection (17 min), (b) pose estimation and depth refinement (11 min), (c) mask-guided point tracking (12 min), (d) lifting and initialization to 3-D (5 min), and (e) reconstruction (25 min).

\subsection{Datasets}
\paragraph{iPhone DyCheck~\cite{gao2022monocular}.}
This benchmark contains 14 casually captured handheld videos (200–500 frames at 360$\times$480) of dynamic scenes, seven of which include two static ``held-out'' cameras for novel-view evaluation.  Besides RGB, ARKit provides noisy LiDAR depth and approximate poses.  We adopt two complementary protocols: (i) \emph{depth \& pose}: both LiDAR depth and ARKit poses are available during training, following setting in MoSca~\cite{lei2024mosca}. (ii) \emph{pose-free RGB}: neither LiDAR depth nor ARKit poses is available during training, making the task relies on pure monocular RGB.  For fair comparison under pose-free RGB, all methods perform identical test-time camera poses optimization while freezing their learned scene representations.  Evaluation reports novel view PSNR, SSIM, and LPIPS.

\paragraph{NVIDIA multi-view videos~\cite{yoon2020novel}.}
This dataset offers eight dynamic scenes filmed by a 16-camera rig with very small baselines. Following RoDynRF~\cite{liu2023robust}, we use a single forward-facing view as training input and treat the remaining 15 as unseen target views. No depth and camera poses are provided in our evaluation. We also adopt the same test-time poses optimization for all methods. Quantitative assessment uses PSNR and LPIPS.

\paragraph{DAVIS~\cite{perazzi2016benchmark}.}
To examine real-world robustness, we employ the 50 validation clips 
of DAVIS, which contain high-motion Internet videos (960x540, 25-40 fps) that lack calibration, depth, or multi-view supervision. Each sequence is treated as an unconstrained monocular video. All methods train using only RGB frames and synthesize novel viewpoints on a circular trajectory for qualitative evaluations. Because ground truth geometry is unavailable, we provide side-by-side renderings and highlight typical success and failure modes rather than reporting quantitative scores.

\subsection{Baselines}
We benchmark our pipeline against two state-of-the-art monocular 4D reconstruction systems chosen for their relevance to our pose-free setting and their publicly available code.

\paragraph{Shape of Motion~\cite{wang2024shape}} represents dynamic scenes with per-point linear combinations of $\mathfrak{se}(3)$ motion bases, but requires accurate camera extrinsics as input. Following its public repository, we first estimate the training-view poses and depths with MegaSAM~\cite{li2024megasam} and then feed these data to Shape of Motion for full-sequence optimization. At test time, we keep the learned scene parameters fixed and perform the same pose refinement that we apply to all methods in our pose-free protocol, ensuring a fair comparison.


\paragraph{MoSca~\cite{lei2024mosca}} is a complete system that jointly solves for camera parameters, and dynamic scene reconstruction. The original evaluation exploits either LiDAR depth or ground truth poses provided by the DyCheck dataset. To follow the setting of dynamic reconstruction from monocular RGB video, we run the official code \emph{without} providing LiDAR depth maps, or ground truth camera parameters thereby limiting supervision to the same RGB frames available to our method. Following the information from their paper and code repository, we substitute LiDAR with~\cite{piccinelli2024unidepth} on DyCheck and~\cite{hu2024depthcrafter} on DAVIS. All other hyperparameters are kept at their default values.

For completeness, we also compare our method with DpDy~\cite{wang2024diffusion} and Cat4D~\cite{wu2024cat4d}. Note that, unlike our approach, these baselines require additional information (e.g. camera poses).

\subsection{Qualitative Comparison}
\label{sec:visual_comparison}

\begin{figure}[h]
\centering

\begin{subfigure}[t]{0.33\linewidth}\centering
  \includegraphics[width=\linewidth]{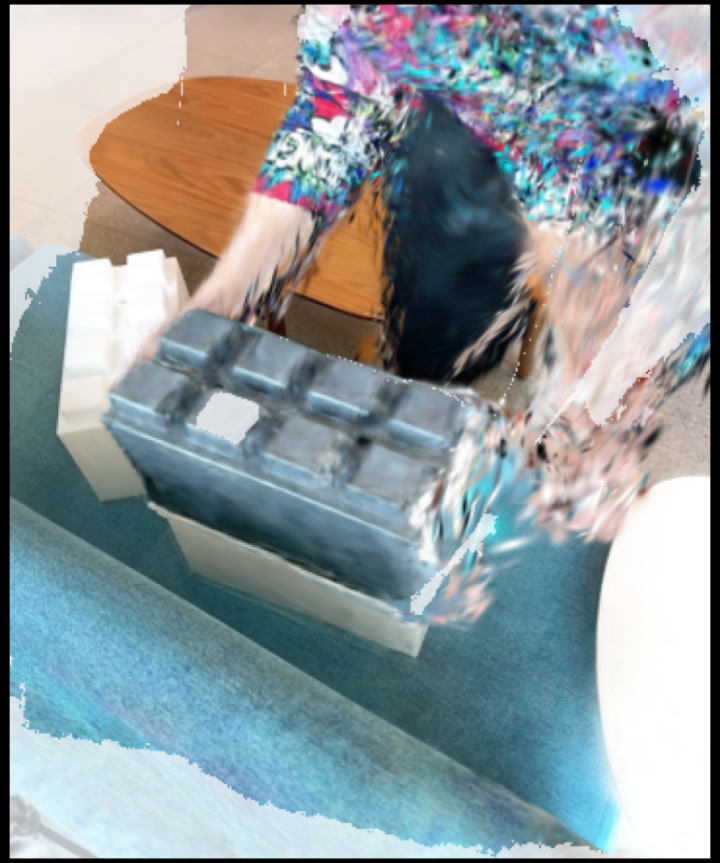}
  \caption{Shape of Motion}
\end{subfigure}\hfill
\begin{subfigure}[t]{0.33\linewidth}\centering
  \includegraphics[width=\linewidth]{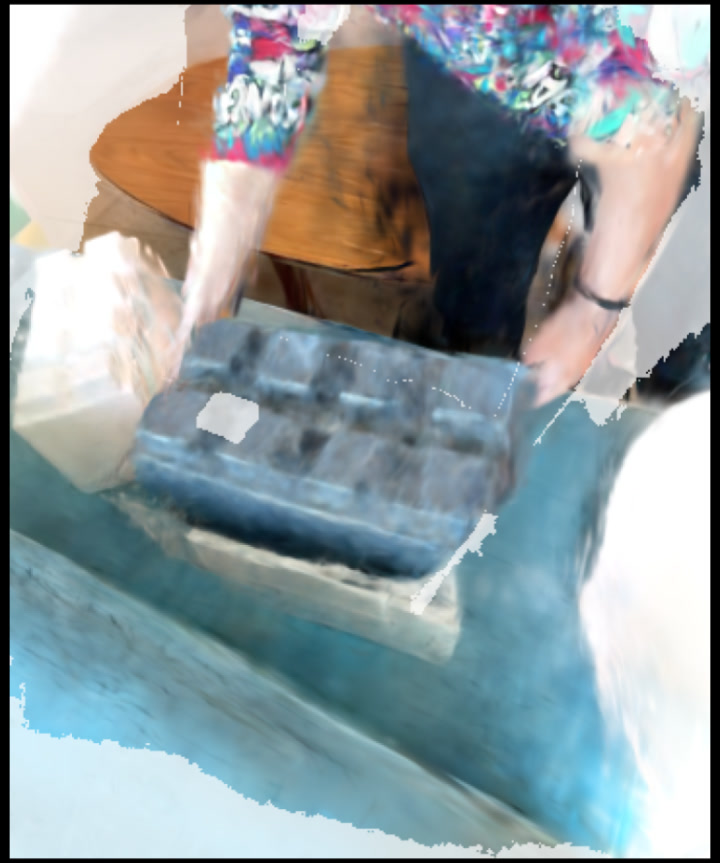}
  \caption{MoSca}
\end{subfigure}\hfill
\begin{subfigure}[t]{0.33\linewidth}\centering
  \includegraphics[width=\linewidth]{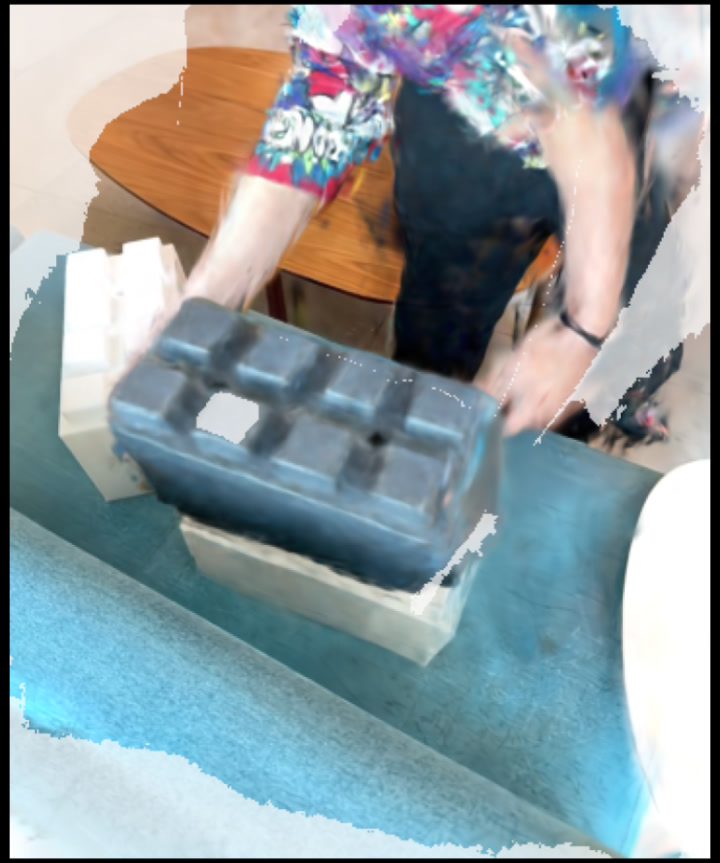}
  \caption{Ours}
\end{subfigure}


\begin{subfigure}[t]{0.33\linewidth}\centering
  \includegraphics[width=\linewidth]{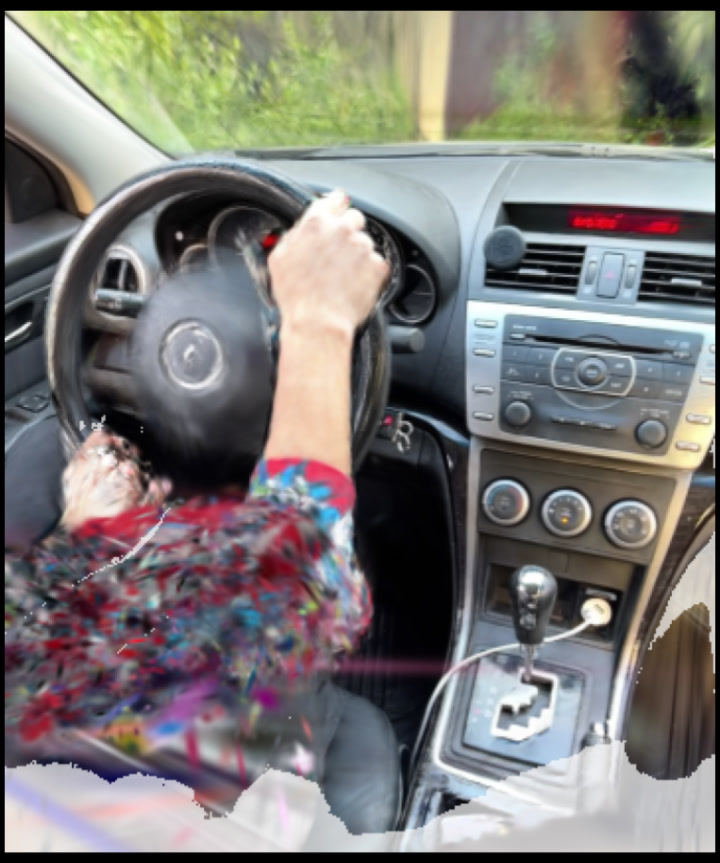}
  \caption{Shape of Motion}
\end{subfigure}\hfill
\begin{subfigure}[t]{0.33\linewidth}\centering
  \includegraphics[width=\linewidth]{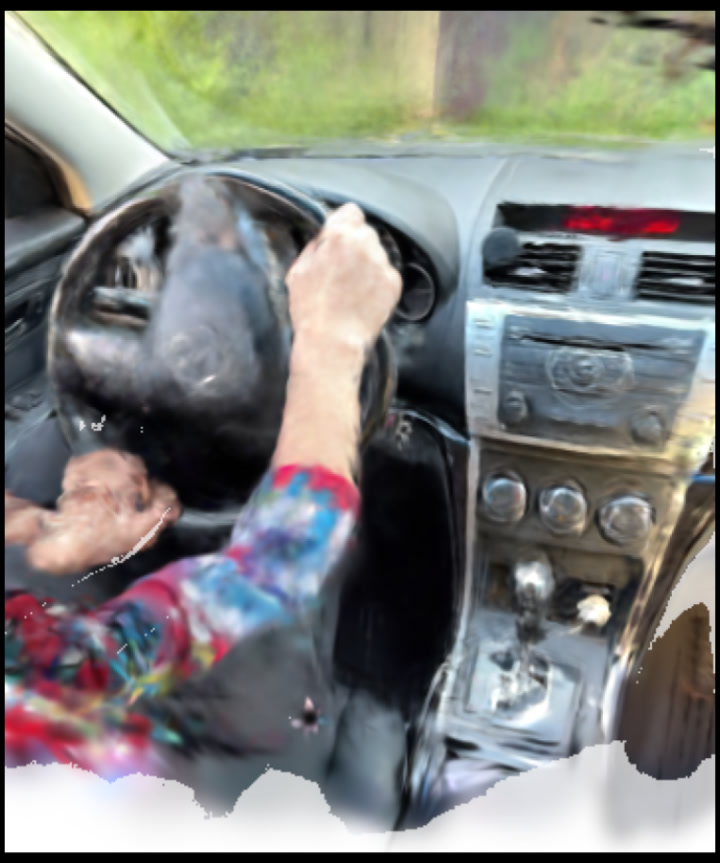}
  \caption{MoSca}
\end{subfigure}\hfill
\begin{subfigure}[t]{0.33\linewidth}\centering
  \includegraphics[width=\linewidth]{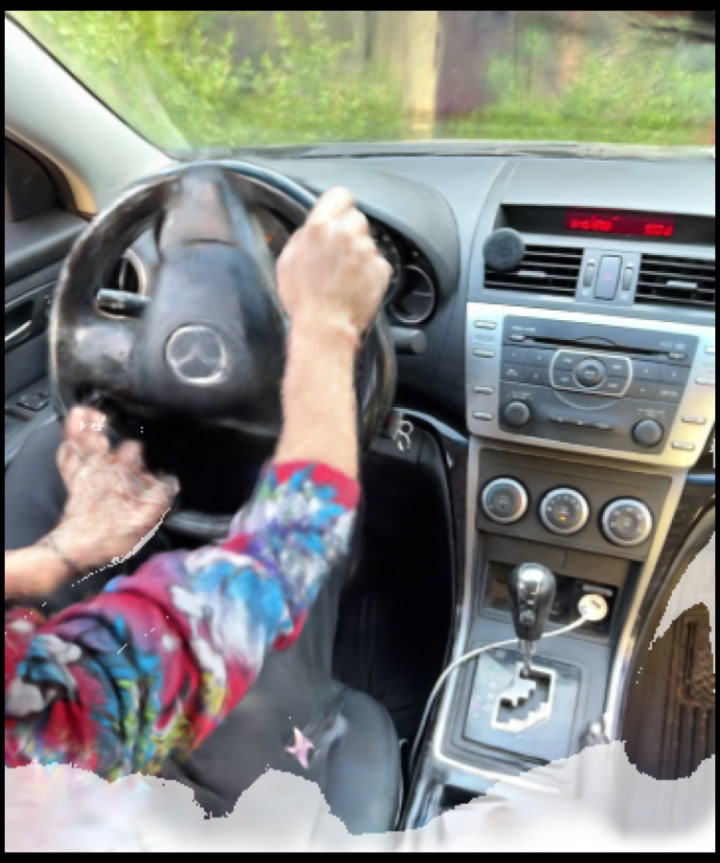}
  \caption{Ours}
\end{subfigure}

\caption{\textbf{Qualitative Comparison on iPhone Dataset.}
Without ground-truth camera poses and without LiDAR depth, we use only RGB
frames as input to the method.}
\label{fig:qual_iphone}
\end{figure}

\paragraph{iPhone DyCheck.}
Fig.~\ref{fig:qual_iphone} compares rendering results produced by Shape of Motion and MoSca with those of ours on DyCheck sequences. Shape of Motion suffers from severe tearing artifacts along depth discontinuities (e.g., box edges and driver arms), and its color bleeding reveals motion basis over-fitting. MoSca mitigates some tearing through joint pose optimization, yet its reliance on dense per-frame depth warping leaves large holes and ``melting'' geometry on rapidly moving limbs. In comparison, our method reconstructs complete geometry with crisp textures. Thin structures such as the cardboard slots and steering wheel spokes are preserved, and hand motion appears smooth. 

\begin{figure*}[h]    
\centering
\includegraphics[width=\linewidth]{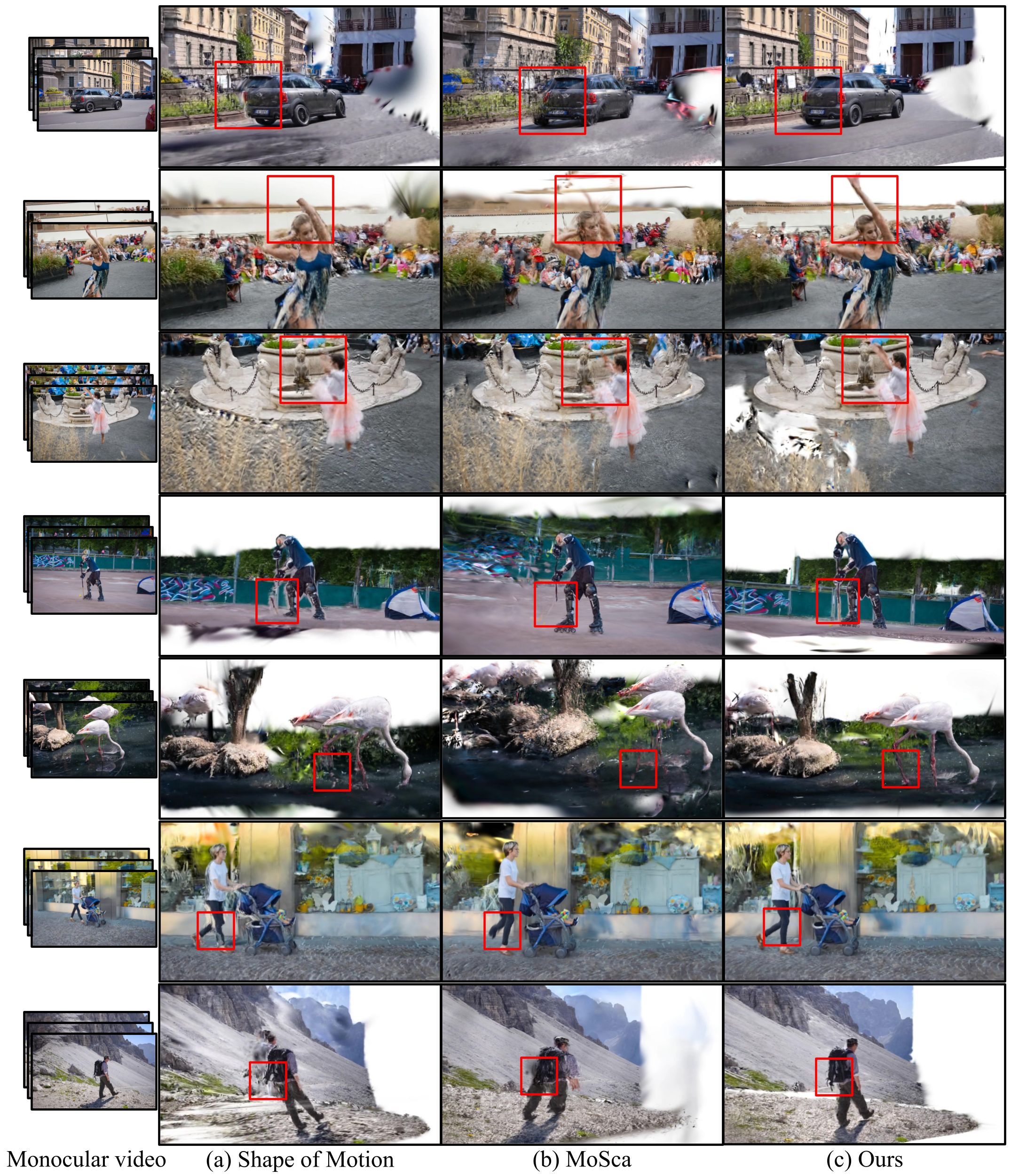}
\caption{
\textbf{Qualitative Comparison on DAVIS Dataset.}
}
\label{fig:qual_davis}
\end{figure*}

\paragraph{DAVIS in-the-wild videos.}
Fig.~\ref{fig:qual_davis} showcases challenging DAVIS clips containing fast-moving cars, dancers, fountains, and roller skaters. All sequences exhibit large inter-frame motion and complex occlusions with no auxiliary sensors. Shape of Motion frequently collapses background geometry outside the narrow view frustum estimated from its noisy monocular poses (see blurred facades and missing road), while MoSca’s per-pixel depth fusion produces artifacts on thin, rapidly moving objects (highlighted by red boxes). Our method retains fine details such as the dancer’s raised arm and the fountain’s statues by globally aggregating track-conditioned observations. It maintains consistent background parallax even where foreground motion is extreme. 

\subsection{Quantitative Comparison}
\label{sec:quanitative}

\paragraph{iPhone DyCheck.}
Table~\ref{tab:quantitative_iphone} summarizes results for the (a) \emph{pose-free RGB} and the (b) \emph{depth \& pose} settings, respectively.  In the strictly monocular regime (Table~\ref{tab:quantitative_iphone} (a)), our method performs on par with Shape of Motion and MoSca. When LiDAR depth and ground truth poses are provided (Table~\ref{tab:quantitative_iphone} (b)), all methods improve, and our scores again slightly outperform the baselines.

\paragraph{NVIDIA multi-view videos.}
On the forward-facing NVIDIA benchmark (Table~\ref{tab:quantitative_nvidia}), our method is marginally ahead of MoSca and clearly above RoDynRF.  Although the absolute gains are modest due to the small-baseline setup, the results confirm that our tracker-guided 4D splatting generalizes beyond handheld monocular footage to multi-camera data.

\subsection{Ablation Study}
\label{sec:ablation}
\begin{figure}[h]
\centering
\begin{subfigure}[t]{0.495\linewidth}\centering
  \includegraphics[width=\linewidth]{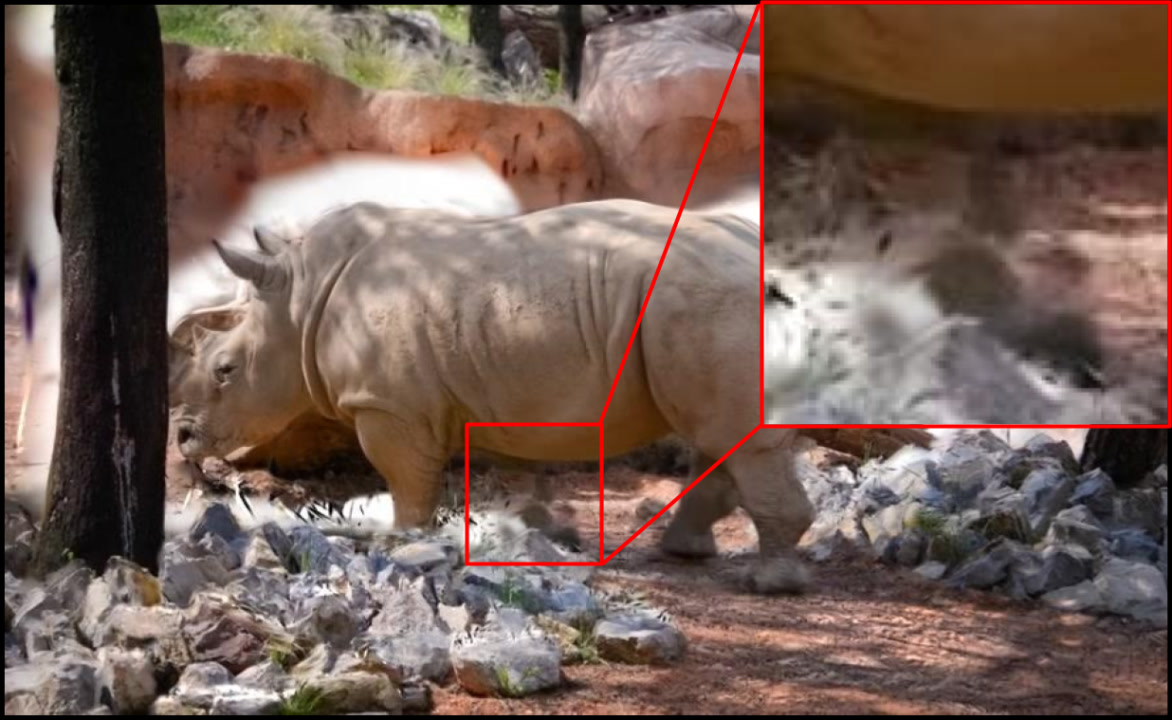}
  \caption{w/o skeleton sampling}
\end{subfigure}\hfill
\begin{subfigure}[t]{0.495\linewidth}\centering
  \includegraphics[width=\linewidth]{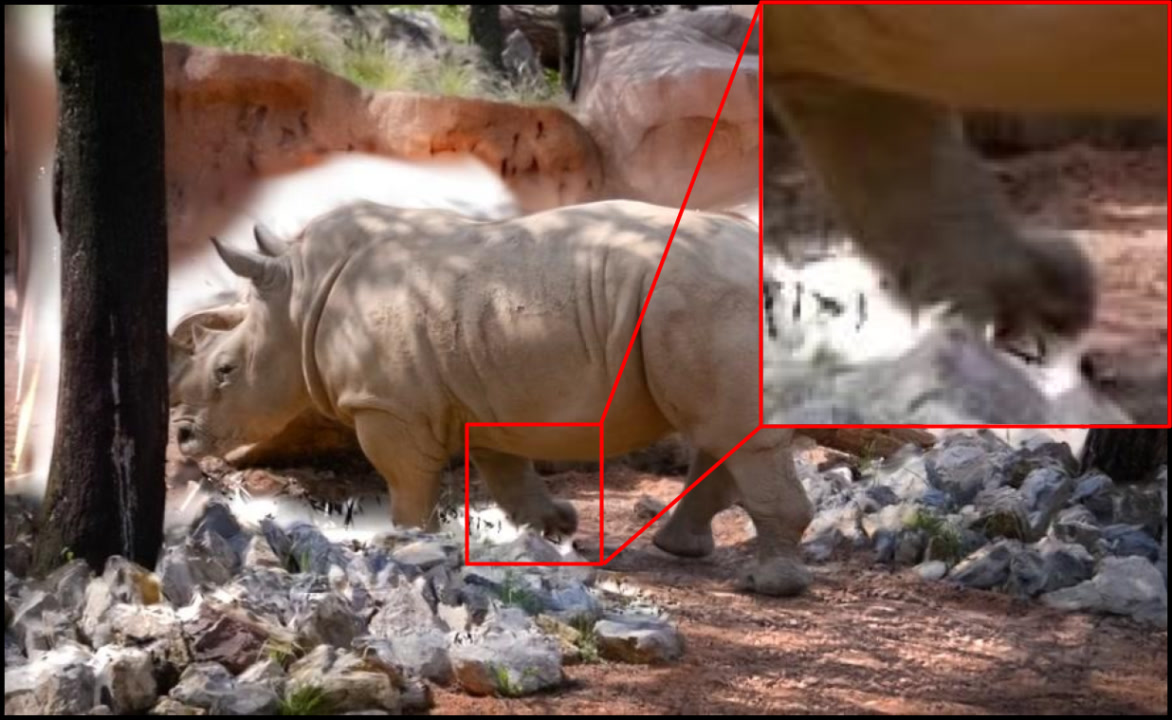}
  \caption{w/ skeleton sampling}
\end{subfigure}

\caption{Qualitative comparison of skeleton sampling.}
\label{fig:qual_skeleton}
\end{figure}

\begin{figure}[h]
\centering

\begin{subfigure}[t]{0.495\linewidth}\centering
  \includegraphics[width=\linewidth]{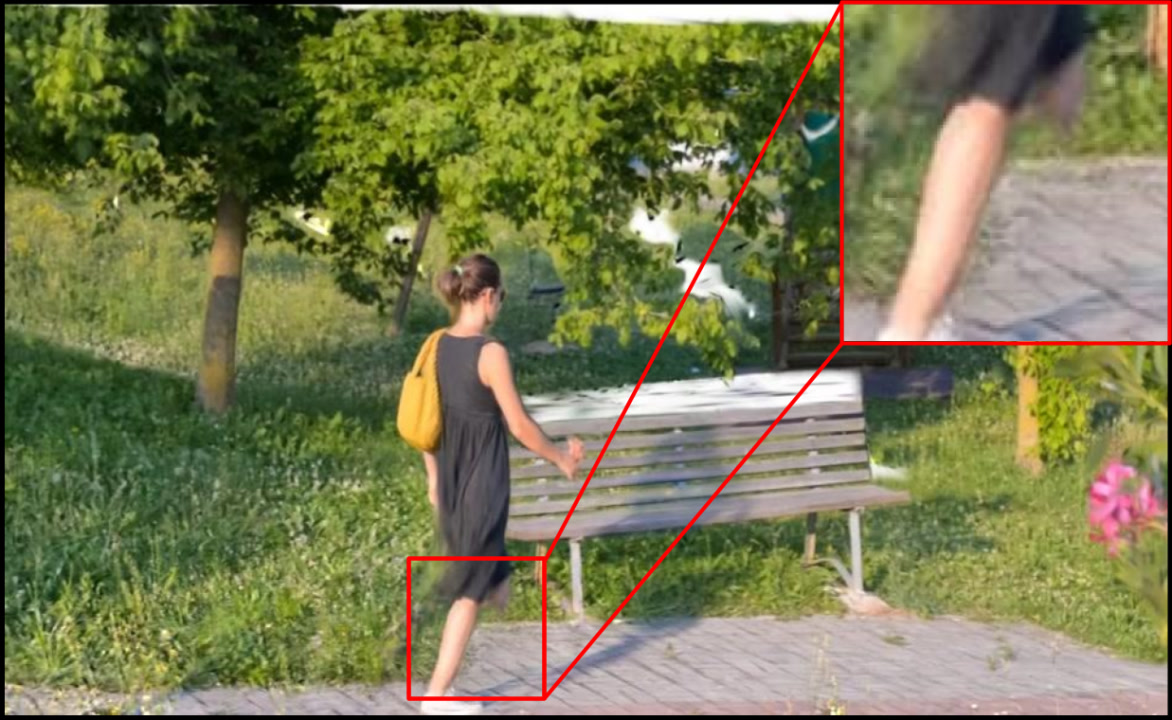}
  \caption{w/o track re-identification}
\end{subfigure}\hfill
\begin{subfigure}[t]{0.495\linewidth}\centering
  \includegraphics[width=\linewidth]{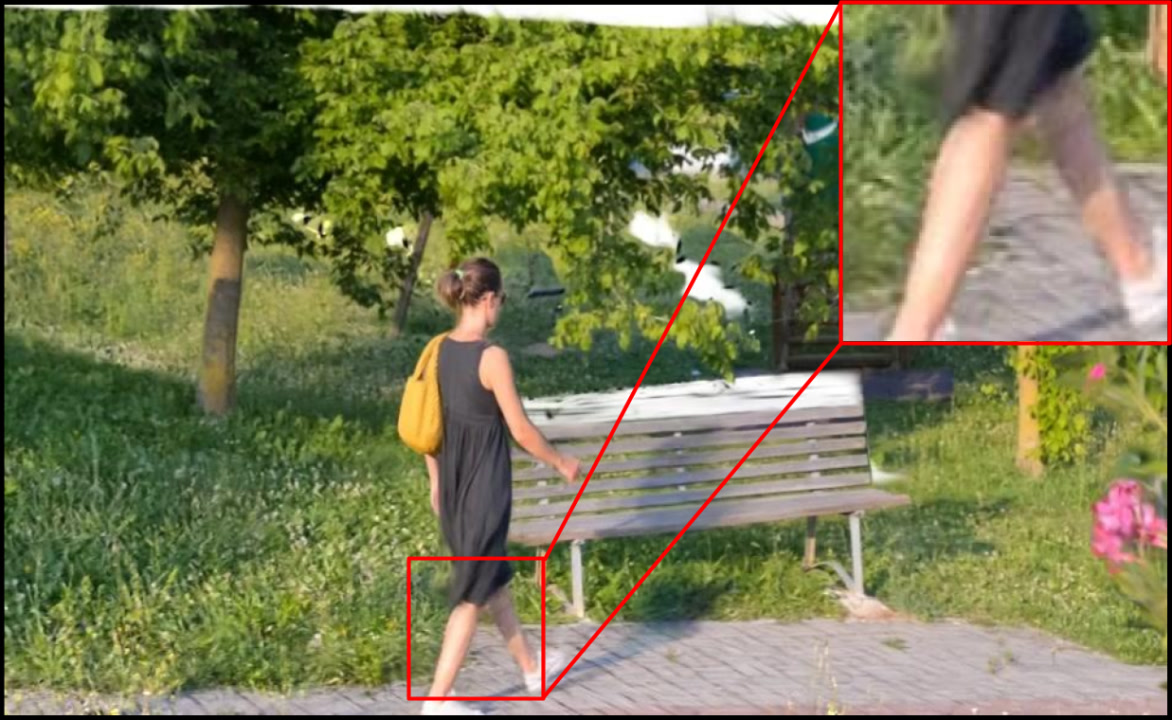}
  \caption{w/ track re-identification}
\end{subfigure}

\caption{Qualitative comparison of track re-identification.}
\label{fig:qual_track_reid}
\end{figure}

\begin{figure}[h]
\centering
\begin{subfigure}[t]{0.495\linewidth}\centering
  \includegraphics[width=\linewidth]{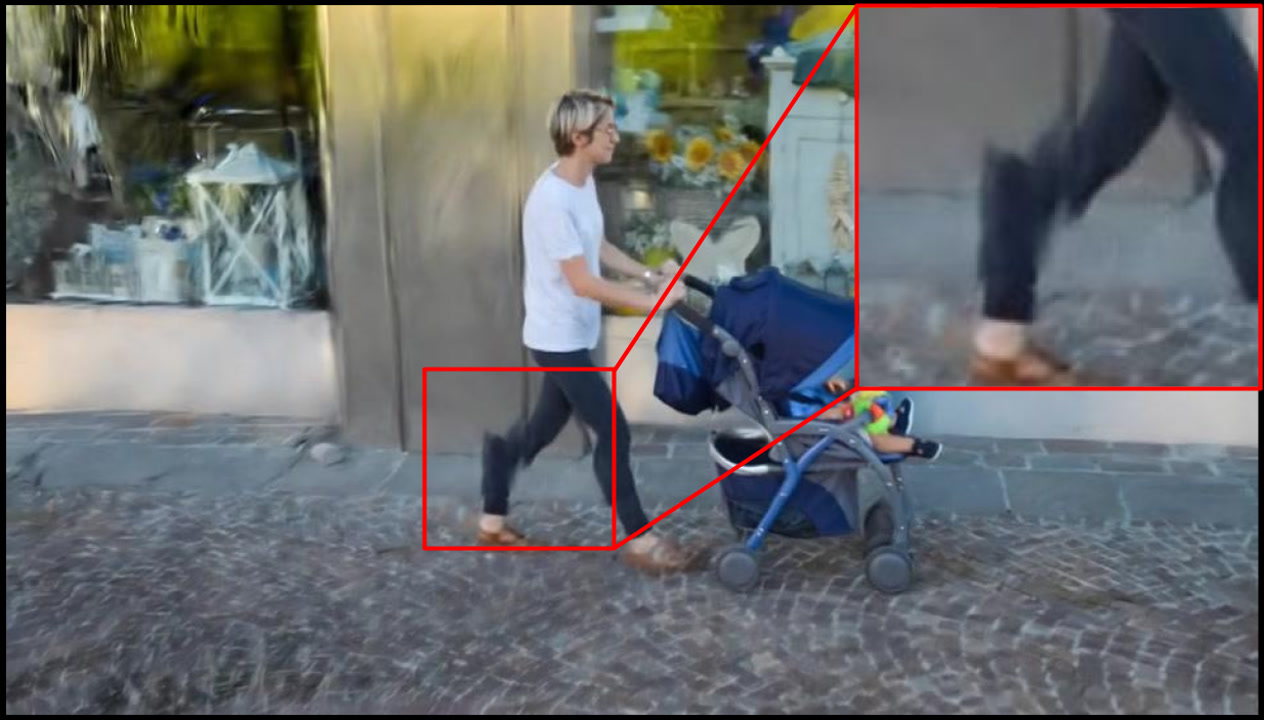}
  \caption{w/o $L^{\text{scaffold}}_{\text{track}}$}
\end{subfigure}\hfill
\begin{subfigure}[t]{0.495\linewidth}\centering
  \includegraphics[width=\linewidth]{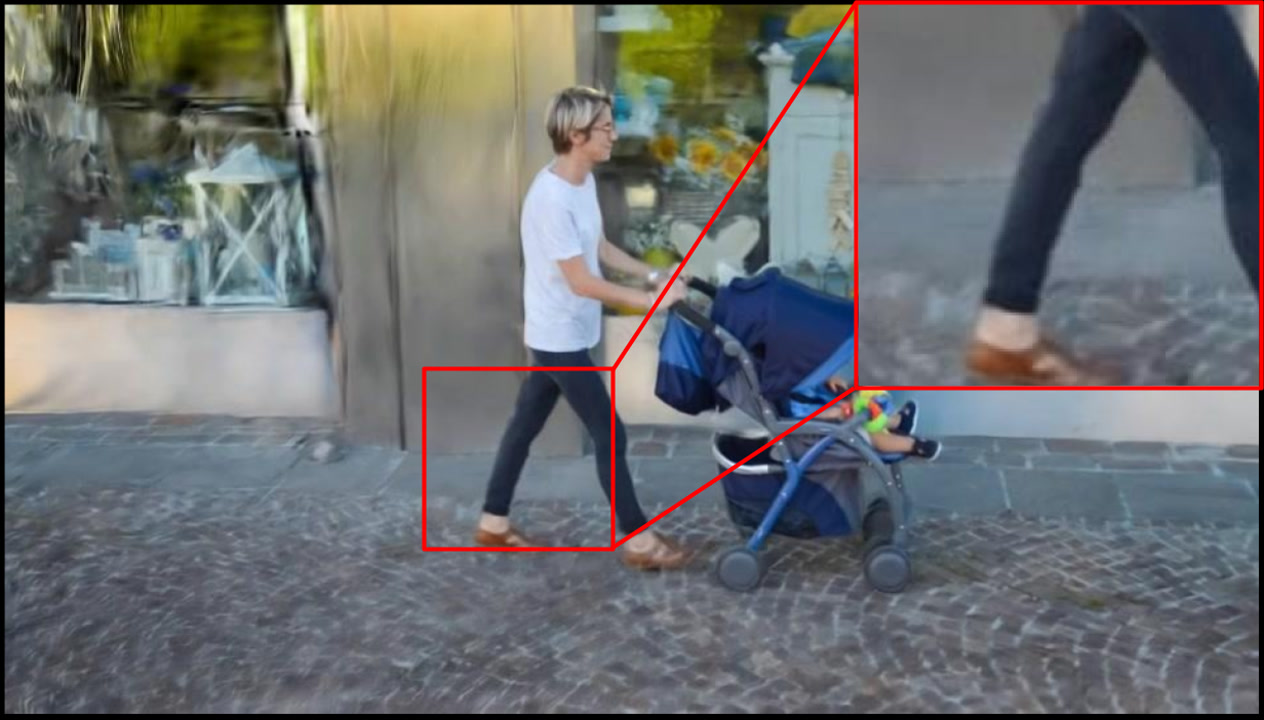}
  \caption{w/ $L^{\text{scaffold}}_{\text{track}}$}
\end{subfigure}

\caption{Effect of scaffold-track loss $L^{\text{scaffold}}_{\text{track}}$.}
\label{fig:qual_scaffold_loss}
\end{figure}

\begin{figure}[h]
\centering
\begin{subfigure}[t]{0.495\linewidth}\centering
  \includegraphics[width=\linewidth]{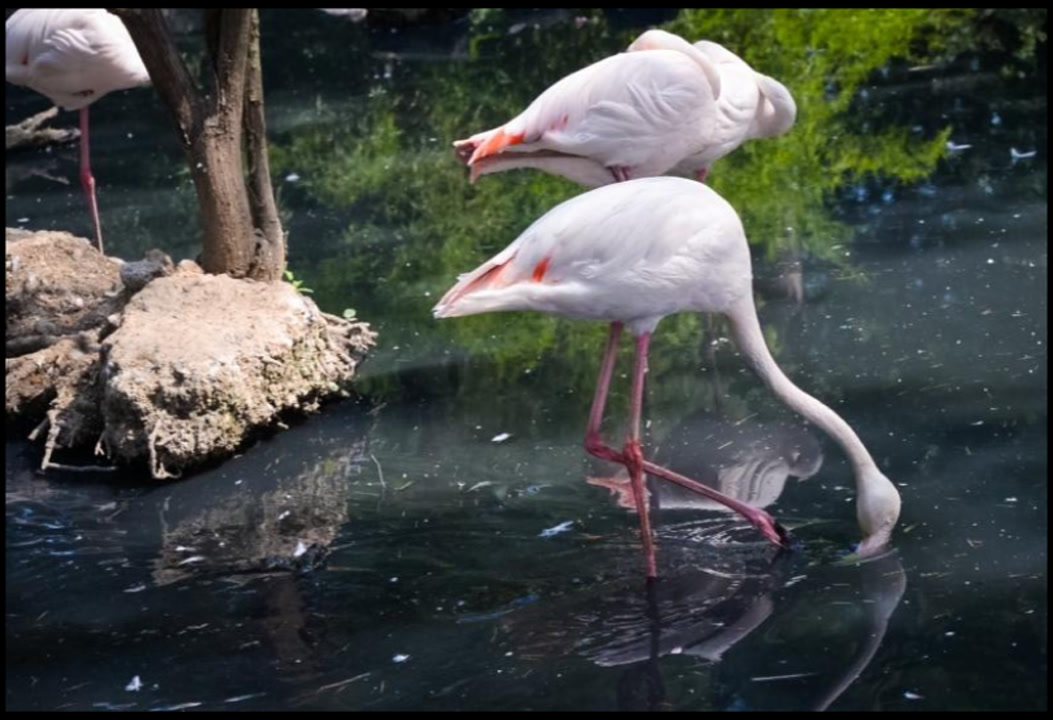}
  \caption{w/o $L^{\text{scaffold}}_{\text{track}}$}
\end{subfigure}\hfill
\begin{subfigure}[t]{0.495\linewidth}\centering
  \includegraphics[width=\linewidth]{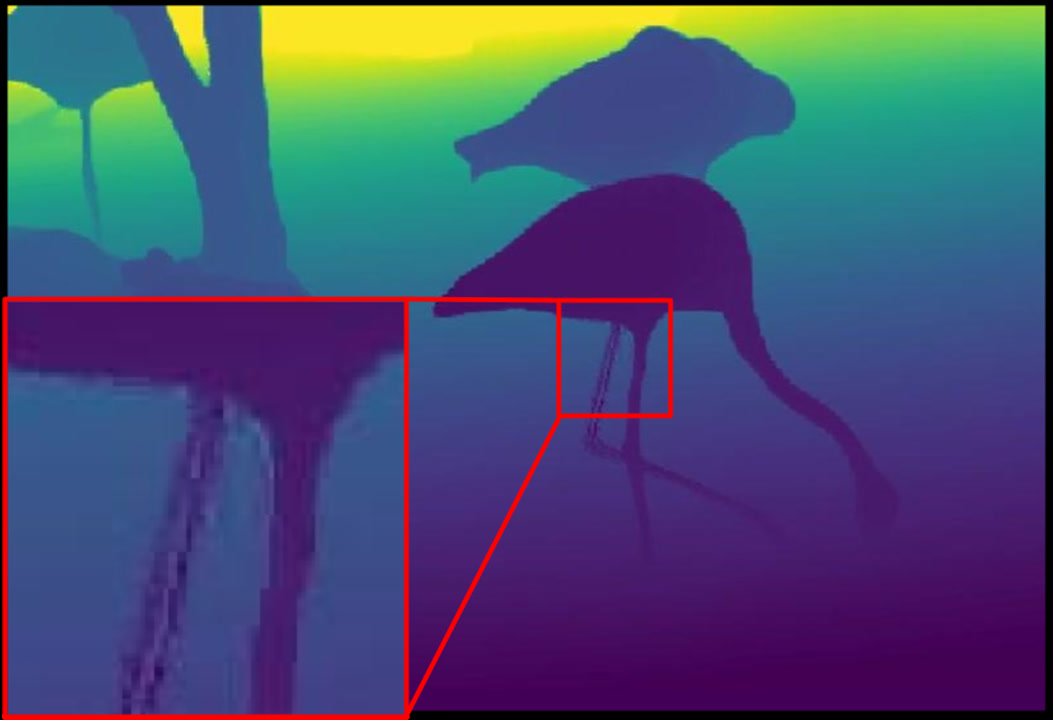}
  \caption{w/ $L^{\text{scaffold}}_{\text{track}}$}
\end{subfigure}

\caption{Effect of depth refinement on video depth $D$.}
\label{fig:qual_local_align}
\end{figure}

\begin{figure}[h]
\centering

\begin{subfigure}[t]{\linewidth}\centering
  \includegraphics[width=0.495\linewidth]{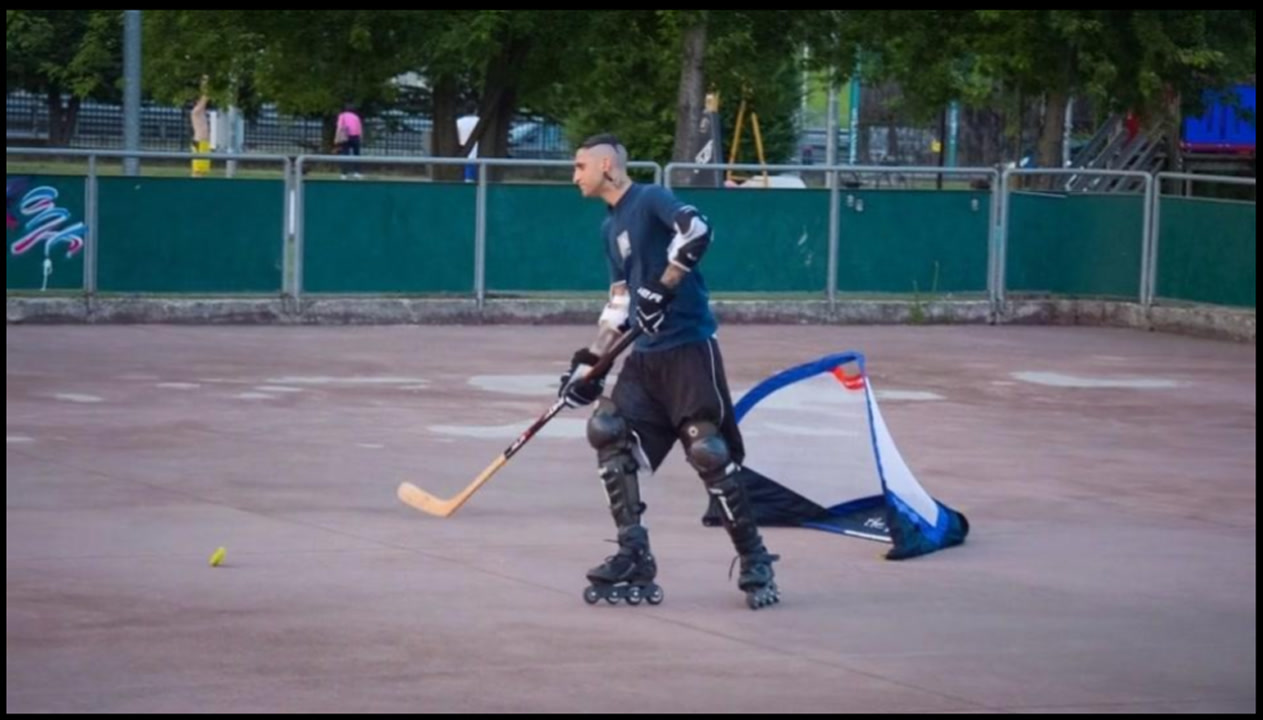}\hfill
  \includegraphics[width=0.495\linewidth]{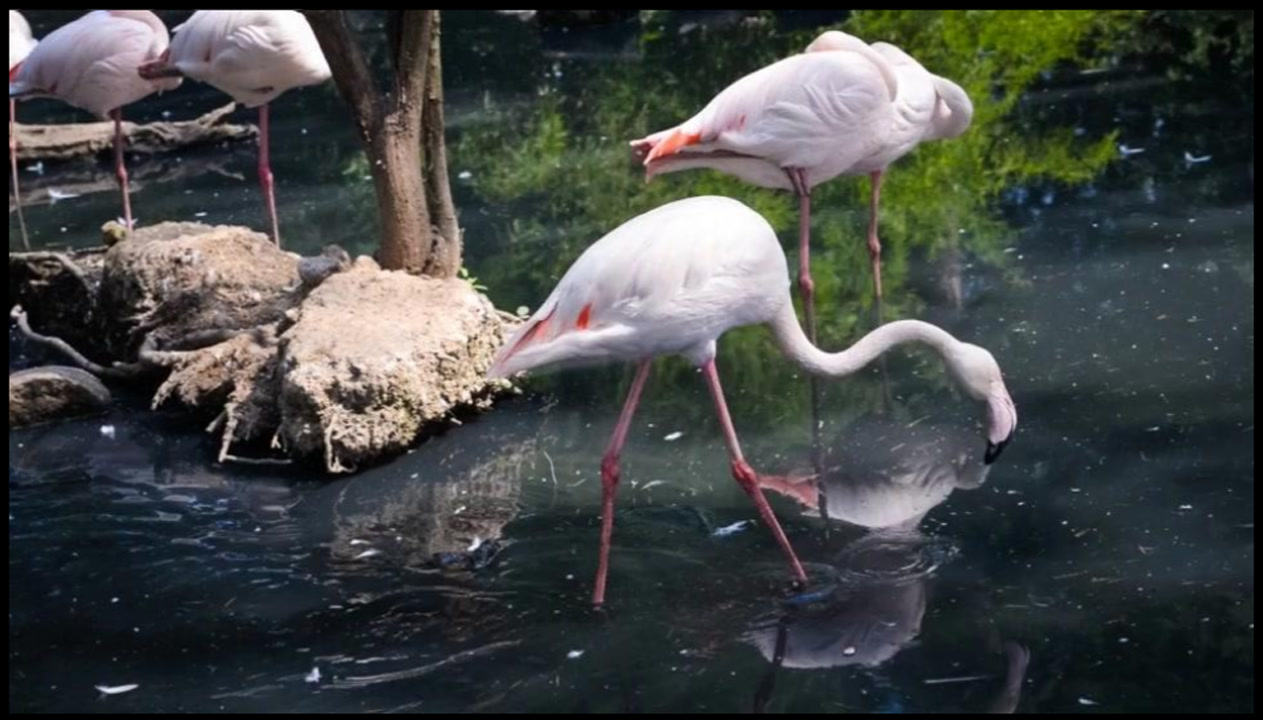}
  \caption{Monocular video}
\end{subfigure}


\begin{subfigure}[t]{\linewidth}\centering
  \includegraphics[width=0.495\linewidth]{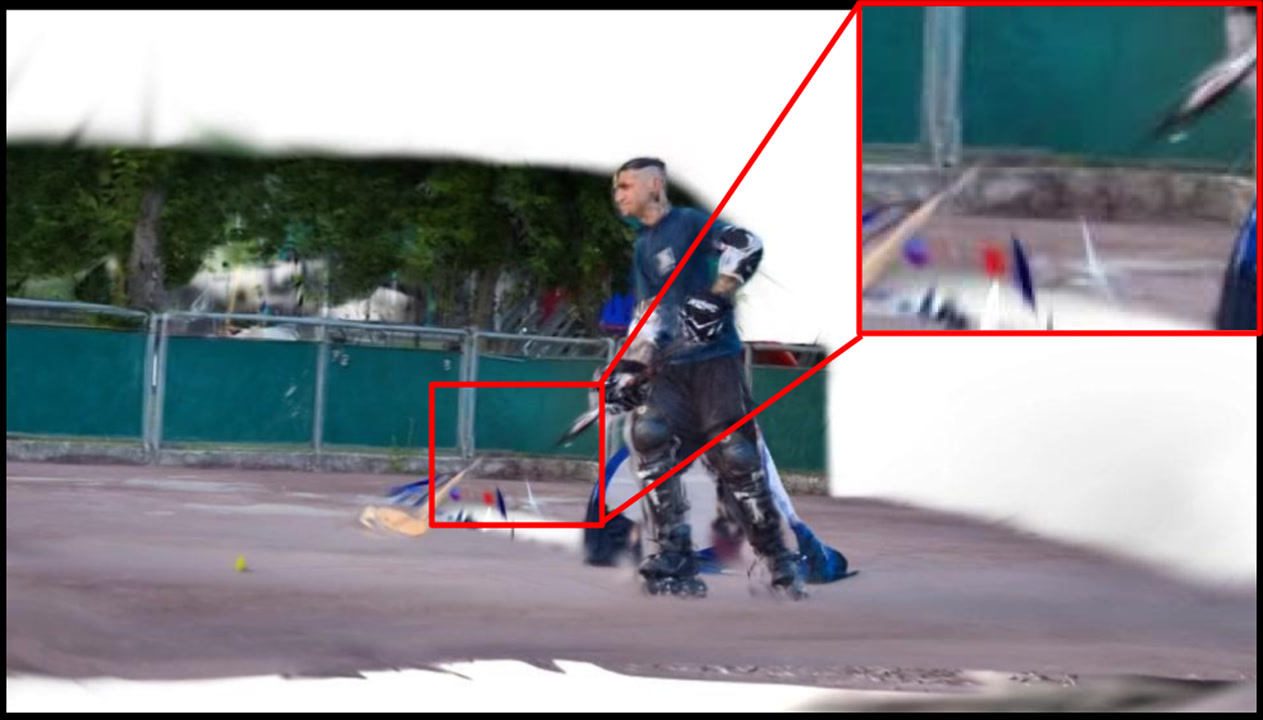}\hfill
  \includegraphics[width=0.495\linewidth]{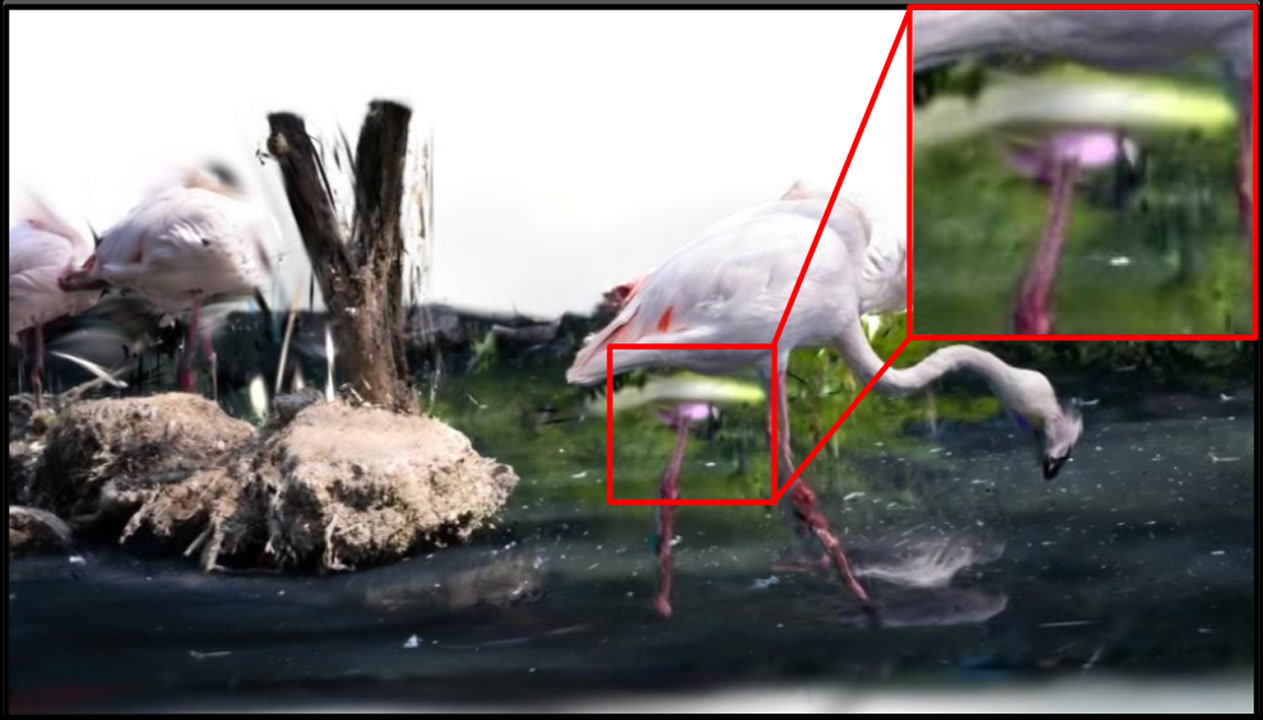}
  \caption{Novel view synthesis w/o Depth Refinement}
\end{subfigure}


\begin{subfigure}[t]{\linewidth}\centering
  \includegraphics[width=0.495\linewidth]{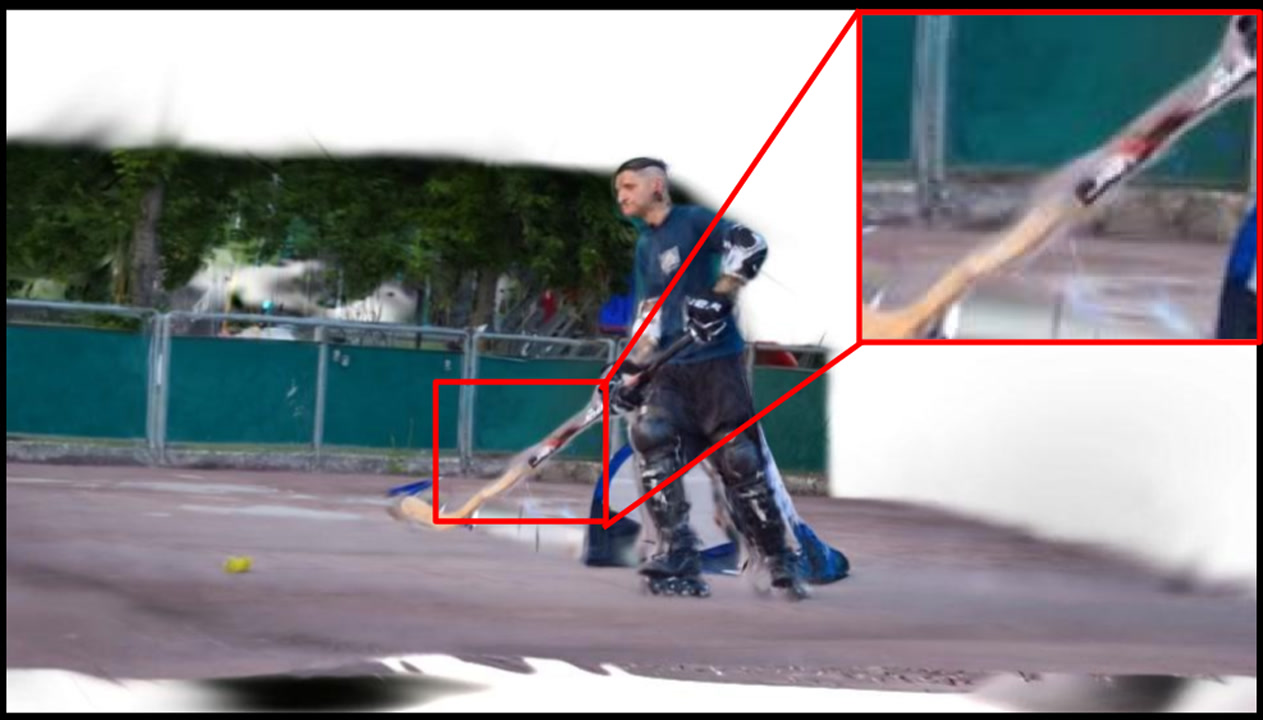}\hfill
  \includegraphics[width=0.495\linewidth]{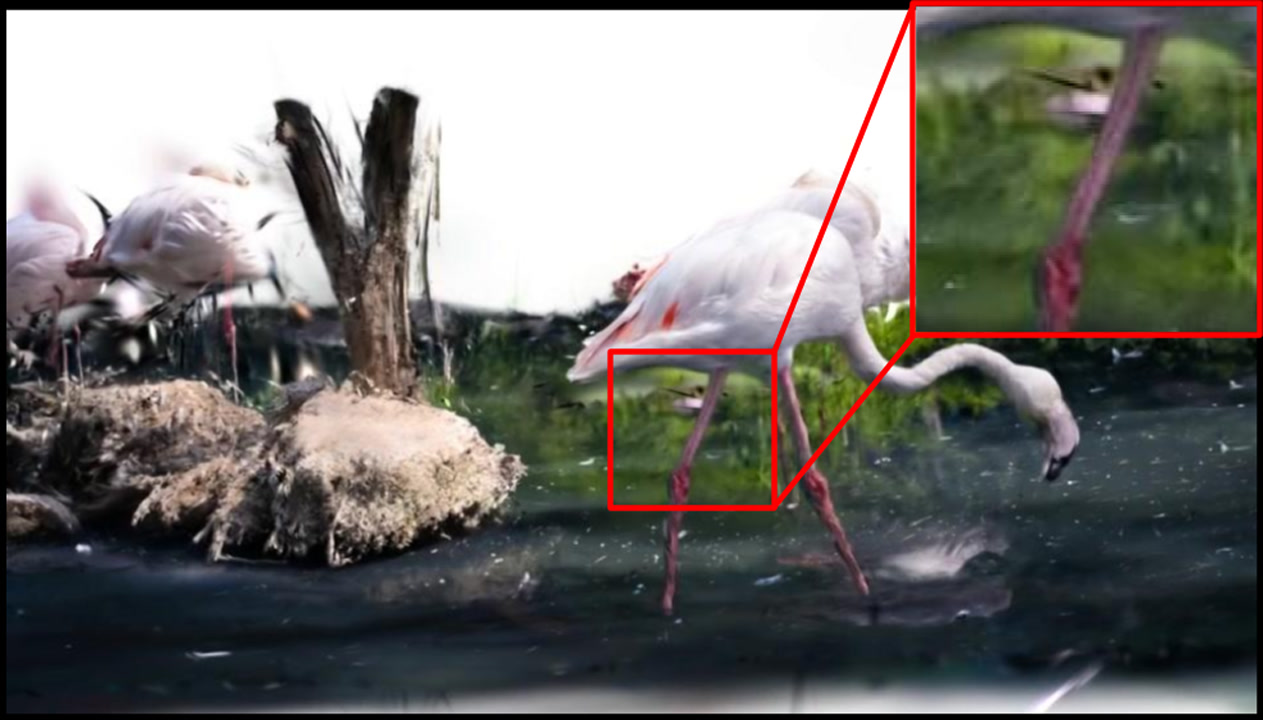}
  \caption{Novel view synthesis w/ Depth Refinement}
\end{subfigure}

\caption{Effect of depth refinement on novel view synthesis.}
\label{fig:qual_depth_refine_recon}
\end{figure}

\begin{figure}[h]
\centering
\begin{subfigure}[t]{0.495\linewidth}\centering
  \includegraphics[width=\linewidth]{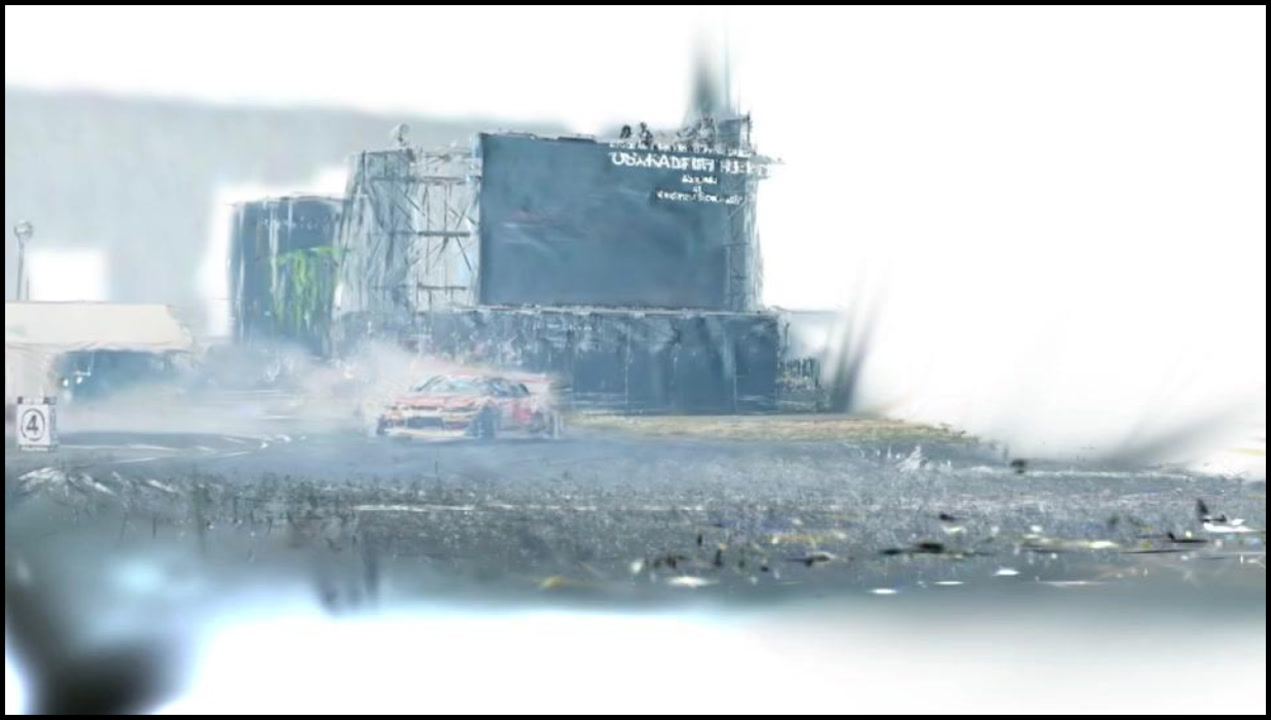}
  \caption{w/o $L^{\text{virtual}}_{\text{depth}}$}
\end{subfigure}\hfill
\begin{subfigure}[t]{0.495\linewidth}\centering
  \includegraphics[width=\linewidth]{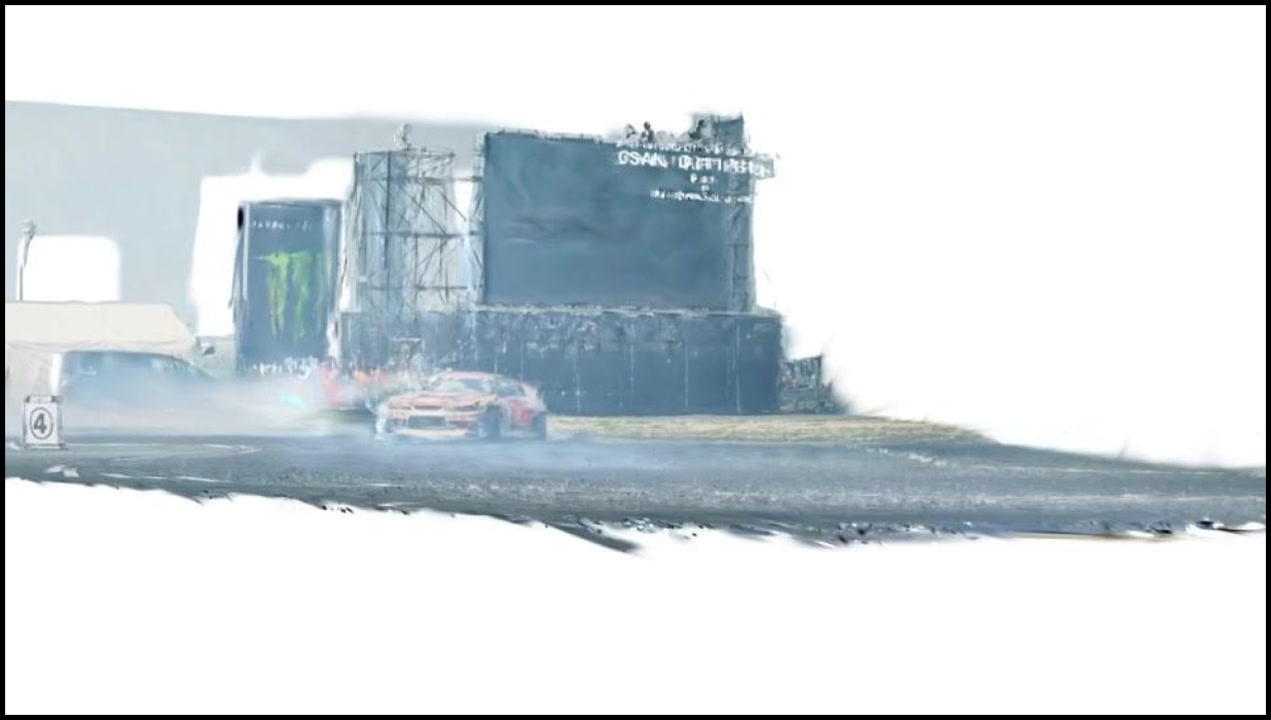}
  \caption{w/ $L^{\text{virtual}}_{\text{depth}}$}
\end{subfigure}

\caption{Qualitative comparison of virtual-view depth loss
$L^{\text{virtual}}_{\text{depth}}$.}
\label{fig:qual_virtual_view_depth_loss}
\end{figure}

We ablate the three main ingredients of our pipeline on the \emph{pose-free} iPhone DyCheck setting (see Tab.~\ref{tab:ablation_study}) and also show representative visual effects on DAVIS (see  Fig.~\ref{fig:qual_skeleton}, 
~\ref{fig:qual_track_reid},
~\ref{fig:qual_scaffold_loss},
~\ref{fig:qual_local_align}, 
~\ref{fig:qual_depth_refine_recon}, and
~\ref{fig:qual_virtual_view_depth_loss}).


\paragraph{Mask-Guided Point Tracker.}
Omitting this module's \textbf{skeleton sampling} leaves thin regions badly under-sampled.
In Fig.~\ref{fig:qual_skeleton}, the rhino’s fore-leg is almost lost, whereas the full model places ample tracks along the limb and recovers both geometry and texture.
Long-range tracks also break at occlusions. This module's \textbf{track re-ID} step, guided by the dynamic masks, stitches them back together. Without it, the walker’s left leg (Fig.~\ref{fig:qual_track_reid}) vanishes while briefly hidden behind the right, and the reconstruction fails.
Removing this module overall causes a perceptible drop in fidelity (see Table~\ref{tab:ablation_study}).





\paragraph{Scaffold-track loss $L_{\text{track}}^{\text{scaffold}}$.}
This loss anchors scaffold nodes to the 2-D tracks.
Without it, the ARAP term $L_\mathrm{arap}$ pulls nodes on thin moving parts together, producing artifacts around leg in Fig.~\ref{fig:qual_scaffold_loss} (a).
While the numerical gap in Table~\ref{tab:ablation_study} is small, visual continuity improves markedly with this constraint.

\paragraph{Virtual-view depth loss $L_{\text{depth}}^{\text{virtual}}$.}
Without this loss, floaters can proliferate: in Fig.~\ref{fig:qual_virtual_view_depth_loss} the floor and building facade fragment into streaks.
Introducing $L_{\text{depth}}^{\text{virtual}}$ removes these artifacts, yielding a clean result. Consistently, LPIPS rises without this loss as shown in Table~\ref{tab:ablation_study}.

Furthermore, our ~\textbf{depth refinement} step raises depth quality, recovers the true geometry of thin structures, and markedly improves the final reconstruction (see Fig.~\ref{fig:qual_local_align} and Fig.~\ref{fig:qual_depth_refine_recon}).
\vspace{-3mm}
\section{Conclusions}
\label{sec:conclusions}
This work demonstrates that stronger \emph{priors} can markedly boost
the performance of Dynamic Gaussian Splatting.  By extracting salient
object masks, refining video depth, and building reliable 2-D tracks, we
supply the Gaussian cloud and its motion scaffold with much richer
supervision.  Two complementary losses introduced in
Sec.~\ref{sec:dyn_recon} further suppress floaters and preserve
coherence on thin, fast-moving parts.  The resulting pipeline is fully
automatic and surpasses previous monocular DGS methods.

\begin{figure}[h]
\centering
\begin{subfigure}[t]{0.495\linewidth}\centering
  \includegraphics[width=\linewidth]{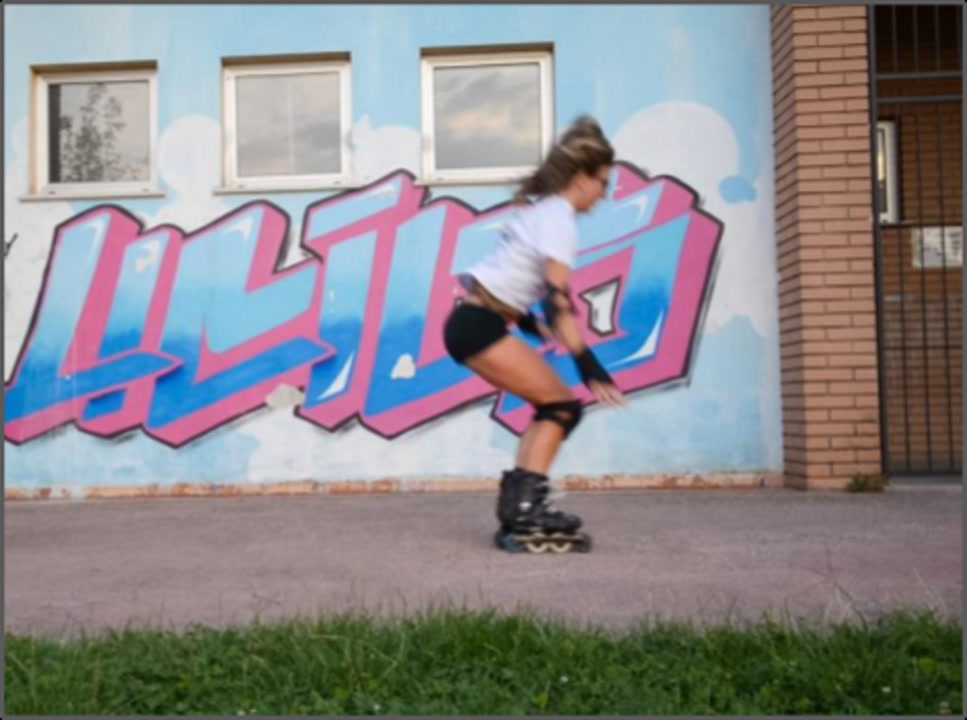}
  \caption{Monocular video}
\end{subfigure}\hfill
\begin{subfigure}[t]{0.495\linewidth}\centering
  \includegraphics[width=\linewidth]{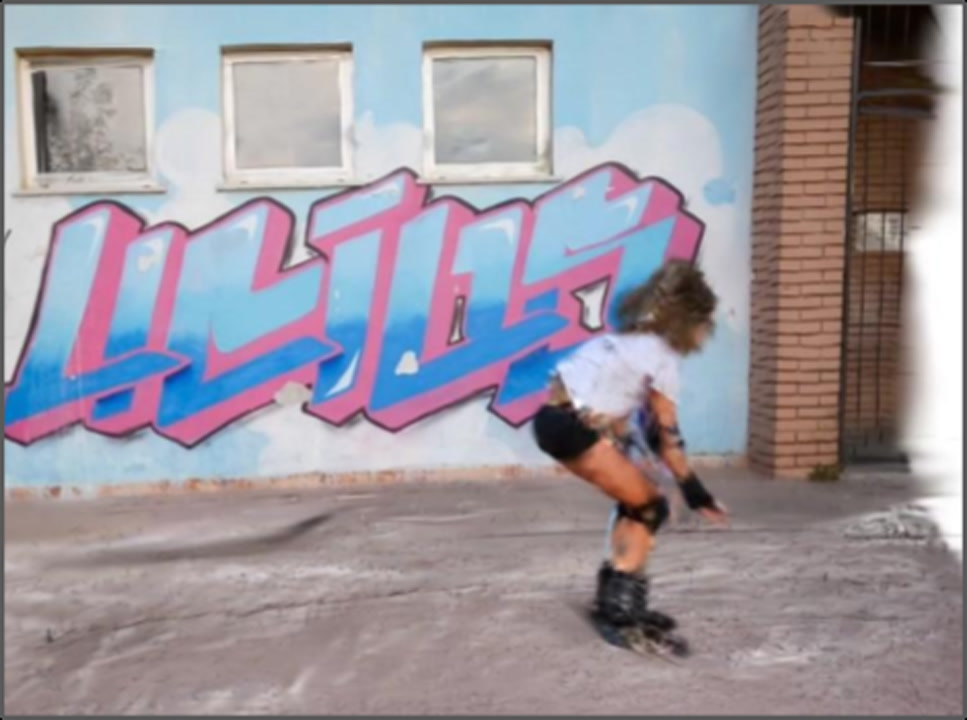}
  \caption{Novel view synthesis}
\end{subfigure}

\caption{\textbf{Failure case.} Our system copies motion- or focus-blur from the input video, so blur visible in the monocular frame (a) persists in the synthesized novel view (b).}
\label{fig:failure}
\end{figure}

\paragraph{Limitations.}
Our system copies motion- or focus-blur from the input video (see Fig.~\ref{fig:failure}) and leaves unseen regions empty. Integrating a video generative model could deblur the frames and hallucinate plausible content for those gaps.

\begin{acks}
This work was supported by Lenovo and the UW Reality Lab.
\end{acks}

\FloatBarrier        
\clearpage           
\bibliographystyle{ACM-Reference-Format}
\bibliography{main}

@String(CVPR= {IEEE Conf. Comput. Vis. Pattern Recog.})

@String(ICCV= {Int. Conf. Comput. Vis.})

@String(ECCV= {Eur. Conf. Comput. Vis.})

@String(TOG= {ACM Trans. Graph.})

@String(CVPR  = {CVPR})

@String(ICCV  = {ICCV})

@String(ECCV  = {ECCV})

@String(TOG   = {ACM TOG})

@inproceedings{chen2022geoaug,
  title={Geoaug: Data augmentation for few-shot nerf with geometry constraints},
  author={Chen, Di and Liu, Yu and Huang, Lianghua and Wang, Bin and Pan, Pan},
  booktitle={European Conference on Computer Vision},
  pages={322--337},
  year={2022},
  organization={Springer}
}

@inproceedings{niemeyer2022regnerf,
  title={Regnerf: Regularizing neural radiance fields for view synthesis from sparse inputs},
  author={Niemeyer, Michael and Barron, Jonathan T and Mildenhall, Ben and Sajjadi, Mehdi SM and Geiger, Andreas and Radwan, Noha},
  booktitle={Proceedings of the IEEE/CVF conference on computer vision and pattern recognition},
  pages={5480--5490},
  year={2022}
}

@inproceedings{jain2021putting,
  title={Putting nerf on a diet: Semantically consistent few-shot view synthesis},
  author={Jain, Ajay and Tancik, Matthew and Abbeel, Pieter},
  booktitle={Proceedings of the IEEE/CVF International Conference on Computer Vision},
  pages={5885--5894},
  year={2021}
}

@inproceedings{truong2023sparf,
  title={Sparf: Neural radiance fields from sparse and noisy poses},
  author={Truong, Prune and Rakotosaona, Marie-Julie and Manhardt, Fabian and Tombari, Federico},
  booktitle={Proceedings of the IEEE/CVF Conference on Computer Vision and Pattern Recognition},
  pages={4190--4200},
  year={2023}
}

@article{yin2024fewviewgs,
  title={FewViewGS: Gaussian Splatting with Few View Matching and Multi-stage Training},
  author={Yin, Ruihong and Yugay, Vladimir and Li, Yue and Karaoglu, Sezer and Gevers, Theo},
  journal={arXiv preprint arXiv:2411.02229},
  year={2024}
}

@article{ye2025gsplat,
  title={gsplat: An open-source library for Gaussian splatting},
  author={Ye, Vickie and Li, Ruilong and Kerr, Justin and Turkulainen, Matias and Yi, Brent and Pan, Zhuoyang and Seiskari, Otto and Ye, Jianbo and Hu, Jeffrey and Tancik, Matthew and Angjoo Kanazawa},
  journal={Journal of Machine Learning Research},
  volume={26},
  number={34},
  pages={1--17},
  year={2025}
}

@inproceedings{doersch2024bootstap,
  title={Bootstap: Bootstrapped training for tracking-any-point},
  author={Doersch, Carl and Luc, Pauline and Yang, Yi and Gokay, Dilara and Koppula, Skanda and Gupta, Ankush and Heyward, Joseph and Rocco, Ignacio and Goroshin, Ross and Carreira, Jo{\~a}o and others},
  booktitle={Proceedings of the Asian Conference on Computer Vision},
  pages={3257--3274},
  year={2024}
}

@software{AutoSeg_SAM2,
  author = {Zrporz},
  title = {AutoSeg-SAM2},
  year = {2024},
  publisher = {GitHub},
  url = {https://github.com/zrporz/AutoSeg-SAM2},
  version = {Latest}, 
  license = {MIT}, 
  note = {Automated image segmentation tool based on Segment Anything Model (SAM)}
}

@inproceedings{teed2020raft,
  title={Raft: Recurrent all-pairs field transforms for optical flow},
  author={Teed, Zachary and Deng, Jia},
  booktitle={Computer Vision--ECCV 2020: 16th European Conference, Glasgow, UK, August 23--28, 2020, Proceedings, Part II 16},
  pages={402--419},
  year={2020},
  organization={Springer}
}

@article{yang2023real,
  title={Real-time photorealistic dynamic scene representation and rendering with 4d gaussian splatting},
  author={Yang, Zeyu and Yang, Hongye and Pan, Zijie and Zhu, Xiatian and Zhang, Li},
  journal={arXiv preprint arXiv:2310.10642},
  year={2023}
}

@article{kratimenos2023dynmf,
  title={DynMF: Neural Motion Factorization for Real-time Dynamic View Synthesis with 3D Gaussian Splatting},
  author={Kratimenos, Agelos and Lei, Jiahui and Daniilidis, Kostas},
  journal={arXiv preprint arXiv:2312.00112},
  year={2023}
}

@article{das2023neural,
  title={Neural Parametric Gaussians for Monocular Non-Rigid Object Reconstruction},
  author={Das, Devikalyan and Wewer, Christopher and Yunus, Raza and Ilg, Eddy and Lenssen, Jan Eric},
  journal={arXiv preprint arXiv:2312.01196},
  year={2023}
}

@article{liang2023gaufre,
  title={GauFRe: Gaussian Deformation Fields for Real-time Dynamic Novel View Synthesis},
  author={Liang, Yiqing and Khan, Numair and Li, Zhengqin and Nguyen-Phuoc, Thu and Lanman, Douglas and Tompkin, James and Xiao, Lei},
  journal={arXiv preprint arXiv:2312.11458},
  year={2023}
}

@inproceedings{li2023dynibar,
  title={Dynibar: Neural dynamic image-based rendering},
  author={Li, Zhengqi and Wang, Qianqian and Cole, Forrester and Tucker, Richard and Snavely, Noah},
  booktitle={Proceedings of the IEEE/CVF Conference on Computer Vision and Pattern Recognition},
  pages={4273--4284},
  year={2023}
}

@inproceedings{li2021neural,
  title={Neural scene flow fields for space-time view synthesis of dynamic scenes},
  author={Li, Zhengqi and Niklaus, Simon and Snavely, Noah and Wang, Oliver},
  booktitle={Proceedings of the IEEE/CVF Conference on Computer Vision and Pattern Recognition},
  pages={6498--6508},
  year={2021}
}

@inproceedings{gao2021dynamic,
  title={Dynamic view synthesis from dynamic monocular video},
  author={Gao, Chen and Saraf, Ayush and Kopf, Johannes and Huang, Jia-Bin},
  booktitle={Proceedings of the IEEE/CVF International Conference on Computer Vision},
  pages={5712--5721},
  year={2021}
}

@inproceedings{liu2023robust,
  author    = {Liu, Yu-Lun and Gao, Chen and Meuleman, Andreas and Tseng, Hung-Yu and Saraf, Ayush and Kim, Changil and Chuang, Yung-Yu and Kopf, Johannes and Huang, Jia-Bin},
  title     = {Robust Dynamic Radiance Fields},
  booktitle = {Proceedings of the IEEE/CVF Conference on Computer Vision and Pattern Recognition},
  year      = {2023}
}

@inproceedings{du2021neural,
  title={Neural radiance flow for 4d view synthesis and video processing},
  author={Du, Yilun and Zhang, Yinan and Yu, Hong-Xing and Tenenbaum, Joshua B and Wu, Jiajun},
  booktitle={2021 IEEE/CVF International Conference on Computer Vision (ICCV)},
  pages={14304--14314},
  year={2021},
  organization={IEEE Computer Society}
}

@article{lin2023im4d,
  title={Im4D: High-Fidelity and Real-Time Novel View Synthesis for Dynamic Scenes},
  author={Lin, Haotong and Peng, Sida and Xu, Zhen and Xie, Tao and He, Xingyi and Bao, Hujun and Zhou, Xiaowei},
  journal={arXiv preprint arXiv:2310.08585},
  year={2023}
}

@article{song2023nerfplayer,
  title={Nerfplayer: A streamable dynamic scene representation with decomposed neural radiance fields},
  author={Song, Liangchen and Chen, Anpei and Li, Zhong and Chen, Zhang and Chen, Lele and Yuan, Junsong and Xu, Yi and Geiger, Andreas},
  journal={IEEE Transactions on Visualization and Computer Graphics},
  volume={29},
  number={5},
  pages={2732--2742},
  year={2023},
  publisher={IEEE}
}

@inproceedings{wang2022fourier,
  title={Fourier plenoctrees for dynamic radiance field rendering in real-time},
  author={Wang, Liao and Zhang, Jiakai and Liu, Xinhang and Zhao, Fuqiang and Zhang, Yanshun and Zhang, Yingliang and Wu, Minye and Yu, Jingyi and Xu, Lan},
  booktitle={Proceedings of the IEEE/CVF Conference on Computer Vision and Pattern Recognition},
  pages={13524--13534},
  year={2022}
}

@article{kerbl20233d,
  title={3D Gaussian Splatting for Real-Time Radiance Field Rendering},
  author={Kerbl, Bernhard and Kopanas, Georgios and Leimk{\"u}hler, Thomas and Drettakis, George},
  journal={ACM Transactions on Graphics},
  volume={42},
  number={4},
  year={2023}
}

@inproceedings{chan2022efficient,
  title={Efficient geometry-aware 3D generative adversarial networks},
  author={Chan, Eric R and Lin, Connor Z and Chan, Matthew A and Nagano, Koki and Pan, Boxiao and De Mello, Shalini and Gallo, Orazio and Guibas, Leonidas J and Tremblay, Jonathan and Khamis, Sameh and others},
  booktitle=CVPR,
  year={2022}
}

@article{wu2023four,
  title={4D Gaussian Splatting for Real-Time Dynamic Scene Rendering},
  author={Wu, Guanjun and Yi, Taoran and Fang, Jiemin and Xie, Lingxi and Zhang, Xiaopeng and Wei, Wei and Liu, Wenyu and Tian, Qi and Wang, Xinggang},
  journal={arXiv preprint arXiv:2310.08528},
  year={2023}
}

@article{attal2023hyperreel,
  title={HyperReel: High-Fidelity 6-DoF Video with Ray-Conditioned Sampling},
  author={Attal, Benjamin and Huang, Jia-Bin and Richardt, Christian and Zollhoefer, Michael and Kopf, Johannes and O'Toole, Matthew and Kim, Changil},
  journal={arXiv preprint arXiv:2301.02238},
  year={2023}
}

@article{bansal2020four,
  title={4D Visualization of Dynamic Events from Unconstrained Multi-View Videos},
  author={Bansal, Aayush and Vo, Minh and Sheikh, Yaser and Ramanan, Deva and Narasimhan, Srinivasa},
  journal={arXiv preprint arXiv:2005.13532},
  year={2020}
}

@article{cao2023hexplane,
  title={HexPlane: A Fast Representation for Dynamic Scenes},
  author={Cao, Ang and Johnson, Justin},
  journal={arXiv preprint arXiv:2301.09632},
  year={2023}
}

@article{fridovich2023kplanes,
  title={K-Planes: Explicit Radiance Fields in Space, Time, and Appearance},
  author={Fridovich-Keil, Sara and Meanti, Giacomo and Warburg, Frederik and Recht, Benjamin and Kanazawa, Angjoo},
  journal={arXiv preprint arXiv:2301.10241},
  year={2023}
}

@article{lombardi2019neural,
  title={Neural Volumes: Learning Dynamic Renderable Volumes from Images},
  author={Lombardi, Stephen and Simon, Tomas and Saragih, Jason and Schwartz, Gabriel and Lehrmann, Andreas and Sheikh, Yaser},
  journal={arXiv preprint arXiv:2011.13961},
  year={2019}
}

@article{luiten2023dynamic,
  title={Dynamic 3D Gaussians: Tracking by Persistent Dynamic View Synthesis},
  author={Luiten, Jonathon and Kopanas, Georgios and Leibe, Bastian and Ramanan, Deva},
  journal={arXiv preprint arXiv:2308.09713},
  year={2023}
}

@article{lei2024mosca,
  title={MoSca: Dynamic Gaussian Fusion from Casual Videos via 4D Motion Scaffolds},
  author={Lei, Jiahui and Weng, Yijia and Harley, Adam and Guibas, Leonidas and Daniilidis, Kostas},
  journal={arXiv preprint arXiv:2405.17421},
  year={2024}
}

@article{wang2024shape,
  title={Shape of Motion: 4D Reconstruction from a Single Video},
  author={Wang, Qianqian and Ye, Vickie and Gao, Hang and Austin, Jake and Li, Zhengqi and Kanazawa, Angjoo},
  journal={arXiv preprint arXiv:2407.13764},
  year={2024}
}

@article{park2024splinegs,
  title={SplineGS: Robust Motion-Adaptive Spline for Real-Time Dynamic 3D Gaussians from Monocular Video},
  author={Park, Jongmin and Bui, Minh-Quan Viet and Gonzalez Bello, Juan Luis and Moon, Jaeho and Oh, Jihyong and Kim, Munchurl},
  journal={arXiv preprint arXiv:2412.09982},
  year={2024}
}

@article{kwak2025modec,
  title={MoDec-GS: Global-to-Local Motion Decomposition and Temporal Interval Adjustment for Compact Dynamic 3D Gaussian Splatting},
  author={Kwak, Sangwoon and Kim, Joonsoo and Jeong, Jun Young and Cheong, Won-Sik and Oh, Jihyong and Kim, Munchurl},
  journal={arXiv preprint arXiv:2501.03714},
  year={2025}
}

@article{liu2024modgs,
  title={MoDGS: Dynamic Gaussian Splatting from Casually-captured Monocular Videos},
  author={Liu, Qingming and Liu, Yuan and Wang, Jiepeng and Lyv, Xianqiang and Wang, Peng and Wang, Wenping and Hou, Junhui},
  journal={arXiv preprint arXiv:2406.00434},
  year={2024}
}

@article{stearns2024dynamic,
  title={Dynamic Gaussian Marbles for Novel View Synthesis of Casual Monocular Videos},
  author={Stearns, Colton and Harley, Adam and Uy, Mikaela and Dubost, Florian and Tombari, Federico and Wetzstein, Gordon and Guibas, Leonidas},
  journal={arXiv preprint arXiv:2406.18717},
  year={2024}
}

@article{zhao2024dynomo,
  title={DynOMo: Online Point Tracking by Dynamic Monocular Reconstruction},
  author={Zhao, Xiaoming and Wang, Yifan and Zhang, Yifan and Liu, Yebin},
  journal={arXiv preprint arXiv:2409.02104},
  year={2024}
}

@article{bui2023dyblurf,
  title={DyBluRF: Dynamic Deblurring Neural Radiance Fields for Blurry Monocular Video},
  author={Bui, Minh-Quan Viet and Park, Jongmin and Oh, Jihyong and Kim, Munchurl},
  journal={arXiv preprint arXiv:2312.13528},
  year={2023}
}

@article{zhao2023pseudo,
  title={Pseudo-Generalized Dynamic View Synthesis from a Video},
  author={Zhao, Xiaoming and Wang, Yifan and Zhang, Yifan and Liu, Yebin},
  journal={arXiv preprint arXiv:2310.08587},
  year={2023}
}

@article{zhang2023dynpoint,
  title={DynPoint: Dynamic Neural Point For View Synthesis},
  author={Zhang, Yifan and Wang, Yifan and Zhao, Xiaoming and Liu, Yebin},
  journal={arXiv preprint arXiv:2310.18999},
  year={2023}
}

@article{miao2024ctnerf,
  title={CTNeRF: Cross-Time Transformer for Dynamic Neural Radiance Field from Monocular Video},
  author={Miao, Xingyu and Bai, Yang and Duan, Haoran and Huang, Yawen and Wan, Fan and Long, Yang and Zheng, Yefeng},
  journal={arXiv preprint arXiv:2401.04861},
  year={2024}
}

@article{li2022layeredgs,
  title={LayeredGS: Efficient Dynamic Scene Rendering and Point Tracking with Multi-Layer Deformable Gaussian Splatting},
  author={Li, Zhengqi and Niklaus, Simon and Snavely, Noah and Wang, Oliver},
  journal={arXiv preprint arXiv:2211.11082},
  year={2022}
}

@article{li2020neural,
  title={Neural Scene Flow Fields for Space-Time View Synthesis of Dynamic Scenes},
  author={Li, Zhengqi and Niklaus, Simon and Snavely, Noah and Wang, Oliver},
  journal={arXiv preprint arXiv:2011.12950},
  year={2020}
}

@article{li2023fast,
  title={Fast View Synthesis of Casual Videos with Soup-of-Planes},
  author={Li, Zhengqi and Niklaus, Simon and Snavely, Noah and Wang, Oliver},
  journal={arXiv preprint arXiv:2304.01716},
  year={2023}
}

@article{you2023decoupling,
  title={Decoupling Dynamic Monocular Videos for Dynamic View Synthesis},
  author={You, Meng and Guo, Mantang and Lyu, Xianqiang and Liu, Hui and Hou, Junhui},
  journal={arXiv preprint arXiv:2304.01716},
  year={2023}
}

@inproceedings{shih2024modeling,
  title={Modeling ambient scene dynamics for free-view synthesis},
  author={Shih, Meng-Li and Huang, Jia-Bin and Kim, Changil and Shah, Rajvi and Kopf, Johannes and Gao, Chen},
  booktitle={ACM SIGGRAPH 2024 Conference Papers},
  pages={1--11},
  year={2024}
}

@article{li2024megasam,
  title={MegaSaM: Accurate, Fast, and Robust Structure and Motion from Casual Dynamic Videos},
  author={Li, Zhengqi and Tucker, Richard and Cole, Forrester and Wang, Qianqian and Jin, Linyi and Ye, Vickie and Kanazawa, Angjoo and Holynski, Aleksander and Snavely, Noah},
  journal={arXiv preprint arXiv:2412.04463},
  year={2024}
}

@article{kopf2020robust,
  title={Robust Consistent Video Depth Estimation},
  author={Kopf, Johannes and Rong, Xuejian and Huang, Jia-Bin},
  journal={arXiv preprint arXiv:2012.05901},
  year={2020}
}

@article{zhang2024monst3r,
  title={MonST3R: A Simple Approach for Estimating Geometry in the Presence of Motion},
  author={Zhang, Junyi and Herrmann, Charles and Hur, Junhwa and Jampani, Varun and Darrell, Trevor and Cole, Forrester and Sun, Deqing and Yang, Ming-Hsuan},
  journal={arXiv preprint arXiv:2410.03825},
  year={2024}
}

@article{wang2025cut3r,
  title={Continuous 3D Perception Model with Persistent State},
  author={Wang, Qianqian and Zhang, Yifei and Holynski, Aleksander and Efros, Alexei A. and Kanazawa, Angjoo},
  journal={arXiv preprint arXiv:2501.12387},
  year={2025}
}

@article{feng2025st4rtrack,
  title={St4RTrack: Simultaneous 4D Reconstruction and Tracking in the World},
  author={Feng, Haiwen and Zhang, Junyi and Wang, Qianqian and Ye, Yufei and Yu, Pengcheng and Black, Michael J. and Darrell, Trevor and Kanazawa, Angjoo},
  journal={arXiv preprint arXiv:2504.13152},
  year={2025}
}

@inproceedings{10.1007/978-3-031-19827-4_2,
author = {Zhang, Zhoutong and Cole, Forrester and Li, Zhengqi and Rubinstein, Michael and Snavely, Noah and Freeman, William T.},
title = {Structure and Motion from Casual Videos},
year = {2022},
isbn = {978-3-031-19826-7},
publisher = {Springer-Verlag},
address = {Berlin, Heidelberg},
url = {https://doi.org/10.1007/978-3-031-19827-4_2},
doi = {10.1007/978-3-031-19827-4_2},
booktitle = {Computer Vision – ECCV 2022: 17th European Conference, Tel Aviv, Israel, October 23–27, 2022, Proceedings, Part XXXIII},
pages = {20–37},
numpages = {18},
location = {Tel Aviv, Israel}
}

@article{ravi2024sam,
  title={Sam 2: Segment anything in images and videos},
  author={Ravi, Nikhila and Gabeur, Valentin and Hu, Yuan-Ting and Hu, Ronghang and Ryali, Chaitanya and Ma, Tengyu and Khedr, Haitham and R{\"a}dle, Roman and Rolland, Chloe and Gustafson, Laura and others},
  journal={arXiv preprint arXiv:2408.00714},
  year={2024}
}

@inproceedings{yang2024depth,
  title={Depth anything: Unleashing the power of large-scale unlabeled data},
  author={Yang, Lihe and Kang, Bingyi and Huang, Zilong and Xu, Xiaogang and Feng, Jiashi and Zhao, Hengshuang},
  booktitle={Proceedings of the IEEE/CVF Conference on Computer Vision and Pattern Recognition},
  pages={10371--10381},
  year={2024}
}

@article{wang2024moge,
  title={Moge: Unlocking accurate monocular geometry estimation for open-domain images with optimal training supervision},
  author={Wang, Ruicheng and Xu, Sicheng and Dai, Cassie and Xiang, Jianfeng and Deng, Yu and Tong, Xin and Yang, Jiaolong},
  journal={arXiv preprint arXiv:2410.19115},
  year={2024}
}

@article{hu2024depthcrafter,
  title={Depthcrafter: Generating consistent long depth sequences for open-world videos},
  author={Hu, Wenbo and Gao, Xiangjun and Li, Xiaoyu and Zhao, Sijie and Cun, Xiaodong and Zhang, Yong and Quan, Long and Shan, Ying},
  journal={arXiv preprint arXiv:2409.02095},
  year={2024}
}

@article{harley2022particle,
  title={Particle Video Revisited: Tracking Through Occlusions Using Point Trajectories},
  author={Harley, Adam W. and Fang, Zhaoyuan and Fragkiadaki, Katerina},
  journal={arXiv preprint arXiv:2204.04153},
  year={2022}
}

@article{doersch2022tap,
  title={TAP-Vid: A Benchmark for Tracking Any Point in a Video},
  author={Doersch, Carl and Gupta, Ankush and Markeeva, Larisa and Recasens, Adrià and Smaira, Lucas and Aytar, Yusuf and Carreira, João and Zisserman, Andrew and Yang, Yi},
  journal={arXiv preprint arXiv:2211.03726},
  year={2022}
}

@article{doersch2023tapir,
  title={TAPIR: Tracking Any Point with per-frame Initialization and temporal Refinement},
  author={Doersch, Carl and Yang, Yi and Vecerik, Mel and Gokay, Dilara and Gupta, Ankush and Aytar, Yusuf and Carreira, Joao and Zisserman, Andrew},
  journal={arXiv preprint arXiv:2306.08637},
  year={2023}
}

@article{karaev2023cotracker,
  title={CoTracker: It is Better to Track Together},
  author={Karaev, Nikita and Rocco, Ignacio and Graham, Benjamin and Neverova, Natalia and Vedaldi, Andrea and Rupprecht, Christian},
  journal={arXiv preprint arXiv:2307.07635},
  year={2023}
}

@article{xiao2024spatialtracker,
  title={SpatialTracker: Tracking Any 2D Pixels in 3D Space},
  author={Xiao, Yuxi and Wang, Qianqian and Zhang, Shangzhan and Xue, Nan and Peng, Sida and Shen, Yujun and Zhou, Xiaowei},
  journal={arXiv preprint arXiv:2404.04319},
  year={2024}
}

@article{kirillov2023segment,
  title={Segment Anything},
  author={Kirillov, Alexander and Mintun, Eric and Ravi, Nikhila and Mao, Hanzi and Rolland, Chloe and Gustafson, Laura and Xiao, Tete and Whitehead, Spencer and Berg, Alexander C and Lo, Wan-Yen and Doll{\'a}r, Piotr and Girshick, Ross},
  journal={arXiv preprint arXiv:2304.02643},
  year={2023}
}

@article{ravi2024sam2,
  title={SAM 2: Segment Anything in Images and Videos},
  author={Ravi, Nikhila and Gabeur, Valentin and Hu, Yuan-Ting and Hu, Ronghang and Ryali, Chaitanya and Ma, Tengyu and Khedr, Haitham and R{\"a}dle, Roman and Rolland, Chloe and Gustafson, Laura and Mintun, Eric and Pan, Junting and Alwala, Kalyan Vasudev and Carion, Nicolas and Wu, Chao-Yuan and Girshick, Ross and Doll{\'a}r, Piotr and Feichtenhofer, Christoph},
  journal={arXiv preprint arXiv:2408.00714},
  year={2024}
}

@article{yang2023track,
  title={Track Anything: Segment Anything Meets Videos},
  author={Yang, Jinyu and Gao, Mingqi and Li, Zhe and Gao, Shang and Wang, Fangjing and Zheng, Feng},
  journal={arXiv preprint arXiv:2304.11968},
  year={2023}
}

@article{huang2025segment,
  title={Segment Any Motion in Videos},
  author={Huang, Nan and Zheng, Wenzhao and Xu, Chenfeng and Keutzer, Kurt and Zhang, Shanghang and Kanazawa, Angjoo and Wang, Qianqian},
  journal={arXiv preprint arXiv:2503.22268},
  year={2025}
}

@article{zhang2025motion,
  title={Motion Blender Gaussian Splatting for Dynamic Reconstruction},
  author={Zhang, Xinyu and Chang, Haonan and Liu, Yuhan and Boularias, Abdeslam},
  journal={arXiv preprint arXiv:2503.09040},
  year={2025}
}

@article{xu2024representing,
  title={Representing long volumetric video with temporal gaussian hierarchy},
  author={Xu, Zhen and Xu, Yinghao and Yu, Zhiyuan and Peng, Sida and Sun, Jiaming and Bao, Hujun and Zhou, Xiaowei},
  journal={ACM Transactions on Graphics (TOG)},
  volume={43},
  number={6},
  pages={1--18},
  year={2024},
  publisher={ACM New York, NY, USA}
}

@inproceedings{yan20244d,
  title={4D Gaussian Splatting with Scale-aware Residual Field and Adaptive Optimization for Real-time rendering of temporally complex dynamic scenes},
  author={Yan, Jinbo and Peng, Rui and Tang, Luyang and Wang, Ronggang},
  booktitle={Proceedings of the 32nd ACM International Conference on Multimedia},
  pages={7871--7880},
  year={2024}
}

@article{liang2024feed,
  title={Feed-Forward Bullet-Time Reconstruction of Dynamic Scenes from Monocular Videos},
  author={Liang, Hanxue and Ren, Jiawei and Mirzaei, Ashkan and Torralba, Antonio and Liu, Ziwei and Gilitschenski, Igor and Fidler, Sanja and Oztireli, Cengiz and Ling, Huan and Gojcic, Zan and others},
  journal={arXiv preprint arXiv:2412.03526},
  year={2024}
}

@article{wu2024cat4d,
  title={Cat4d: Create anything in 4d with multi-view video diffusion models},
  author={Wu, Rundi and Gao, Ruiqi and Poole, Ben and Trevithick, Alex and Zheng, Changxi and Barron, Jonathan T and Holynski, Aleksander},
  journal={arXiv preprint arXiv:2411.18613},
  year={2024}
}

@article{lee2024fully,
  title={Fully explicit dynamic gaussian splatting},
  author={Lee, Junoh and Won, ChangYeon and Jung, Hyunjun and Bae, Inhwan and Jeon, Hae-Gon},
  journal={Advances in Neural Information Processing Systems},
  volume={37},
  pages={5384--5409},
  year={2024}
}

@article{zhu2024motiongs,
  title={Motiongs: Exploring explicit motion guidance for deformable 3d gaussian splatting},
  author={Zhu, Ruijie and Liang, Yanzhe and Chang, Hanzhi and Deng, Jiacheng and Lu, Jiahao and Yang, Wenfei and Zhang, Tianzhu and Zhang, Yongdong},
  journal={Advances in Neural Information Processing Systems},
  volume={37},
  pages={101790--101817},
  year={2024}
}

@inproceedings{bae2024per,
  title={Per-gaussian embedding-based deformation for deformable 3d gaussian splatting},
  author={Bae, Jeongmin and Kim, Seoha and Yun, Youngsik and Lee, Hahyun and Bang, Gun and Uh, Youngjung},
  booktitle={European Conference on Computer Vision},
  pages={321--335},
  year={2024},
  organization={Springer}
}

@inproceedings{sun20243dgstream,
  title={3dgstream: On-the-fly training of 3d gaussians for efficient streaming of photo-realistic free-viewpoint videos},
  author={Sun, Jiakai and Jiao, Han and Li, Guangyuan and Zhang, Zhanjie and Zhao, Lei and Xing, Wei},
  booktitle={Proceedings of the IEEE/CVF Conference on Computer Vision and Pattern Recognition},
  pages={20675--20685},
  year={2024}
}

@inproceedings{duan20244d,
  title={4d-rotor gaussian splatting: towards efficient novel view synthesis for dynamic scenes},
  author={Duan, Yuanxing and Wei, Fangyin and Dai, Qiyu and He, Yuhang and Chen, Wenzheng and Chen, Baoquan},
  booktitle={ACM SIGGRAPH 2024 Conference Papers},
  pages={1--11},
  year={2024}
}

@inproceedings{li2024spacetime,
  title={Spacetime gaussian feature splatting for real-time dynamic view synthesis},
  author={Li, Zhan and Chen, Zhang and Li, Zhong and Xu, Yi},
  booktitle={Proceedings of the IEEE/CVF Conference on Computer Vision and Pattern Recognition},
  pages={8508--8520},
  year={2024}
}

@article{shaw2023swings,
  title={SWinGS: Sliding Windows for Dynamic 3D Gaussian Splatting},
  author={Shaw, Richard and Nazarczuk, Michal and Song, Jifei and Moreau, Arthur and Catley-Chandar, Sibi and Dhamo, Helisa and Perez-Pellitero, Eduardo},
  journal={arXiv preprint arXiv:2312.13308},
  year={2023}
}

@inproceedings{liang2025gaufre,
  title={Gaufre: Gaussian deformation fields for real-time dynamic novel view synthesis},
  author={Liang, Yiqing and Khan, Numair and Li, Zhengqin and Nguyen-Phuoc, Thu and Lanman, Douglas and Tompkin, James and Xiao, Lei},
  booktitle={2025 IEEE/CVF Winter Conference on Applications of Computer Vision (WACV)},
  pages={2642--2652},
  year={2025},
  organization={IEEE}
}

@article{wang2024diffusion,
  title={Diffusion priors for dynamic view synthesis from monocular videos},
  author={Wang, Chaoyang and Zhuang, Peiye and Siarohin, Aliaksandr and Cao, Junli and Qian, Guocheng and Lee, Hsin-Ying and Tulyakov, Sergey},
  journal={arXiv preprint arXiv:2401.05583},
  year={2024}
}

@inproceedings{huang2024sc,
  title={Sc-gs: Sparse-controlled gaussian splatting for editable dynamic scenes},
  author={Huang, Yi-Hua and Sun, Yang-Tian and Yang, Ziyi and Lyu, Xiaoyang and Cao, Yan-Pei and Qi, Xiaojuan},
  booktitle={Proceedings of the IEEE/CVF conference on computer vision and pattern recognition},
  pages={4220--4230},
  year={2024}
}

@inproceedings{kratimenos2024dynmf,
  title={Dynmf: Neural motion factorization for real-time dynamic view synthesis with 3d gaussian splatting},
  author={Kratimenos, Agelos and Lei, Jiahui and Daniilidis, Kostas},
  booktitle={European Conference on Computer Vision},
  pages={252--269},
  year={2024},
  organization={Springer}
}

@inproceedings{wu20244d,
  title={4d gaussian splatting for real-time dynamic scene rendering},
  author={Wu, Guanjun and Yi, Taoran and Fang, Jiemin and Xie, Lingxi and Zhang, Xiaopeng and Wei, Wei and Liu, Wenyu and Tian, Qi and Wang, Xinggang},
  booktitle={Proceedings of the IEEE/CVF conference on computer vision and pattern recognition},
  pages={20310--20320},
  year={2024}
}

@inproceedings{lin2024gaussian,
  title={Gaussian-flow: 4d reconstruction with dynamic 3d gaussian particle},
  author={Lin, Youtian and Dai, Zuozhuo and Zhu, Siyu and Yao, Yao},
  booktitle={Proceedings of the IEEE/CVF Conference on Computer Vision and Pattern Recognition},
  pages={21136--21145},
  year={2024}
}

@article{mildenhall2021nerf,
  title={Nerf: Representing scenes as neural radiance fields for view synthesis},
  author={Mildenhall, Ben and Srinivasan, Pratul P and Tancik, Matthew and Barron, Jonathan T and Ramamoorthi, Ravi and Ng, Ren},
  journal={Communications of the ACM},
  volume={65},
  number={1},
  pages={99--106},
  year={2021},
  publisher={ACM New York, NY, USA}
}

@article{tapip3d,
  title={TAPIP3D: Tracking Any Point in Persistent 3D Geometry},
  author={Zhang, Bowei and Ke, Lei and Harley, Adam W and Fragkiadaki, Katerina},
  journal={arXiv preprint arXiv:2504.14717},
  year={2025}
}

@inproceedings{fang2022fast,
  title={Fast dynamic radiance fields with time-aware neural voxels},
  author={Fang, Jiemin and Yi, Taoran and Wang, Xinggang and Xie, Lingxi and Zhang, Xiaopeng and Liu, Wenyu and Nie{\ss}ner, Matthias and Tian, Qi},
  booktitle={SIGGRAPH Asia 2022 Conference Papers},
  pages={1--9},
  year={2022}
}

@inproceedings{athar2022rignerf,
  title={Rignerf: Fully controllable neural 3d portraits},
  author={Athar, ShahRukh and Xu, Zexiang and Sunkavalli, Kalyan and Shechtman, Eli and Shu, Zhixin},
  booktitle={Proceedings of the IEEE/CVF conference on Computer Vision and Pattern Recognition},
  pages={20364--20373},
  year={2022}
}

@inproceedings{jiang2022neuman,
  title={Neuman: Neural human radiance field from a single video},
  author={Jiang, Wei and Yi, Kwang Moo and Samei, Golnoosh and Tuzel, Oncel and Ranjan, Anurag},
  booktitle={European Conference on Computer Vision},
  pages={402--418},
  year={2022},
  organization={Springer}
}

@inproceedings{yoon2020novel,
  title={Novel view synthesis of dynamic scenes with globally coherent depths from a monocular camera},
  author={Yoon, Jae Shin and Kim, Kihwan and Gallo, Orazio and Park, Hyun Soo and Kautz, Jan},
  booktitle={Proceedings of the IEEE/CVF Conference on Computer Vision and Pattern Recognition},
  pages={5336--5345},
  year={2020}
}

@inproceedings{piccinelli2024unidepth,
  title={UniDepth: Universal monocular metric depth estimation},
  author={Piccinelli, Luigi and Yang, Yung-Hsu and Sakaridis, Christos and Segu, Mattia and Li, Siyuan and Van Gool, Luc and Yu, Fisher},
  booktitle={Proceedings of the IEEE/CVF Conference on Computer Vision and Pattern Recognition},
  pages={10106--10116},
  year={2024}
}

@inproceedings{xian2021space,
  title={Space-time neural irradiance fields for free-viewpoint video},
  author={Xian, Wenqi and Huang, Jia-Bin and Kopf, Johannes and Kim, Changil},
  booktitle={Proceedings of the IEEE/CVF conference on computer vision and pattern recognition},
  pages={9421--9431},
  year={2021}
}

@inproceedings{yang2024deformable,
  title={Deformable 3d gaussians for high-fidelity monocular dynamic scene reconstruction},
  author={Yang, Ziyi and Gao, Xinyu and Zhou, Wen and Jiao, Shaohui and Zhang, Yuqing and Jin, Xiaogang},
  booktitle={Proceedings of the IEEE/CVF conference on computer vision and pattern recognition},
  pages={20331--20341},
  year={2024}
}

@article{gao2022monocular,
  title={Monocular dynamic view synthesis: A reality check},
  author={Gao, Hang and Li, Ruilong and Tulsiani, Shubham and Russell, Bryan and Kanazawa, Angjoo},
  journal={Advances in Neural Information Processing Systems},
  volume={35},
  pages={33768--33780},
  year={2022}
}

@article{goli2024romo,
  title={RoMo: Robust Motion Segmentation Improves Structure from Motion},
  author={Goli, Lily and Sabour, Sara and Matthews, Mark and Brubaker, Marcus and Lagun, Dmitry and Jacobson, Alec and Fleet, David J and Saxena, Saurabh and Tagliasacchi, Andrea},
  journal={arXiv preprint arXiv:2411.18650},
  year={2024}
}

@article{karazija2024learning,
  title={Learning segmentation from point trajectories},
  author={Karazija, Laurynas and Laina, Iro and Rupprecht, Christian and Vedaldi, Andrea},
  journal={Advances in Neural Information Processing Systems},
  volume={37},
  pages={112573--112597},
  year={2024}
}

@inproceedings{perazzi2016benchmark,
  title={A benchmark dataset and evaluation methodology for video object segmentation},
  author={Perazzi, Federico and Pont-Tuset, Jordi and McWilliams, Brian and Van Gool, Luc and Gross, Markus and Sorkine-Hornung, Alexander},
  booktitle={Proceedings of the IEEE conference on computer vision and pattern recognition},
  pages={724--732},
  year={2016}
}

@article{jeong2024rodygs,
  title={RoDyGS: Robust Dynamic Gaussian Splatting for Casual Videos},
  author={Jeong, Yoonwoo and Lee, Junmyeong and Choi, Hoseung and Cho, Minsu},
  journal={arXiv preprint arXiv:2412.03077},
  year={2024}
}
\appendix


\end{document}